\documentclass[natbib]{pretty-preprint}
\usepackage[section]{placeins} 
\usepackage{xurl}
\usepackage{fix-cm} 

\title{Video2World:\\Benchmarking Coding Agents for Interactive World Modeling from Embodied Videos}
\runningtitle{Video2World}
\author[1,2*]{Jinzhou Tang}
\author[1*]{Zijun Zhang}
\author[1]{Jing Yang}
\author[1]{Yuchen Yan}
\author[1\dag]{Kun Zhou}
\author[1]{Lingjun Mao}
\author[1]{Ruobing Han}
\author[1]{Jinglin Cao}
\author[1]{Wenpeng Xu}
\author[1]{Lukun He}
\author[1,2]{Minghao Fu}
\author[2]{Fan Feng}
\author[1,2]{Biwei Huang}
\affiliation[1]{Aether AI}
\affiliation[2]{University of California, San Diego}
\contribution[*]{Equal Contribution}
\contribution[\dag]{Corresponding Author and Project Lead}
\date{\today}

\code{https://github.com/AetherLabsAI/Video2World}
\dataset{https://huggingface.co/datasets/AetherLabs-AI/Video2World}
\webpage{https://aetherlabsai.github.io/Video2World}

\definecolor{oursblue}{HTML}{0B6FB8}
\definecolor{oursbg}{HTML}{EAF1FA}
\DeclareRobustCommand{\tabyes}{\textcolor{green!55!black}{\checkmark}}
\DeclareRobustCommand{\tabno}{\textcolor{red!75!black}{$\times$}}

\newcommand{\ours}[1]{\textcolor{oursblue}{#1}}

\usepackage{listings}
\lstdefinestyle{interfacecode}{basicstyle=\ttfamily\footnotesize,
  breaklines=true,breakatwhitespace=false,columns=fullflexible,
  keepspaces=true,showstringspaces=false,frame=single,escapeinside={(*@}{@*)},
  rulecolor=\color{black!20},backgroundcolor=\color{black!2},
  xleftmargin=4pt,xrightmargin=4pt,aboveskip=8pt,belowskip=8pt,
  literate={—}{{---}}3 {–}{{--}}2 {→}{{$\rightarrow$}}1
  {≤}{{$\leq$}}1 {≥}{{$\geq$}}1 {×}{{$\times$}}1
  {≈}{{$\approx$}}1 {°}{{$^\circ$}}1 {…}{{...}}3
  {µ}{{$\mu$}}1 {−}{{$-$}}1 {’}{{'}}1 {“}{{``}}2 {”}{{''}}2}
\newtcolorbox{interfaceprompt}[2][]{enhanced,breakable,
  colback=black!2,colframe=black!22,colbacktitle=black!7,
  coltitle=black,fonttitle=\bfseries\small,fontupper=\small,
  title={#2},title after break={#2 (continued)},boxrule=0.4pt,arc=1mm,
  left=7pt,right=7pt,top=6pt,bottom=6pt,
  before skip=10pt,after skip=10pt,#1}

\abstract{%
Building interactive simulators from real-world observations is a promising way to scale embodied data, but current pipelines still rely heavily on manual environment construction and calibration. We study whether frontier foundation models and coding agents can automate this process end to end. We formulate \emph{autonomous video-to-simulation} as a software engineering task in which an agent observes an embodied video, constructs the corresponding simulated environment and robot behavior, and iteratively refines the result through execution feedback. To evaluate this capability, we introduce \textbf{Video2World}, a benchmark comprising 222 reconstruction instances derived from 189 robot and human demonstration videos. Video2World measures reconstructed worlds along geometric fidelity, dynamic fidelity, and functional correctness, capturing spatial perception, physical reasoning, and executable interaction. Evaluating 9 frontier coding-agent systems reveals a sharp improvement in Task success beginning with Claude Opus 5, rising from below 5\% to over 15\%, while substantial gaps to human-assisted reconstruction remain. 
We further find that worlds that look better could work worse: better visual fidelity does not always lead to higher task success. This echoes the broader gap between perceptual realism and factual correctness observed in generative models.\\ \
}

\begin{document}
\maketitle

\section{Introduction}
\label{sec:introduction}
Embodied AI aims to build intelligent systems that can perceive, reason, and act effectively in the physical world, enabling applications such as robotic manipulation, household assistance, and industrial automation~\citep{duan2022survey}.
Fully acquiring these capabilities requires large amounts of diverse embodied interaction data, which is expensive, time-consuming, and difficult to scale in the real world~\citep{khazatsky2024droid,mandlekar2023mimicgen}.
To alleviate this bottleneck, many prior works have explored interactive simulation as a scalable data engine, where agents can act, observe state transitions, and automatically collect embodied experience~\citep{nasiriany2024robocasa,
wang2024gensim,wang2023robogen}. Conventional Real-to-Sim pipelines, however, still require substantial engineering
effort to reconstruct and configure simulation environments, calibrate geometry and physical properties, align coordinate systems, and integrate perception and simulation components~\citep{torne2024rialto,han2026re3sim,chen2026agenticreal2sim}.

In recent years, foundation models and agents have made rapid progress in spatial perception, visual understanding and planning, demonstrating increasingly strong capabilities in solving complex software and embodied tasks~\citep{yang2024thinking,jimenez2024swebench,xie2024osworld,fu2026capx,you2026embodiedswe}. These advances make it increasingly promising to automate the full process of interactive world modeling in simulators through an agent itself ~\citep{ye2025video2policy,chen2026agenticreal2sim}. Such automation could substantially reduce the human effort, and enable the efficient conversion of real-world observations into scalable interactive environments for data generation, policy learning, and evaluation. However, existing benchmarks typically assess only one or a few isolated capabilities for interactive environment reconstruction, such as spatial understanding, physical rule following, scene reconstruction, or environment editing~\citep{yang2024thinking,wang2026simuscene,gao20263dcodebench,zhao2026sceneactbench,tang2026bvb}. As shown in Table~\ref{tab:benchmark_comparison}, they provide only a fragmented view of performance and do not reveal how effectively current frontier models and agents can integrate these capabilities to autonomously build, execute, and refine interactive worlds.

\begin{table}[t]
    \centering
    \footnotesize
    \setlength{\tabcolsep}{2pt}
    \renewcommand{\arraystretch}{1.05}
    \caption{
        \textbf{Comparison with related benchmarks.}
        \tabyes~yes and \tabno~no denote whether the benchmark focuses on
        evaluating the corresponding capability:
        \emph{Embodied} for embodied reasoning and planning;
        \emph{Physics} for physical-world dynamics and interaction;
        \emph{Spatial} for spatial understanding;
        \emph{Agent} for agentic planning and iterative interaction;
        \emph{Coding} for executable code generation.
    }
    \label{tab:benchmark_comparison}
    \begin{tabular}{@{}llcccccl@{}}
        \toprule
        Benchmark & Input & {Embodied } & {Physics } & {Spatial } & {Agent } & {Coding } & Evaluation Task \\
        \midrule

        SWE-bench~\citeyearpar{jimenez2024swebench}
            & Code Repository
            & \tabno & \tabno & \tabno & \tabno & \tabyes
            & Code Generation \\

        VSI-Bench~\citeyearpar{yang2024thinking}
            & RGB Video
            & \tabno & \tabno & \tabyes & \tabno & \tabno
            & Question Answer \\

        RoboDojo~\citeyearpar{chen2026robodojo}
            & Interactive Env.
            & \tabyes & \tabyes & \tabno & \tabno & \tabno
            & Robot Manipulation \\

        CaP-X~\citeyearpar{fu2026capx}
            & Interactive Env.
            & \tabyes & \tabyes & \tabno & \tabyes & \tabyes
            & Robot Manipulation \\

        RLE-Bench~\citeyearpar{rlebench2026}
            & Interactive Env.
            & \tabyes & \tabyes & \tabyes & \tabyes & \tabyes
            & Robotics Engineering \\

        EmbodiedSWE-Bench~\citeyearpar{you2026embodiedswe}
            & Interactive Env.
            & \tabyes & \tabyes & \tabno & \tabyes & \tabyes
            & Task Completion \\

        SimuScene~\citeyearpar{wang2026simuscene}
            & Text
            & \tabno & \tabyes & \tabno & \tabno & \tabyes
            & Physics Simulation \\

        3DCodeBench~\citeyearpar{gao20263dcodebench}
            & Text / Image
            & \tabno & \tabno & \tabyes & \tabyes & \tabyes
            & 3D Modeling \\

        SceneActBench~\citeyearpar{zhao2026sceneactbench}
            & Image / Video
            & \tabno & \tabno & \tabyes & \tabyes & \tabyes
            & 3D Scene Action \\

        BVB~\citeyearpar{tang2026bvb}
            & RGB Video
            & \tabno & \tabno & \tabyes & \tabyes & \tabyes
            & Editable Env. Rec. \\

        \midrule

        \ours{\textbf{Video2World} (Ours)}
            & \ours{RGB Video}
            & \ours{\tabyes}
            & \ours{\tabyes}
            & \ours{\tabyes}
            & \ours{\tabyes}
            & \ours{\tabyes}
            & \ours{Interactive Env. Rec.} \\

        \bottomrule
    \end{tabular}
\end{table}

To fill this gap and directly test whether foundation models and coding agents can autonomously construct interactive physical worlds, we formulate \emph{autonomous video-to-simulation} 
as an end-to-end software engineering task ~\citep{ye2025video2policy,chen2026agenticreal2sim}. Concretely, given an RGB video of an embodied interaction together with a target simulator, an agent must construct the environment, implement the robot behavior, and iteratively refine the simulation through execution feedback. To support systematic evaluation, we construct a dataset that pairs embodied videos with their corresponding simulator environments, including aligned scene configurations, object states, and interaction trajectories. This paired structure allows us to compare an agent-generated simulation directly against the underlying target environment and assess the full reconstruction quality. In this way, we measure not only whether an agent can understand the world, but also whether it can integrate embodied reasoning, physical understanding, spatial perception, and agentic coding abilities to recover an interactive environment that faithfully reproduces the demonstrated physical world.

We introduce \textbf{Video2World}, a benchmark for evaluating foundation models and coding agents on autonomous interactive world modeling from embodied videos. Video2World comprises 222 reconstruction instances across 39 task families, constructed from 189 video clips spanning robot executions, egocentric human recordings, and third-person human demonstrations. To evaluate reconstruction quality comprehensively, we design metrics from three complementary perspectives: \emph{geometric fidelity}, which measures how accurately the reconstructed scene matches the spatial structure of the source environment; \emph{dynamic fidelity}, which evaluates whether simulated object motions and state transitions reproduce the observed physical evolution; and \emph{functional correctness}, which assesses whether the reconstructed environment supports successful task execution through physical interaction. Together, these metrics reflect an agent's ability to integrate spatial perception, physical reasoning, and executable coding for end-to-end interactive world modeling.

We evaluate 9 frontier coding-agent systems under the same interface and uncover several key findings. First, current coding agents can construct executable simulated worlds, but reliably completing the demonstrated interactions remains challenging.
Second, reconstruction fidelity and functional correctness are related but not equivalent: models with stronger geometric accuracy do not necessarily achieve task success. Third, human-assisted reconstruction provides a substantially stronger reference, while still falls short of perfect task completion. Finally, different coding-agent systems exhibit distinct strengths across the evaluated dimensions, suggesting that overall video-to-simulation performance emerges from their combination rather than any single ability.

\begin{figure}[t]
    \centering
    \includegraphics[width=\columnwidth]{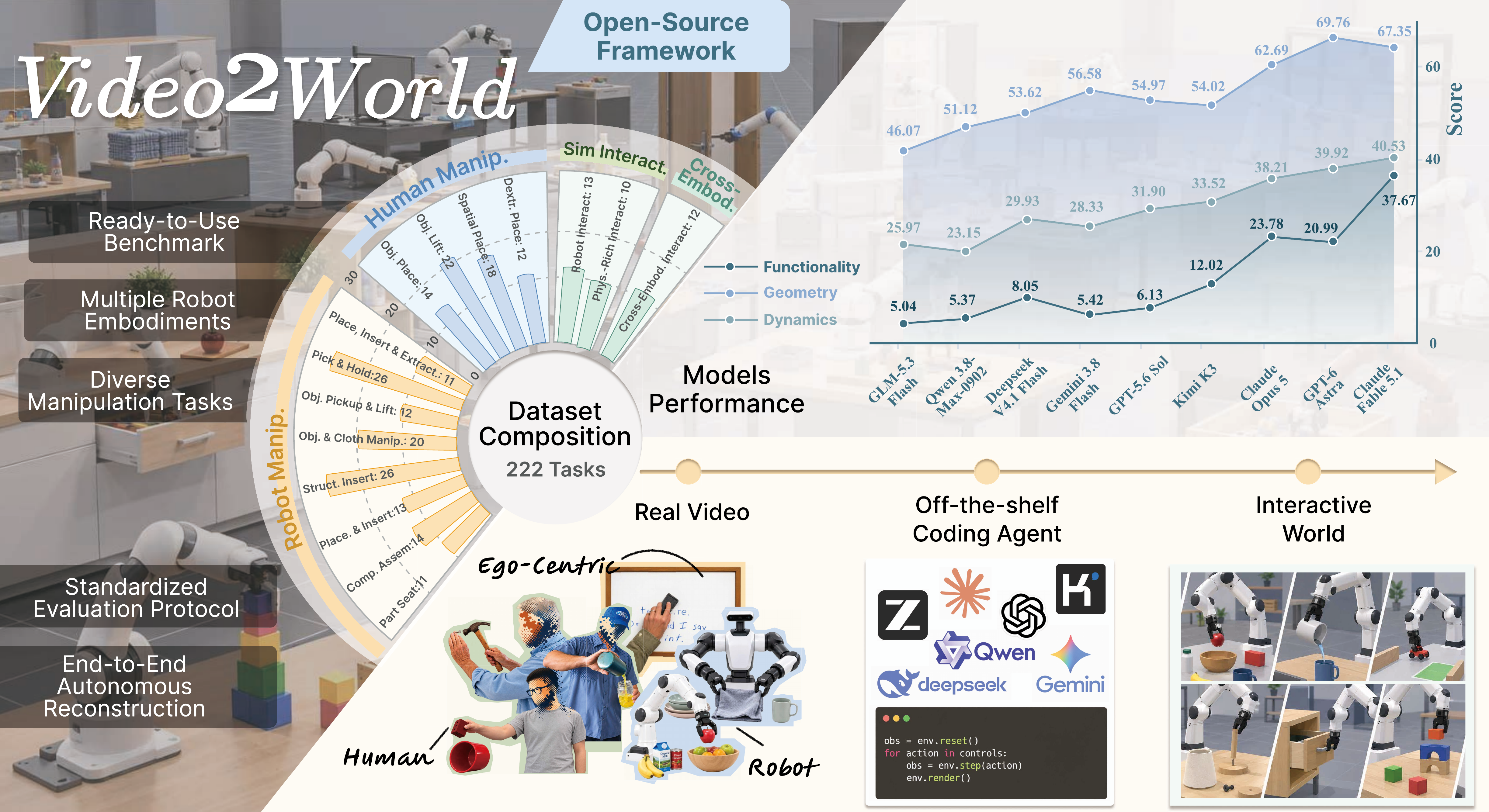}
    \caption{
        \textbf{Video2World} benchmarks the conversion of embodied videos
        into simulated worlds. Given an RGB video and
        simulator tools, a coding agent constructs the environment and
        robot behavior, then revises its implementation using execution
        feedback. Evaluation examines geometric fidelity, physically
        generated object motion, and task completion.
    }
    \label{fig:teaser}
\end{figure}

\section{Preliminary: Autonomous Video-to-Sim via Coding Agents}
\label{sec:formulation}
We study the autonomous video-to-sim task that asks a coding agent to implement an executable simulator program as an interactive environment from an embodied video. The agent must recover task-relevant scene structure, physical behavior, and robot control from visual observations.

\paragraph{Problem Formulation.}
Given an RGB embodied video $v$, a target simulator $\mathcal{S}$, and a set of general-purpose tools $\mathcal{T}$, a coding agent $\mathcal{A}$ produces
\begin{equation}
P=(\eta,\pi)
\sim
\mathcal{A}(\cdot\mid v,\mathcal{S},\mathcal{T}),
\end{equation}
where $\eta$ specifies the reconstructed simulation environment and $\pi$ specifies the robot behavior under the robot embodiment and control interface provided by $\mathcal{S}$. Executing $P$ yields a rollout consisting of a sequence of simulated states and robot actions:
\begin{equation}
\tau_P=(s_0,a_0,s_1,\ldots,a_{T-1},s_T),
\qquad
a_t=\pi(t,s_{\leq t}),
\qquad
s_{t+1}=F_{\eta}(s_t,a_t),
\end{equation}
where $F_{\eta}$ denotes the simulator dynamics instantiated by $\eta$, while $s_t$ and $a_t$ denote the simulated state and robot action at step $t$, respectively. A reconstruction is functionally correct if the resulting rollout reproduces the task demonstrated in $v$ through robot execution in the reconstructed environment.

\paragraph{Agentic Video-to-Sim.}
We implement agentic video-to-sim as an iterative coding process. Given the source video $v$, target simulator $\mathcal{S}$, and tools $\mathcal{T}$, the agent inspects visual observations and available resources, edits the simulation program, executes the current reconstruction, and uses the resulting feedback to revise both the environment $\eta$ and robot behavior $\pi$. This loop allows the agent to progressively correct implementation errors and discrepancies between the simulated interaction and the source demonstration.
Construction terminates upon submission or when the runtime or tool-call budget is exhausted. The resulting program $P=(\eta,\pi)$ is then fixed for evaluation.

\section{Video2World: Benchmarking Coding Agent Video-to-Sim}
\label{sec:benchmark}
Video2World evaluates coding agents for end-to-end reconstruction of interactive physical worlds from embodied videos. We first collect embodied videos and construct the corresponding world assets and annotations, then design a standardized agentic video-to-sim protocol, and finally define evaluation metrics for assessing the reconstructed worlds.

\subsection{Property Construction}
\label{sec:property_construction}
Video2World constructs object-centric world properties from embodied videos to support evaluation of reconstructed interactive environments. We focus on the objects, physical context, motion, and task states that determine how the demonstrated interaction unfolds. These properties provide common evaluation targets across different robot embodiments and control strategies.

\paragraph{Video and Task Collection.}
To cover diverse scenarios and tasks, we assemble 189 video clips forming 222 reconstruction instances from FurnitureBench~\cite{heo2023furniturebench}, DROID~\cite{khazatsky2024droid}, RoboDojo~\cite{chen2026robodojo}, HOI4D~\cite{liu2022hoi4d}, HOT3D~\cite{banerjee2025hot3d}, DexYCB~\cite{chao2021dexycb}, OakInk2~\cite{zhan2024oakink2}, and an in-house household dataset. 
The collection includes both real and simulated demonstrations spanning diverse embodiments, viewpoints, scene configurations, and interaction mechanisms. 
Each reconstruction instance pairs a source video with a target configuration specifying the simulator, robot embodiment, and control interface.
We select clips in which the participating objects, essential interaction process, and task outcome are visually discernible. 
Instances are organized into fixed task families according to
source-specific task categories, target embodiments, and annotation
configurations (Appendices~\ref{app:task_organization}).

\begin{figure}[t]
    \centering
    \includegraphics[width=\columnwidth]{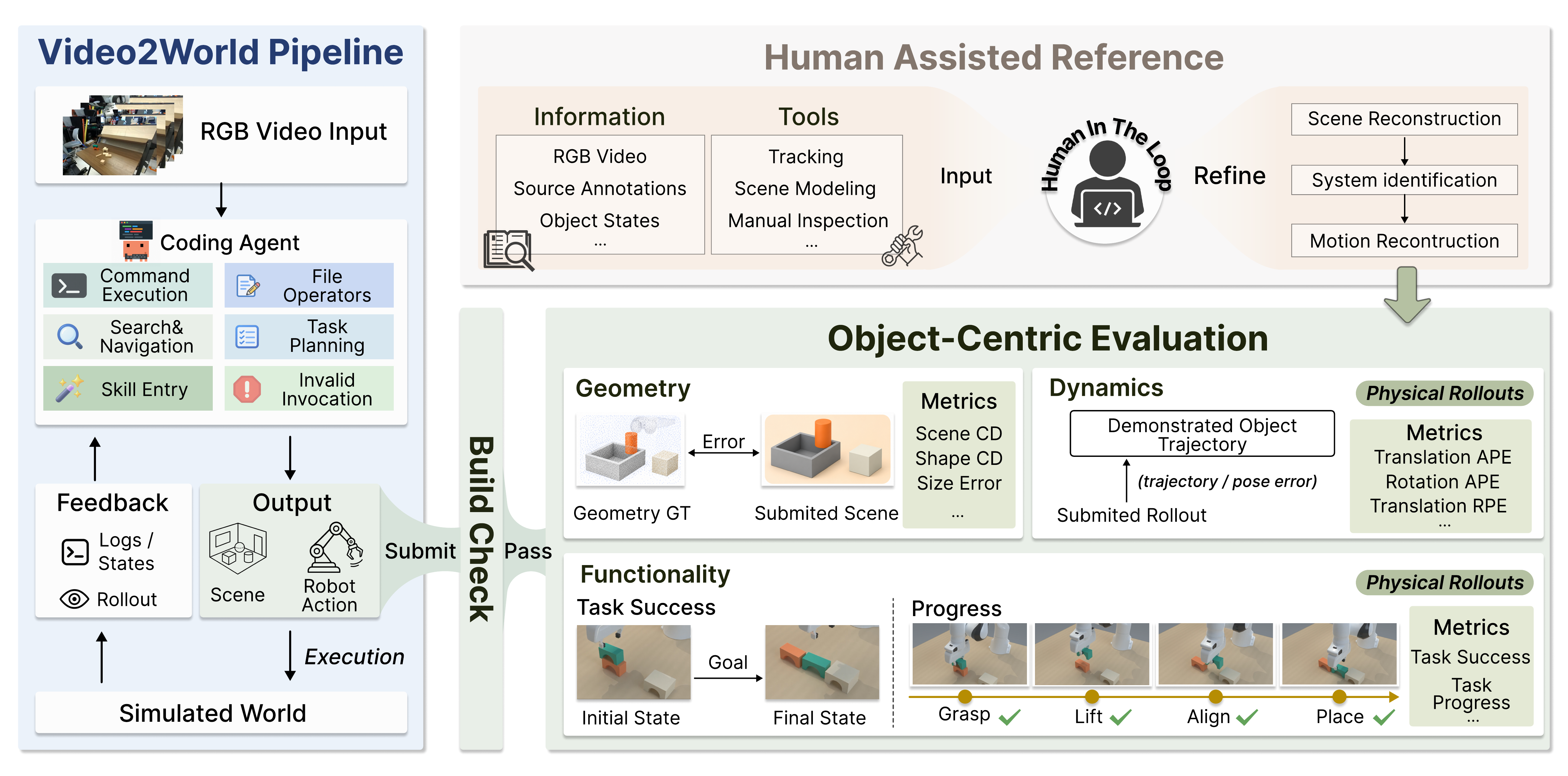}
    \caption{
        \textbf{Overview of Video2World.}
        A coding agent constructs the scene and robot behavior in a
        sandbox, using video inspection, simulator APIs, asset resources,
        and execution feedback. Submitted reconstructions are evaluated
        against hidden scene geometry, demonstrated object motion, and
        task requirements. Evaluation separately measures geometric
        fidelity, dynamic fidelity of object motion, and functional
        correctness under physical execution.
    }
    \label{fig:manicoder_main}
\end{figure}

\paragraph{Object-Centric World Annotations.}
To evaluate whether an agent reconstructs an interactive world rather than merely a visually similar scene, Video2World requires task-relevant properties that capture how the demonstrated interaction is structured and evolves. We therefore represent each video with object-centric annotations covering three aspects: scene geometry, object motion, and task requirements. Concretely, scene geometry records object shape, scale, articulation, and spatial relations; object-motion annotations describe how participating objects move and change state; and task specifications define entity roles, initial conditions, essential interaction events, terminal goals, and progress milestones. These annotations provide a common target for measuring what the agent has reconstructed and reproduced, while remaining independent of the particular simulator implementation or robot control strategy. The annotation sources, hidden specifications, and verification process are described in Appendices~\ref{app:hidden_specifications} and~\ref{app:annotation_verification}.

\paragraph{Cross-Embodiment Alignment.}
Video2World aligns human and robot demonstrations through task-relevant object states and motions, providing common evaluation targets across embodiments and control strategies. Concretely, we derive a reference behavior for each target robot from the hidden annotations and execute it under contact physics to verify that the demonstrated interaction can be physically reproduced. This object-centric alignment preserves shared task-level targets while allowing embodiment-specific robot trajectories. Reference construction is detailed in Appendix~\ref{app:human_assisted}.

As a supplementary check, we compare source and simulated effector motion on 40 human-assisted reference instances from 35 videos, selected by task success in the original evaluation. The six motion statistics show Spearman correlations of 0.74--0.99, including 0.80 for net rotation, suggesting that these reference executions also preserve key motion characteristics beyond object motion (Figure~\ref{fig:effector_motion}). The comparison protocol and metric definitions are provided in Appendix~\ref{app:robot_motion}.

\subsection{Agentic Protocol Design}
\label{sec:agentic_protocol}

With the above collected properties, we devise a standardized protocol for testing how coding agents construct interactive worlds from videos. It controls the agent's available resources, reconstruction budget, and execution setting, while leaving the internal construction strategy fully autonomous. This design evaluates deployable coding-agent systems under a common task interface rather than enforcing a common internal scaffold.

\paragraph{Agent Workspace.}
Each reconstruction instance pairs a source video with a target configuration specifying the simulator, robot embodiment, and control interface. The coding agent operates in an isolated sandbox containing the source video, simulator APIs, documentation, available assets, and general-purpose coding tools. Within this workspace, the agent constructs the simulated environment $\eta_i$ and robot behavior $\pi_i$, and may iteratively inspect simulation outputs and revise its implementation through execution feedback. Each reconstruction instance specifies the source video, target simulator, robot embodiment, and control interface. Construction follows framework-specific resource limits and stopping rules, without a fixed episode-wide tool-call or model-turn budget. More details are specified in Appendices~\ref{app:agent_sandbox}, \ref{app:interface_tools}, and~\ref{app:interface_budget}.

\paragraph{Environment Execution.}
After submission, the reconstructed environment and robot behavior are executed by the evaluator under the assigned simulator, embodiment, and control settings. The evaluator initializes the scene, controls simulation stepping, and records object states, articulation, and contacts for evaluation. After initialization, all object changes must arise through simulator dynamics rather than direct state assignment. Each model receives one independent construction attempt per instance, with submitted code and execution traces retained for reproducibility. More details are in Appendices~\ref{app:simulation_environment} and~\ref{app:submission_protocol}.

\subsection{Evaluation Metrics Design}
\label{sec:evaluation}

We evaluate each reconstructed world from complementary perspectives that capture whether it can be executed, whether it preserves the source scene and physical evolution, and whether it supports the demonstrated interaction. Specifically, Video2World reports executability, geometric fidelity, dynamic fidelity, and functional correctness.

\paragraph{Executability.}
We first measure whether the submitted reconstruction can be successfully executed through the prescribed simulator interface. Build Rate reports the fraction of submissions that produce a valid executable environment and robot behavior under the assigned target configuration.

\paragraph{Geometric Fidelity.}
Geometric fidelity measures how accurately the reconstructed initial scene matches the annotated source environment. We evaluate both scene-level surface correspondence and object-level reconstruction quality, including object shape, size, position, articulation, and relative spatial relations. For scene-level evaluation, let $\mathcal{E}$ denote the annotated non-support entities, $\mathcal S_e$ their reference surfaces, $\widehat{\mathcal C}$ the registered reconstructed non-support surface, and $\mathcal B$ the reference surface including supports. We compute an interaction-weighted geometric error
\begin{equation}
E_{\mathrm{geo}}
=
\sum_{e\in\mathcal E}
\bar w_e\,\mathcal D\!\left(\mathcal S_e,\widehat{\mathcal C}\right)
+
\mathcal D_{\omega}\!\left(\widehat{\mathcal C},\mathcal B\right),
\end{equation}
where $\bar w_e=w_e/\sum_{j\in\mathcal E}w_j$, and $w_e$ assign larger weights to manipulated objects, receiving objects, and nearby structures that directly affect the demonstrated interaction. The directed distances $\mathcal D$ and $\mathcal D_{\omega}$ use the capped median and capped weighted-median definitions in Appendix~\ref{app:geometric_fidelity}, respectively, with candidate-point weights $\omega$ inherited from the nearest reference non-support entity. Object-level metrics further report shape, scale, position, and relative-pose errors. Full surface-sampling, registration, and object-level definitions are provided in Appendix~\ref{app:geometric_fidelity}.

\begin{table}[t]
\centering
\footnotesize
\setlength{\tabcolsep}{2.5pt}
\renewcommand{\arraystretch}{1.08}
\caption{
\textbf{Overall performance on Video2World.}
V2WScore combines Build, functionality, geometry, and dynamics into a single 0--100 score, macro-averaged across task families. Build measures executability; Functionality reports task success and stage-wise progress; Geometry measures scene- and object-level reconstruction accuracy; and Dynamics reports translation APE, rotation APE, and translation RPE. The human-assisted reference additionally uses source information and manual refinement. Best and second-best automated results are shown in bold and underlined, respectively; ties share the same formatting. Metric definitions, deformation and partial-observation terms, aggregation, and missing-data handling are detailed in Appendices~\ref{app:v2w_score} and~\ref{app:aggregation}.
}
\label{tab:main_results}

\begin{tabular}{lcccccccccc}
\toprule
& \multicolumn{1}{c}{V2WScore}
& \multicolumn{1}{c}{Build}
& \multicolumn{2}{c}{Functionality}
& \multicolumn{3}{c}{Geometry}
& \multicolumn{3}{c}{Dynamics} \\
\cmidrule(lr){2-2}
\cmidrule(lr){3-3}
\cmidrule(lr){4-5}
\cmidrule(lr){6-8}
\cmidrule(lr){9-11}
Method
& \shortstack{Score\\(0--100) $\uparrow$}
& \shortstack{Rate\\(\%) $\uparrow$}
& \shortstack{Success\\(\%) $\uparrow$}
& \shortstack{Progress\\(0--1) $\uparrow$}
& \shortstack{Scene CD\\(cm) $\downarrow$}
& \shortstack{Shape CD\\(cm) $\downarrow$}
& \shortstack{Size err.\\(cm) $\downarrow$}
& \shortstack{T-APE\\(cm) $\downarrow$}
& \shortstack{R-APE\\(deg) $\downarrow$}
& \shortstack{T-RPE\\(cm) $\downarrow$} \\
\midrule
Human-assisted Ref & 74.24 & 100.0 & 58.8 & 0.70 & 2.19 & 0.15 & 0.43 & 9.40 & 41.09 & 2.42 \\
\midrule
Claude Fable 5.1 & \textbf{48.52} & \textbf{93.1} & \textbf{25.5} & \textbf{0.53} & \textbf{5.35} & \underline{1.00} & \underline{2.78} & \textbf{18.71} & 104.61 & \underline{8.25} \\
GPT-6 Astra & \underline{43.55} & \textbf{93.1} & 10.7 & 0.33 & 6.23 & \textbf{0.69} & \textbf{1.62} & 20.78 & \textbf{95.08} & \textbf{6.19} \\
Claude Opus5 & 41.56 & \underline{92.8} & \underline{16.0} & \underline{0.34} & \underline{6.01} & 1.14 & 3.13 & \underline{19.37} & 110.85 & 8.42 \\
Kimi K3 & 33.19 & 92.0 & 4.8 & 0.21 & 8.08 & 1.41 & 4.23 & 27.28 & 111.16 & 14.29 \\
GPT-5.6 Sol & 31.00 & 92.5 & 2.5 & 0.11 & 7.79 & 1.37 & 3.76 & 31.68 & 105.03 & 18.00 \\
DeepSeek-V4.1 Flash & 30.54 & 92.4 & 4.1 & 0.13 & 8.59 & 1.50 & 4.13 & 34.64 & 104.52 & 20.85 \\
Gemini-3.8 Flash & 30.11 & \textbf{93.1} & 2.0 & 0.09 & 8.57 & 1.25 & 3.67 & 37.97 & 114.82 & 28.65 \\
Qwen-3.8 Max-0902 & 26.55 & 87.6 & 1.5 & 0.11 & 8.36 & 1.51 & 4.13 & 41.28 & 110.06 & 27.38 \\
GLM-5.3 Flash & 25.69 & 87.7 & 2.0 & 0.09 & 9.58 & 1.62 & 5.04 & 41.16 & \underline{103.83} & 30.16 \\
\bottomrule
\end{tabular}
\end{table}

\paragraph{Dynamic Fidelity.}
Dynamic fidelity measures how closely physical execution in the reconstructed environment reproduces the object motion observed in the source video. For object $e$, let $p_e(t)$ and $p_e^*(t)$ denote its simulated and demonstrated positions at time $t$. We measure translational motion discrepancy relative to each trajectory's initial state:
\begin{equation}
E_{\mathrm{dyn}}^{e}
=
\frac{1}{|\mathcal T|}
\sum_{t\in\mathcal T}
\rho_{c_p}\!\left(
\left\lVert
\Delta p_e(t)-\Delta p_e^{*}(t)
\right\rVert_2
\right).
\end{equation}
Here $\Delta p_e(t)=p_e(t)-p_e(0)$ and $\Delta p_e^*(t)=p_e^*(t)-p_e^*(0)$, $\mathcal T$ contains the comparison times after temporal correspondence, and $\rho_{c_p}(r)=\min\{r,c_p\}$ with $c_p=1\,\mathrm m$, following Appendix~\ref{app:dynamic_fidelity}. This removes global translation offsets while preserving differences in motion magnitude and temporal evolution. We additionally evaluate rotational motion, account for known object symmetries, and use frame-level surface correspondence for deformable objects. These metrics focus on task-relevant object dynamics, allowing different control strategies to realize the same physical interaction.

\paragraph{Functional Correctness.}
Functional correctness evaluates whether the reconstructed world supports successful execution of the demonstrated task. For each task family, we define object-centric requirements covering task-defining initial conditions, essential interaction events and their ordering, and the terminal goal. We report both Task Success and graded Task Progress through intermediate milestones, allowing different embodiments, trajectories, and execution timings as long as the required interaction is physically realized. More details are in Appendix~\ref{app:functional_correctness}.

\begin{figure}[t]
    \centering
    \includegraphics[width=\columnwidth]{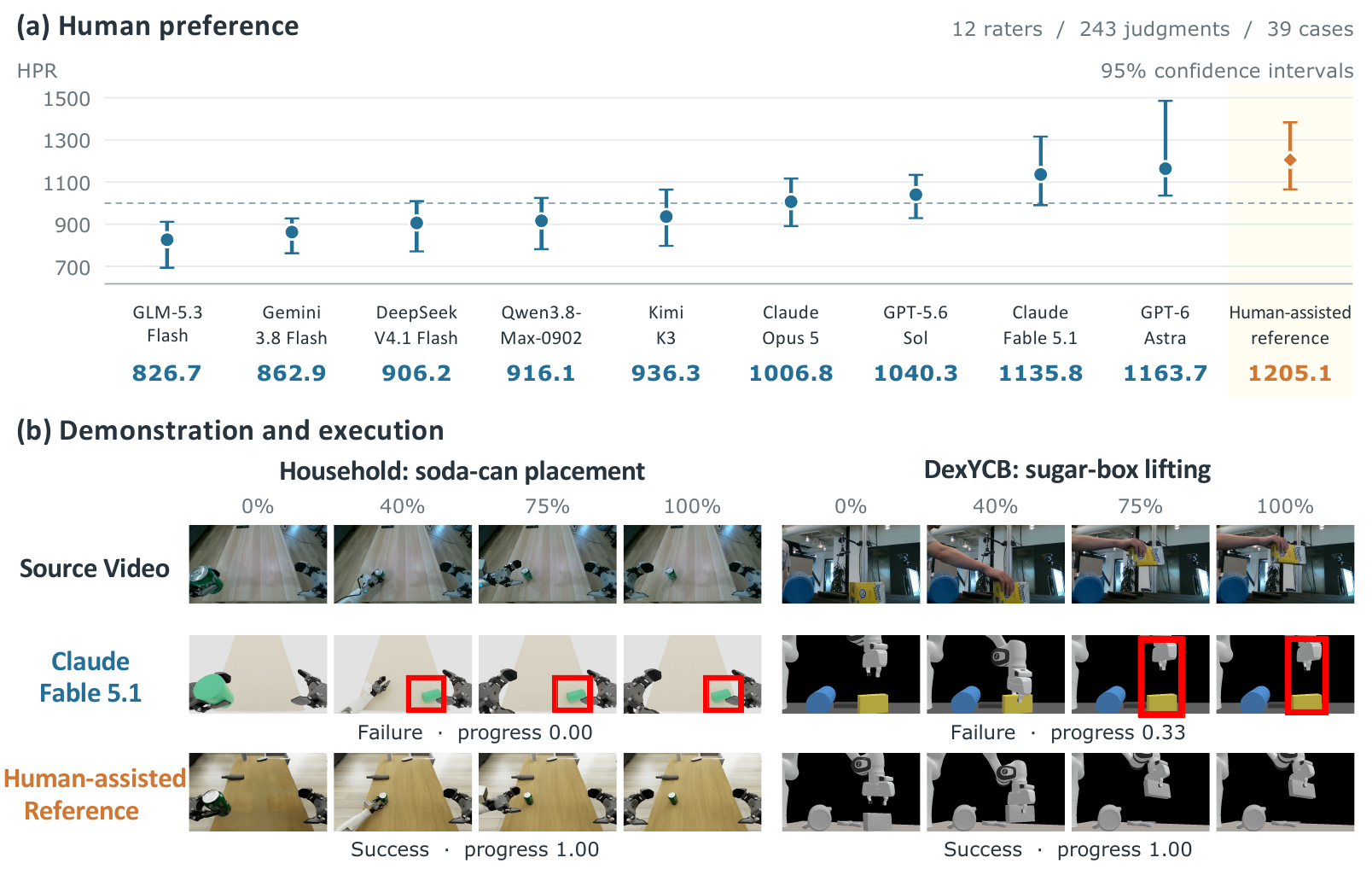}
    \caption{
        \textbf{Human in the loop reference and evaluation}
        (a) Human preference ratings of coding models and the human-assisted reference, based on 243 judgments from 12 raters across 39 cases. Error bars indicate 95\% confidence intervals.
        (b) Source demonstrations and simulated executions for two representative tasks. Each method executes its own behavior in its reconstructed environment. Outcomes and task progress are reported below each simulated sequence; red boxes highlight failed executions.
    }
    \label{fig:human_comparison}
\end{figure}

\section{Experiments}
\label{sec:experiments}

We investigate the reliability of autonomous video-to-simulation via coding agents, the performance relative to human-assisted reconstruction, the relationship between geometric fidelity and functional correctness, and the effector-motion correspondence in selected HAR executions.

\subsection{Experimental Setup}
\label{sec:experimental-setup}

\paragraph{Coding agents.}
We evaluate 9 coding-agent models on 222 instances, covering 189 video clips and 39 task families. Seven models run through OpenCode, while GPT-6 Astra and Fable-5.1 use Codex CLI and Claude Code, respectively. We evaluate each configuration as a complete agent system, including its underlying model, coding scaffold, and inference settings. Each model makes one autonomous construction attempt per instance under the benchmark interface and assigned target configuration (Section~\ref{sec:benchmark}). All agents have access to physical simulation and rendered rollout feedback and may iteratively revise both the scene and robot behavior within a 180-minute wall-clock limit. The submission retained by the prescribed stopping rule is frozen for evaluation, without human correction or selection based on hidden scores. More details are in Appendix~\ref{app:environment}.

\paragraph{Human-assisted reference.}
HAR leads across the reported automatic metrics and obtains the highest Human Preference Rating (HPR) point estimate in our pairwise study of overall scene-and-interaction faithfulness (Figure~\ref{fig:human_comparison}a). The study protocol, participant information, and rating procedure are detailed in Appendix~\ref{app:human_evaluation}. Successful HAR rollouts further demonstrate that the target tasks are feasible under the benchmark configuration, including cases where coding agents produce incomplete or unsuccessful executions (Figure~\ref{fig:human_comparison}b). However, HAR itself does not achieve uniformly reliable task completion, showing that improved scene knowledge and manual refinement do not eliminate the difficulty of reconstructing executable behavior.

\begin{figure}[t]
    \centering
    \includegraphics[width=\columnwidth]{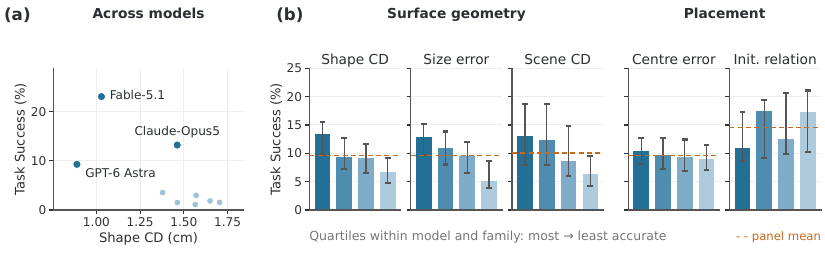}
    \caption{
        \textbf{Geometric fidelity versus functional correctness.} (a) System-level Shape CD against Task Success; both are macro-averaged over task families, and Task Success counts failed builds as failures. (b) Task Success across geometric-error quartiles formed within each model and task family, averaged equally across families and then models. Dashed lines mark the mean of each panel; error bars show 95\% source-video bootstrap intervals. The initial relation is the object--receiver relation before execution, since the terminal relation enters the success predicate.
    }
    \label{fig:geometry_function}
\end{figure}

\paragraph{Evaluation.}
We report V2WScore, a 0--100 aggregate of Build, functionality, geometry, and dynamics, alongside its component
metrics. The score is computed independently for each method, which executes its own robot behavior in its
reconstructed environment. Task judgments use the recorded rollout and configuration-specific physical-validity
checks. Build Rate, Task Success, and Task Progress are aggregated over all applicable instances, with failed
builds assigned zero. Continuous errors are computed on evaluable comparisons, with coverage reported. Each metric is averaged within task families and then across eligible families with equal weight. Reported 95\% confidence intervals use stratified source-cluster bootstrapping, retaining related clips and target configurations together. 
More details and experiments are in Appendix~\ref{app:evaluation} and Appendix~\ref{app:world_model}.

\subsection{Main Results}
\label{sec:main-results}

\paragraph{Coding agents build executable worlds but remain unreliable at full task completion.}
Most agent-generated reconstructions can be simulated, as reflected by high Build Rates, but Task Progress consistently exceeds Task Success (Table~\ref{tab:main_results}), showing that agents often achieve partial interaction without completing all required conditions. Even Fable-5.1, the strongest model by V2WScore, exhibits this gap. Moreover, strong object reconstruction does not necessarily yield stronger functional performance: GPT-6 Astra leads in object shape and size accuracy, while Fable-5.1 achieves better scene-level and task performance. These results highlight a gap between reconstructing individual assets and jointly recovering a scene and behavior that support successful interaction.

\paragraph{Human-assisted reconstruction provides a strong but empirical reference.}
HAR leads across the reported automatic metrics and is also preferred by human evaluators (Figure~\ref{fig:human_comparison}a), supporting its use as a high-quality reference. Successful HAR rollouts further demonstrate that the target tasks are feasible under the benchmark configuration, including cases where coding agents produce incomplete or unsuccessful executions (Figure~\ref{fig:human_comparison}b). However, HAR itself does not achieve uniformly reliable task completion, showing that improved scene knowledge and manual refinement do not eliminate the difficulty of reconstructing executable behavior.

\begin{figure}[t]
    \centering
    \includegraphics[width=\columnwidth]{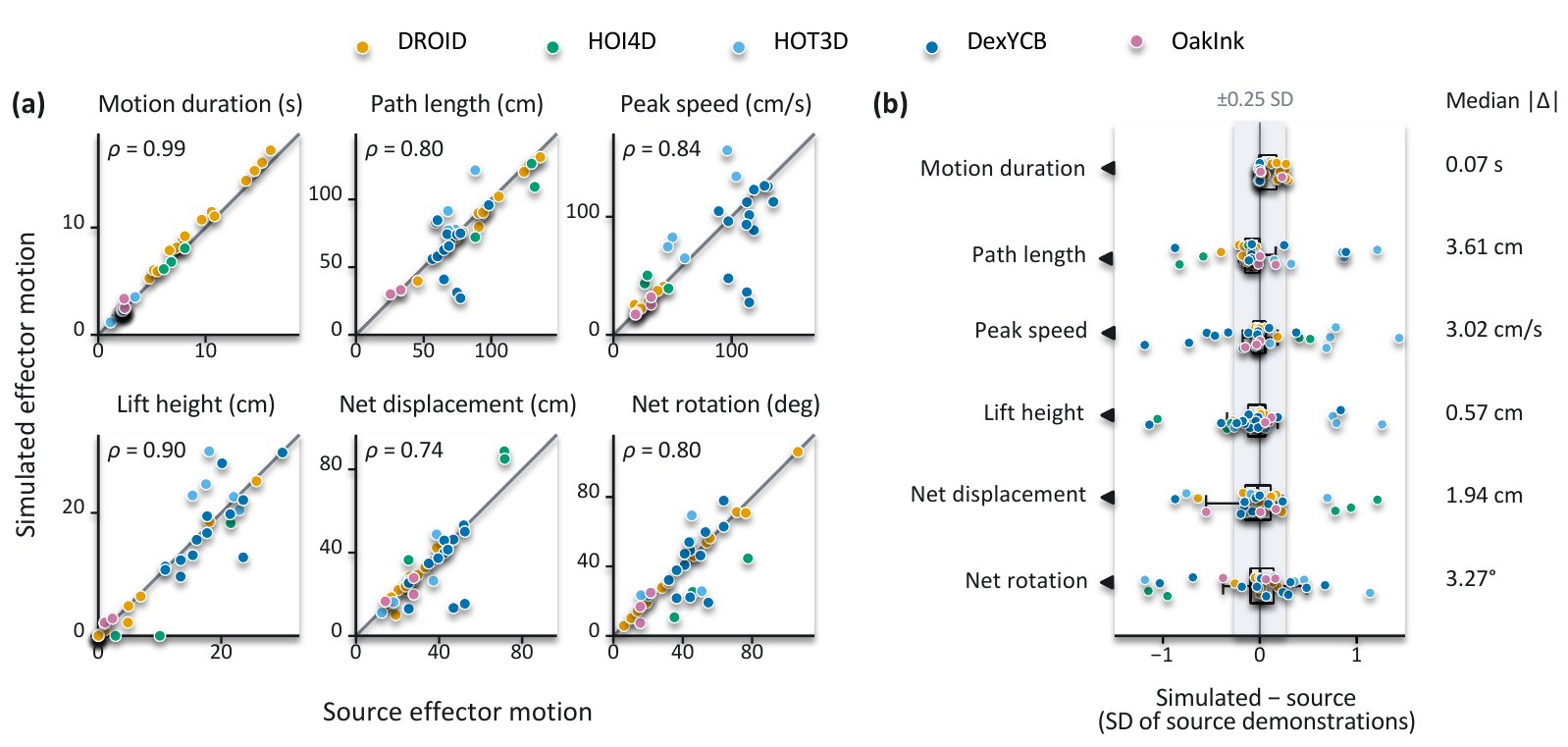}
    \caption{
        \textbf{Robot Effector Motion Beyond Object-Centric Evaluation.}
        (a) Source versus simulated effector statistics for 40 human-assisted reference instances from 35 videos, selected by task success in the original evaluation and colored by source. We compare gripper TCPs on DROID and human wrists with robot hand roots on human-video sources. Diagonals indicate equality; $\rho$ is Spearman correlation.
        (b) Simulated-minus-source differences normalized by the source sample SD. Boxes show median/IQR; whiskers use $1.5\times$ IQR. Shading marks $\pm0.25$ SD; arrows count off-scale points; the right column gives median absolute differences. See Appendix~\ref{app:robot_motion} for definitions.
    }
    \label{fig:effector_motion}
\end{figure}

\paragraph{Geometric fidelity helps but does not guarantee functional correctness.}
Lower shape, size, and scene errors are generally associated with higher Task Success within models and task families (Figure~\ref{fig:geometry_function}b), yet many geometrically accurate reconstructions still fail. The system-level comparison shows the same mismatch: GPT-6 Astra achieves the strongest object-shape fidelity, whereas Fable-5.1 completes more tasks (Figure~\ref{fig:geometry_function}a). Conversely, some reconstructions remain functional despite spatial disagreement with the source, since robot behavior can adapt to the reconstructed layout. Geometry and functionality should therefore be evaluated separately to distinguish faithful reconstruction from task-effective reconstruction.

\paragraph{Successful reconstructions also recover key motion characteristics.}
We compare effector-motion statistics between source demonstrations and successful simulated reference executions, including motion duration, path length, peak speed, lift height, net displacement, and net rotation. Figure~\ref{fig:effector_motion} shows broadly similar motion characteristics across the two, providing additional evidence that successful reconstructions capture not only terminal states but also important temporal and spatial properties of the demonstrated interaction. Detailed statistics are provided in Appendix~\ref{app:robot_motion}.

\section{Related Work}
\label{sec:related-work}

\paragraph{Real-to-Sim Reconstruction for Robot Manipulation.}
Real-to-Sim methods reconstruct interactive simulations from real-world observations for robot learning and evaluation. RialTo and Re$^3$Sim reconstruct environments for policy learning and transfer~\citep{torne2024rialto,han2026re3sim}, while PhysTwin recovers deformable-object models that reproduce observed interactions~\citep{jiang2025phystwin}. GenSim and RoboGen use foundation models to generate tasks, environments, and robot behaviors for scalable simulation-based learning~\citep{wang2024gensim,wang2023robogen}. More closely related, Video2Policy reconstructs manipulation tasks from internet videos and uses language models to generate and refine rewards for policy learning~\citep{ye2025video2policy}. Agentic Real-to-Sim coordinates vision-language agents with specialized perception and simulation tools to convert recorded interactions into executable episode twins~\citep{chen2026agenticreal2sim}. 

\paragraph{Foundation-Model Agents for Embodied Tasks.}
Foundation models connect semantic reasoning with embodied action through planning, code generation, and execution feedback. SayCan grounds language-based planning in robotic skill affordances~\citep{ahn2022icanisay}, while PaLM-E integrates visual and continuous sensor observations for embodied reasoning~\citep{driess2023palme}. Code as Policies generates executable programs that compose perception and control APIs~\citep{liang2023codepolicies}, and VoxPoser constructs spatial value maps through language-model-generated code~\citep{huang2023voxposer}. Voyager further demonstrates how execution feedback enables program revision and reusable skill acquisition in Minecraft~\citep{wang2023voyager}.

\paragraph{Benchmarks for Coding and Agentic Construction.}
Coding-agent benchmarks evaluate program execution and refinement. SWE-bench measures repository-level issue resolution~\citep{jimenez2024swebench}; embodied benchmarks include CaP-X for robot-control programs under varied interfaces and observations~\citep{fu2026capx}, RLE-Bench for robot learning engineering~\citep{rlebench2026}, and EmbodiedSWE-Bench for long-horizon manipulation and other physical tasks~\citep{you2026embodiedswe}. Construction benchmarks cover procedural modeling from text and images in 3DCodeBench~\citep{gao20263dcodebench}, visually conditioned scene operations against hidden geometric targets in SceneActBench~\citep{zhao2026sceneactbench}, and code generation for text-specified physical scenarios in SimuScene~\citep{wang2026simuscene}. Particularly related, BVB evaluates video reconstruction as animated Blender scenes through spatiotemporal facts and perceptual similarity~\citep{tang2026bvb}. 

\section{Conclusion}
We introduced \textbf{Video2World}, a benchmark for autonomous reconstruction of interactive physical worlds from embodied videos. Video2World formulates video-to-simulation as an end-to-end software engineering task and evaluates geometric fidelity, dynamic fidelity, and functional correctness. Our experiments show that current coding agents can often build executable simulations, yet reliable task completion remains challenging. We further find that geometric fidelity does not fully determine functional correctness, and even human-assisted reconstruction remains imperfect. These results highlight interactive world reconstruction as a joint scene--behavior problem and motivate execution-grounded evaluation beyond individual fidelity metrics.

\bibliographystyle{plainnat}
\bibliography{reference}

\beginappendix
\setcounter{figure}{0}
\setcounter{table}{0}
\renewcommand{\thefigure}{A\arabic{figure}}
\renewcommand{\thetable}{A\arabic{table}}
\renewcommand{\theHfigure}{appendix.\arabic{figure}}
\renewcommand{\theHtable}{appendix.\arabic{table}}

\section{Environment and Agent Interface}
    \label{app:environment}
    \raggedbottom

    This section specifies the interface through which a coding agent constructs and submits a reconstruction to Video2World. A benchmark instance fixes an RGB video, a target embodiment, an execution backend, and a public task brief. The agent estimates the scene and supplies robot controls; the benchmark reconstructs the submitted environment and measures the resulting physical interaction. It is implemented using the scene, control, and validation facilities of \textsc{Video2Sim}.

    \paragraph{Instantiation of the reconstruction problem.}
    The environment description $\eta$ contains the reconstructed geometry, collision shapes, material and inertial properties, passive mechanisms, and initial object configuration, together with the estimated camera and robot placement required by the assigned interface. The target simulator $\mathcal{S}$ fixes the embodiment, admissible controls, and execution rules. The initial physical state $s_0$ is obtained from the submitted configuration using the initialization procedure specified below. Camera parameters determine rendered observations; they do not prescribe object motion. Thus, $P=(\eta,\pi)$ is an executable scene-and-control package interpreted by the benchmark's simulator, with the agent estimating the scene and behavior within a fixed execution contract.

    \paragraph{Representation of robot behavior.}
    In the implemented interface, the agent specifies a finite sequence of control references $U=(u_0,\ldots,u_{N-1})$ and a control interval $\Delta t$. Each reference is a Cartesian target or a vector of joint and gripper commands, as detailed in Table~\ref{tab:simulation_configurations}. The behavior $\pi$ combines these references with the fixed controller for $\mathcal{S}$, which performs the applicable interpolation, inverse kinematics, actuator control, and contact-dependent waits. The general state-dependent notation $\pi(t,s_{\leq t})$ therefore includes feedback supplied by this controller; the submitted behavior is a control sequence, rather than an arbitrary policy callback. The coding agent may revise $U$ after inspecting a construction rollout, but does not revise the frozen sequence during final evaluation.

    \paragraph{State and execution time.}
    The simulated state comprises robot and object configurations and velocities, together with the simulator state required to advance their dynamics. Robot actions $a_t$ in the rollout describe physical execution and are distinct from the coding agent's tool operations during construction. The transition $F_{\eta}$ is understood with the assigned simulator and controller fixed; a recorded transition can span multiple physics integration steps. Source-video frames, submitted control samples, and recorded simulation states consequently need not have equal counts or timestamps. Interpolation, initialization, contact-dependent waits, and terminal settling determine the executed clock. The declared control interval and recorded state timestamps, rather than the video frame index, determine the timing of the reconstruction.

    \subsection{Public Inputs and Information Boundary}
    \label{app:agent_sandbox}

    \paragraph{Episode inputs.}
    The agent receives a local copy of the source video, its image dimensions and frame count, and the instructions for the assigned embodiment. Video inspection, frame extraction, geometric estimation, numerical optimization, and local simulation are available through ordinary coding tools. No particular perception or planning pipeline is prescribed. The agent must infer object geometry, metric scale, initial poses, camera parameters, and robot behavior from the permitted observations. For an egocentric recording, later measurements are expressed in the first camera frame after accounting for camera motion using the static scene.


    \paragraph{Public simulator and robot information.}
Public resources specify the target simulator, robot geometry, joint limits, control conventions, and submission requirements. Agents infer the task objective, participating objects, semantic roles, and interaction sequence from the source video. No additional task descriptions, semantic role annotations, or task-specific success criteria are supplied. Source calibration, reference geometry, measured object trajectories, reference controls, and benchmark scores remain hidden.

    \begin{table}[!htbp]
        \centering\small
        \caption{\textbf{Agent-visible resources and withheld information.} Public resources are selected by the assigned configuration. Candidate observations concern the agent's own reconstruction.}
        \label{tab:sandbox_resources}
        \renewcommand{\arraystretch}{1.15}
        \begin{tabularx}{\linewidth}{@{}p{0.23\linewidth}X@{}}
            \toprule
            Resource & Interface specification \\
            \midrule
            Source observation & RGB video; resolution, frame count, and the sampling information stated in the task brief. No depth or measured source calibration. \\
            Public specification & Target simulator and robot; simulator documentation, robot assets, joint limits, control conventions, and submission requirements. \\
            Workspace tools & Shell execution; reading, writing, and editing files; filename and text search. Image-capable file reads expose inspected source frames or candidate renders. \\
            Construction feedback & Program output and errors; package-validation errors; candidate physical rollouts, rendered observations, and available execution diagnostics. \\
            Withheld information & Reference scene and object meshes, measured source trajectories, source robot logs, hidden camera calibration, reference submissions, and benchmark scores. \\
            \bottomrule
        \end{tabularx}
    \end{table}

    \paragraph{Workspace and observations.}
    The standard construction driver stages the video separately from the reference data and allocates episode-specific output and temporary directories. Instructions restrict episode-specific reads to the staged inputs and prohibit obtaining the source dataset's ground truth or calibration. Construction transcripts are retained and audited for accesses to protected sample paths. These are workspace restrictions and an access audit; the audit alone does not establish operating-system isolation. A tool observation consists of the content actually returned to the agent: file contents, search results, execution output, errors, or inspected images. A generated image or video becomes a visual observation only when it is opened through an image-capable interface. Hidden benchmark scores are never included in the standard validation-repair loop.

    \paragraph{Initialization of a benchmark episode.}
    A reconstruction trial starts from the selected video and embodiment in the experiment manifest. Before launching the agent, the benchmark verifies the input video and public-resource versions, creates a fresh workspace and conversation, and supplies the fully instantiated task brief. The source video and released resources are staged separately from evaluation annotations; previous candidates and conversations are not reused. Thus, reproducing a trial fixes both the video bytes and the assigned configuration, together with the prompt revision, public assets, agent invocation, and evaluation version. The supported configurations in Table~\ref{tab:simulation_configurations} describe the available interface; the selected experiment manifest determines which instances are run.

    \paragraph{Execution environment for the model comparison.}
    The seven OpenCode configurations in Table~\ref{tab:agent_models} use a benchmark-managed executor that additionally enforces filesystem and network restrictions. Public inputs are read-only, and each agent receives one CPU, 4 GiB of memory, and no GPU device access. Network connections are restricted to the model API proxy and the submission service. Web access, external plugins, human clarification, and delegated agents are disabled. These resource limits describe the coding process. In the current protocol, every configuration also provides physical rollout feedback during construction, as specified in Section~\ref{app:interface_budget}; lack of direct GPU access by the coding process does not remove this requirement. Final evaluation is performed separately with access to the withheld annotations. Astra and Fable use externally launched coding-agent sessions, whose configuration and delivery procedure are specified below.

    \paragraph{Visual observations and media limits.}
    The public frame sampler defaults to 16 evenly spaced frame indices, including both endpoints, with original timestamps retained. Images are resized to a maximum side length of 1,024 pixels and encoded as JPEG at quality 85. An extraction may contain 2--48 frames and must write to a new output directory. For the OpenCode configurations, each model request admits at most 16 images, 4 MB per image, and 20 MB of aggregate image data before base64 encoding; the serialized request is limited to 30 MB. Context management preferentially retains recent images and the first image of each source within this budget. Removed images are represented by visible placeholders and may be inspected again. Frame extraction and image observation are separate operations: only images actually returned by a read or attachment enter the model's visual context.


    \subsection{Simulation Configurations and Control Conventions}
    \label{app:simulation_environment}

    Table~\ref{tab:simulation_configurations} summarizes the simulation configurations and control interfaces for the 222 benchmark instances. The two human-video configurations use the same 33 source clips with different target embodiments. All configurations are included in the main evaluation.

    \begin{table}[!htbp]
        \centering\small
        \caption{\textbf{Registered candidate interfaces.} $N$ is the number of reconstruction instances; the total is 222. Frequencies refer to submitted controls, not physics integration. For each floating hand, the seventh palm value records the open/closed convention; its twenty finger angles determine the actual posture.}
        \label{tab:simulation_configurations}
        \setlength{\tabcolsep}{4pt}
        \renewcommand{\arraystretch}{1.16}
        \begin{tabularx}{\linewidth}{@{}>{\raggedright\arraybackslash}p{0.19\linewidth}r >{\raggedright\arraybackslash}p{0.20\linewidth}>{\raggedright\arraybackslash}X@{}}
            \toprule
            Configuration & $N$ & Simulator / robot & Submitted control \\
            \midrule
            Furniture assembly & 64 & SAPIEN / Panda & TCP position, orientation, and gripper; 7 values at 5 Hz. \\
            DROID tasks & 20 & SAPIEN / Panda with Robotiq 2F-85 & TCP position, orientation, and gripper; 7 values at 5 Hz. \\
            Human video to arm & 33 & SAPIEN / Panda & TCP position, orientation, and gripper; 7 values at 5 Hz. \\
            Human video to hands & 33 & SAPIEN / paired Wuji hands & Two palm streams of 7 values and two finger streams of 20 angles; synchronized timestep. \\
            Planar pushing & 3 & SAPIEN / xArm7 with pusher & TCP position and orientation at 5 Hz; seventh value unused. \\
            RoboDojo & 25 & Isaac Sim / dual X5 or dual xArm7 & 14 or 16 joint/gripper values at 25 Hz. \\
            Household manipulation & 37 & Isaac Sim / G1 with Omnipicker grippers & 20 joint/gripper values at 20 Hz. \\
            Rope and soft toy & 7 & MuJoCo / xArm7 & 7 arm angles and 1 gripper angle; declared control timestep. \\
            \bottomrule
        \end{tabularx}
    \end{table}

    \paragraph{Units, poses, and cameras.}
    Distances are in metres, time in seconds, and joint and orientation angles in radians. Rigid poses use a translation and a scalar-first unit quaternion, $(x,y,z,q_w,q_x,q_y,q_z)$. Camera coordinates follow the OpenCV convention: $+x$ points right, $+y$ down, and $+z$ forward. A transform ${}^{A}\!T_B$ maps coordinates from frame $B$ to frame $A$. The agent estimates the camera parameters requested by the configuration at the delivered image resolution. The rigid camera-frame profiles estimate focal lengths and principal point; RoboDojo assumes square pixels and a centred principal point and asks for the focal length. Resizing and letterbox padding must be accounted for when projecting onto the video.

    For the camera-frame rigid and human-video interfaces, the declared camera-to-delivery transform is the identity. The evaluator applies the withheld registration between the source camera and the evaluation frame; this registration is not returned to the agent. Candidate self-simulation instead uses the agent's own estimate of the robot base pose. Native RoboDojo and household submissions explicitly estimate the camera-to-public-world transform, and evaluation uses that estimate rather than substituting hidden calibration. Joint commands remain in the named robot joint frames.

    \paragraph{Single-arm Cartesian control.}
    A submitted row is
    \begin{equation}
        u_k=(x_k,y_k,z_k,\alpha_k,\beta_k,\gamma_k,g_k),\qquad
        R_k=R_z(\gamma_k)R_y(\beta_k)R_x(\alpha_k).
        \label{eq:appendix_tcp}
    \end{equation}
    The position denotes the tool centre point (TCP); for the parallel-gripper interfaces this is the midpoint between the fingertip pads, with the approach direction along local $+z$ and the opening direction along local $+y$. A gripper command of one means open and zero means closed. The first row specifies the initial tool configuration. The public brief recommends successive translations no larger than 3 cm and rotations no larger than 0.3 rad, and a stationary terminal segment. These are motion-planning recommendations, distinct from the runtime's numerical validity checks.

    For FurnitureBench, rigid DROID, and planar pushing, the benchmark registers the camera-frame submission using the withheld source calibration and executes the controls with the assigned robot at its evaluation placement. Inverse kinematics produces joint targets, with four interpolated joint-control updates between consecutive submitted poses. For the prescribed 5 Hz input, joint control is 20 Hz and physics integration is 100 Hz. Reset uses seed zero. An initially open gripper is followed by 0.5 s of settling. An initially closed stream omits this settling and performs grasp initialization, with closing preparation of up to 2 s for the Panda or 1.5 s for the Robotiq. Closing transitions can insert additional waits: up to 1.5 s for the Panda and 1.2 s for the Robotiq, with earlier termination when the respective contact condition is met. The final pose is measured after a 1 s settling period. These waits belong to the executed clock and are retained in the rollout diagnostics. The human-video arm configuration uses its separate 300 Hz contact executor and also includes 1 s of terminal settling.

    The rigid benchmark uses frictional holding with configuration-specific contact approximations. For an initially held object, it permits a surface-alignment correction of at most 3 cm before closure. Applicable furniture insertion tasks use an evaluator-owned retention approximation: a released part satisfying the predefined seating geometry is fixed at its attained pose to represent a threaded or tight fit. Ordinary DROID placement disables this retention rule. These operations are fixed properties of the evaluator, not additional actions available to the agent. In DROID, the gripper closes kinematically before first contact and then uses force-limited control; its closing rate is derived evaluator-side from the source gripper trace, with a minimum increment of 0.02 rad per control update. This trace is not exposed during reconstruction.

    The public local replay tool supports candidate-only inspection using the agent's estimated robot and camera geometry. It is distinct from the registered benchmark evaluator: its 5 cm inverse-kinematics rejection threshold and 180 s action limit are local replay checks, not universal benchmark admission thresholds. Local replay success therefore does not certify the benchmark outcome.

    \paragraph{Synchronized dexterous control.}
    Each Wuji hand has six wrist degrees of freedom and twenty finger joints. The submission provides a palm-pose stream and a finger-angle stream for each side, with equal lengths and a shared timestep that is a positive integer multiple of $1/300$ s; the task brief gives $1/30$ s as an example. An unused hand still receives a stationary stream. Finger angles are ordered by thumb, index, middle, ring, and little finger, with four joints per digit, and must satisfy the public robot limits. Palm $+z$ points along the extended fingers and $+x$ toward the palm's curling direction; the thumb lies on $+y$ for the right hand and $-y$ for the left.

    Both hands evolve under force-limited dynamics at 300 Hz. Wrist forces are capped at 200 N per translational axis, and wrist torques at 100 N,m per rotational axis; finger effort and velocity limits follow the robot description. The default initialization uses an open posture and a palm position displaced backwards by 8 cm, followed by a shared 1.2 s approach. A submitted initialization offset may have magnitude at most 25 cm. Both hands then follow the same action clock, without independent adaptive waits, and the rollout ends with 1 s of settling. Objects are held through contact and friction. Commanded palm motion is not exact object motion, and contact from either hand is considered when determining release. The hand validity checks allow at most 6 mm of robot penetration, 15 mm of object displacement during initialization, 4 mm of nonadjacent self-overlap diameter, and 0.001 rad of finger-joint limit excess. Joint-speed exceedances are recorded as diagnostics; they do not independently invalidate the rollout.

    \paragraph{Native joint control.}
    RoboDojo accepts the left-arm angles, left gripper, right-arm angles, and right gripper, in that order. Dual X5 has six arm joints per side (14 values total); dual xArm7 has seven (16 values total). The interval is 0.04 s and grippers use zero for closed and one for open. The household interface instead orders seven left-arm angles, seven right-arm angles, waist lift, waist pitch, head yaw, head pitch, and two gripper commands (20 values total), at an interval of 0.05 s. Waist lift is a distance; the other articulated coordinates are angles. Here grippers use zero for open and one for closed. Only the outer gripper joints are driven; mechanical follower joints retain their native coupling. RoboDojo integrates physics at 250 Hz (ten steps per command) with reset seed seven. The 35 rigid household cases integrate at 60 Hz (three steps per command), while the two wallet cases use 120 Hz (six steps per command). The household executors append a task-specific stationary terminal tail of either two or forty command intervals (0.1 s or 2 s); the wallet cases use 2 s. One initialization physics step precedes the recorded household trajectory.

    In the current household brief, the first-frame camera pose must be consistent with the public head-camera mount and the agent's inferred initial waist and head configuration. If ${}^{W}\!T_{H}$ is the resulting USD camera transform expressed using column vectors, then
    \begin{equation}
        {}^{W}\!T_C = {}^{W}\!T_H\,\operatorname{diag}(1,-1,-1,1).
        \label{eq:appendix_camera_fk}
    \end{equation}
    USD row-vector matrices are transposed before this conversion. The scene remains in the first camera frame even when the head subsequently moves. This requirement is not a claim about the prompts used for earlier submissions. Structural validation checks the submitted rigid transform and joint limits; it does not independently certify this forward-kinematics consistency.

    \paragraph{Articulated-gripper MuJoCo control.}
    The rope and soft-toy configurations use the supplied xArm7 collision model, including six mechanically linked gripper joints, five coupling equalities, and four exclusions between adjacent links. The agent estimates the robot base placement but preserves these internal mechanics. The TCP is 0.172 m from the final arm link. Each action contains seven arm angles and the gripper drive angle: zero is open and 0.85 rad is closed. The drive command changes at no more than 2 rad/s. The public brief specifies a physics timestep of $50\,\mu$s and a coupling constraint time constant of $0.1$ ms with damping ratio one. The runtime rejects a coupling residual above 0.001 rad at any physics step. The declared control interval must lie between 1 ms and 1 s; submissions contain at most 20,000 rows and at most 180 s of controls. Executed states are sampled at 20 Hz. Manipulated objects are passive rigid bodies, articulated chains, or deformable elements; controls actuate only the supplied robot.

    \paragraph{Runtime versions.}
    The inspected rigid execution environment uses SAPIEN 3.0.3 and ManiSkill 3.0.1 with Python 3.11.7. The isolated rope and soft-toy environment uses MuJoCo 3.11.0 with Python 3.11.7. The native robot environment uses Isaac Sim 5.1.0.0. These are backend-specific environments, rather than a single shared installation. Robot assets, controller implementations, native runtime dependencies, and the selected evaluation version are bound to the run's reproduction record; library dependency lower bounds do not identify a reproducible runtime.

    \subsection{Submission, Validation, and Execution}
    \label{app:submission_protocol}

    \paragraph{Submission contents.}
    A submission is a self-contained scene-and-control package. It declares the assigned embodiment, object identities and roles, geometry and collision geometry, physical properties, initial poses, camera model, control convention, and timestep. All referenced candidate geometry must resolve to local package assets. A reconstruction report states the observations supporting metric estimates, modeling approximations, and remaining ambiguities. Annotated source frames provide evidence for estimates such as object dimensions and contact timing. Table~\ref{tab:submission_artifacts} gives the concrete interchange files; these filenames are serialization conventions, not a prescribed reconstruction method.

    \begin{table}[!htbp]
        \centering\small
        \caption{\textbf{Submission artifacts.} $T$ denotes source-video frames and $N$ submitted control rows; these lengths need not agree.}
        \label{tab:submission_artifacts}
        \renewcommand{\arraystretch}{1.14}
        \begin{tabularx}{\linewidth}{@{}p{0.29\linewidth}X@{}}
            \toprule
            Artifact & Required content \\
            \midrule
            \texttt{protocol.json} & Version 3.0 manifest: target configuration, scene or scene-file reference, camera, robot, action-file references, timestep, and provenance. \\
            \texttt{actions.npy} & Finite control matrix with the embodiment-specific width in Table~\ref{tab:simulation_configurations}; Cartesian streams are stored as 64-bit floating-point arrays. \\
            Additional hand arrays & For paired hands, a second $N\!\times\!7$ palm stream and two $N\!\times\!20$ finger-angle streams, with four distinct file references. \\
            \texttt{scene.json} / \texttt{scene.xml} & Native Isaac submissions use a separate scene description; MuJoCo submissions use an MJCF scene with the supplied robot and explicit object, robot, and target roles. \\
            \texttt{assets/} & Locally referenced visual and collision geometry. Native mesh submissions provide vertices and face indices, with metric scale incorporated into the vertices. \\
            \texttt{expected/}\newline\texttt{obj\_poses.npy} & Camera-frame $T\!\times\!7$ object-pose estimates for the camera-frame rigid and human-video profiles. These are declarations, not commands; native joint and MuJoCo interfaces do not require them. \\
            \texttt{source/}, \texttt{report.md} & Input video, annotated evidence frames, and reconstruction rationale with uncertainty and approximation disclosures. \\
            \bottomrule
        \end{tabularx}
    \end{table}

    \paragraph{Scene semantics.}
    Manipulated objects are dynamic; receiving objects and supports are represented explicitly. Static and dynamic geometry are distinguished, and a receptacle's collision model must preserve its cavity. The shared rigid schema specifies primitive half-extents, whereas the native Isaac scene format specifies full extents; the conversion must not introduce a factor-of-two scale error. Object names identify semantic roles in the rigid interfaces. Native tasks use the supplied semantic identifiers, while MuJoCo submissions identify disjoint robot, manipulated-object, and target subtrees. Doors and drawers require passive revolute or prismatic joints, respectively, with the joint axis and anchor expressed in the parent object's local frame. Initial joint displacement is zero; limits are in radians for revolute joints and metres for prismatic joints.

    \paragraph{Structural validation.}
    The validator checks that required files exist and that geometry, poses, and action arrays satisfy the assigned contract. Configuration-specific checks include object and receiver presence, finite arrays, synchronized hand streams, dimensions and command rates, admissible joint limits, local asset references, and valid rigid transforms. Native scene checks additionally verify positive dimensions and mass, normalized quaternions, collision participation, and consistency between declared sizes and mesh bounds. A structurally valid package may still fail to initialize, contain unreachable controls, or fail the task; validation is not a success oracle.

    \paragraph{Physical execution.}
    Evaluation initializes the scene from the submitted package and executes its own controls through the registered robot interface. The benchmark owns simulation stepping and state recording. Candidate object-pose predictions do not drive bodies, and candidate instructions cannot replace the measured rollout with a declared trajectory. Direct object animation, object actuators, artificial grasp attachments, and post-initialization object-state overrides are outside the candidate control interface. Passive mechanisms required by an object's physical structure, such as a door hinge or rope joints, are part of the scene model. Native executors record actual robot and object states and, where supported, contacts, constraint residuals, and physical-validity diagnostics. Candidate-only inspection uses the declared geometry and camera where supported; evaluation renders follow the configuration-specific camera and registration policy described above, including native mounted cameras.

    \paragraph{Build and outcome separation.}
    A missing or invalid candidate package is a method failure. Successful parsing, a completed simulator process, and the agent's declared completion status do not imply Task Success. Task judgments use the executed states and the benchmark's evaluation specification. Conversely, an execution that completes and produces valid records can fail the task while remaining evaluable. Runtime or dependency failures attributable to evaluation infrastructure are recorded separately and require a valid rerun before a task conclusion is drawn. The evaluator's recorded failure category distinguishes these cases; no hidden score is used to choose a better intermediate submission.

    \subsection{Agent Tools and Public Operations}
    \label{app:interface_tools}

    Coding tools provide the agent's interaction with the construction environment. They support video inspection, numerical analysis, geometric modeling, and writing scene and control artifacts. Table~\ref{tab:coding_tools} summarizes these capabilities and the feedback available to the agent. The reconstruction procedure is unrestricted within the task's observation and resource constraints. OpenCode selects the editing interface for the configured model. Fable exposes its native shell, read, write, and edit tools, while Astra uses the coding tools of its pinned Codex CLI. Exact invocation schemas accompany the reproducibility material.

    \begin{table}[!htbp]
        \centering\small
        \caption{\textbf{Coding capabilities available to the agent.} All operations are restricted to the permitted workspace and public resources. Model-specific tool definitions are supplied by the pinned agent framework.}
        \label{tab:coding_tools}
        \setlength{\tabcolsep}{4pt}\renewcommand{\arraystretch}{1.14}
        \begin{tabularx}{\linewidth}{@{}>{\raggedright\arraybackslash}p{0.23\linewidth}>{\raggedright\arraybackslash}p{0.31\linewidth}>{\raggedright\arraybackslash}X@{}}
        \toprule
        Capability & Agent-specified input & Agent-visible feedback \\
        \midrule
        Shell execution & Program or command; optional execution limit & Standard output, execution errors, exit status, and timeout information. \\
        File and image reading & File or directory; optional text range & File contents, directory entries, or an image observation. \\
        Filename search & Filename pattern and search location & Matching file paths. \\
        Text search & Text pattern and search scope & Matching passages and file locations. \\
        File creation & File location and contents & Confirmation, diagnostics, or a write error. \\
        Text replacement & File and requested replacement & Modification result or an error. \\
        Patch editing & Additions, changes, or deletions & Results of the requested file modifications. \\
        Task bookkeeping & Task descriptions and completion states & Updated local task list. \\
        \bottomrule
        \end{tabularx}
    \end{table}

    The shell tool defaults to a 120 s timeout, which may be overridden for individual commands. Text output is ordinarily limited to 2,000 lines or 51,200 bytes; oversized shell output is retained in a local file that the agent can subsequently read. Text reads default to 2,000 lines, with a 51,200-byte limit and a 2,000-character limit per line. Filename and text searches return at most 100 results. These observation limits are distinct from the episode's total construction allowance.

    The benchmark provides operations for observing the video, inspecting physical execution, and managing candidate reconstructions (Table~\ref{tab:public_operations}). Frame extraction exposes visual evidence; physical replay provides feedback on the reconstructed interaction; validation checks structural requirements; checkpointing preserves a candidate during refinement; and submission requests final delivery. None of these operations returns reference annotations or benchmark scores. A checkpoint or submission is validated on a protected copy, so subsequent edits to the working candidate cannot change a previously accepted reconstruction.

    \begin{table}[!htbp]
        \centering\small
        \caption{\textbf{Benchmark operations during construction.} The interface separates visual and physical observations, structural feedback, and candidate delivery. An accepted submission terminates generation; successful validation alone does not.}
        \label{tab:public_operations}
        \setlength{\tabcolsep}{4pt}\renewcommand{\arraystretch}{1.17}
        \begin{tabularx}{\linewidth}{@{}>{\raggedright\arraybackslash}p{0.17\linewidth}>{\raggedright\arraybackslash}p{0.21\linewidth}>{\raggedright\arraybackslash}p{0.27\linewidth}>{\raggedright\arraybackslash}X@{}}
        \toprule
        Operation & Input & Agent-visible feedback & Effect on the episode \\
        \midrule
        Frame extraction & Source video and sampling specification & Frame references, timestamps, and image metadata & Makes frames available for inspection; the candidate is unchanged. \\
        Physical replay & Candidate scene and robot controls & Rendered rollout and available execution diagnostics & Supports inspection and revision of the reconstruction; does not submit or score it. \\
        Validation & Candidate reconstruction and task configuration & Missing artifacts and violations of the public submission requirements & Construction continues; no candidate is submitted. \\
        Checkpointing & Candidate reconstruction & Validation feedback and confirmation of preservation & A valid candidate is retained as a fallback during the permitted refinement. \\
        Submission & Candidate reconstruction & Acceptance or validation errors & Acceptance freezes the final candidate and ends generation; an invalid submission consumes one opportunity. \\
        \bottomrule
        \end{tabularx}
    \end{table}

    The agent is informed whether preservation or submission succeeded, whether an error arose from the candidate or the validation service, and how many submission opportunities remain. Only a complete, internally consistent copy can become an accepted candidate. Validation establishes compliance with the public submission requirements; it neither determines physical task success nor changes the candidate selected for evaluation.


    \subsection{Construction Budget and Feedback}
    \label{app:interface_budget}

    \paragraph{Physical feedback during construction.}
    All benchmark configurations provide access to physical rollout feedback, including rendered observations and execution diagnostics. Agents may inspect their reconstructed interactions alongside the source video and revise both the environment and robot behavior within the construction budget. Feedback is restricted to the agent’s own reconstruction; reference annotations and benchmark scores remain unavailable. Final evaluation executes the selected, frozen reconstruction without further agent revision.

    Construction feedback is restricted to the candidate's own scene, controls, rendered observations, and available execution diagnostics. It does not expose reference geometry, source calibration, reference controls, or hidden evaluation scores. A rollout becomes visual feedback when its rendered observations are supplied to or opened by the agent; merely creating a video file is insufficient. The agent compares these observations with the permitted source video. This construction loop is distinct from final evaluation, which executes the selected, frozen package and computes the benchmark outcomes without further agent revision. Candidate replay and final evaluation retain their respective camera and registration conventions described in Section~\ref{app:simulation_environment}.

    \paragraph{Budget semantics.}
    Construction is bounded by the configured time and, where applicable, API-cost allowance. There is no fixed episode-wide tool-call or model-turn budget. Coding, validation, and rollout inspection take place within construction; the separate evaluation allowance is used after the candidate is frozen. Submission opportunities constrain delivery attempts and do not limit the number of analysis, editing, or inspection operations. Likewise, a per-command timeout or per-response output limit is not an episode-wide tool-call budget. The model-specific stopping and candidate-selection rules below determine which package is retained when construction ends.

    \paragraph{Scope of the construction protocols.}
    The following three paragraphs document the legacy reference two-round authoring driver. The multi-model experiments additionally use a managed executor with explicit submissions and protected checkpoints, and externally launched native coding agents. Their respective stopping and delivery rules are specified below; the two-round default is not a model-turn limit for those experiments.


    The reference authoring driver allows at most two construction rounds: the initial task brief and, if necessary, one repair round containing the concrete validation errors. Both rounds belong to one attempt; the default adapter resumes the same agent session for the second round. Validation is performed after each round, and construction stops at the first validation-passing submission or after the second round. The package left by this process is handed to evaluation, including an invalid final package if repair fails. A run producing no package is retained as a failed build. A fresh experimental attempt must use a new run location, because the outer runner treats an existing submission as completed work.

    The configured outer limits are 10,800 s for construction and 5,400 s for evaluation, with timeout status retained. These limits bound the respective runner processes; there is no additional fixed token budget or tool-call count in the standard interface. The default author adapter requests Claude Opus 5 and exposes the the tool capabilities in Table~\ref{tab:coding_tools}; the benchmark also permits an explicitly configured external author command. The requested model, adapter, command, and effective limits therefore belong to each run record. These defaults describe the current interface and do not retroactively establish the settings of every historical result.

    Between rounds, feedback concerns the submitted package's validity. During a round, the agent can use available candidate-only simulation and inspection tools to refine its reconstruction. Such outputs describe the consequences of its own scene and controls. Reference-based geometric errors, trajectory errors, Task Progress, and Task Success are computed after construction and are not returned as repair feedback.

    \paragraph{Managed construction protocol.}
    The multi-model executor uses explicit submission rather than the two-round return-and-validate procedure above. Each episode receives one generation attempt, a cumulative construction allowance of 10,800 s, and a \$60 observed API-cost threshold. Delayed usage reporting may cause the billed cost to exceed the threshold. There is no fixed total number of model turns or tool calls. A normal agent return without delivery can resume the same conversation and workspace; three cumulative empty turns, defined as turns with neither a tool call nor a candidate, terminate the attempt. Resource exhaustion and repeated interaction without progress may also stop construction. A terminal method failure does not automatically trigger a fresh generation attempt.

    \begin{table}[!htbp]
        \centering\small
        \caption{\textbf{Construction schedule for benchmark-managed episodes.} Times are cumulative within one attempt. The first milestone is a prompt instruction; the generation and publication deadlines are enforced by the executor.}
        \label{tab:delivery_clock}
        \setlength{\tabcolsep}{4pt}\renewcommand{\arraystretch}{1.14}
        \begin{tabularx}{\linewidth}{@{}>{\raggedright\arraybackslash}p{0.18\linewidth}>{\raggedright\arraybackslash}X@{}}
        \toprule
        Elapsed time & Construction and delivery policy \\
        \midrule
        About 30 min & Complete an initial evidence-based candidate; validate it and repair missing or malformed artifacts. \\
        120 min & Restrict remaining work to completing and repairing delivery. \\
        150 min & Close the optional refinement allowance. \\
        165 min & Stop generation and new delivery requests; select the latest valid protected checkpoint if no final submission has been accepted. \\
        180 min & Complete process termination, integrity verification, and publication of the selected candidate. \\
        \bottomrule
        \end{tabularx}
    \end{table}

    \paragraph{Submission allowance and candidate selection.}
    After the first successful public validation, the agent may submit immediately or save a checkpoint before one bounded refinement. Up to three submission attempts are permitted. An invalid package or invalid submission path consumes an attempt; a failure of the executor's validation service does not. Saving a checkpoint neither consumes nor replenishes submission opportunities. Once a valid checkpoint exists, only one optional refinement checkpoint is permitted, before the 150-minute cutoff.

    A successful explicit submission ends generation and determines the package evaluated, without requiring a final natural-language response. If generation reaches the 165-minute deadline, the latest valid executor-saved checkpoint is selected. An invalid revision cannot replace that checkpoint. Unsaved workspace files and claims in an agent response are not eligible substitutes. The selected package is frozen and verified before evaluation, and hidden scores never determine which candidate is chosen. An episode without a valid delivery, or with an explicit failure or infeasibility declaration, is retained as a method failure. Process-cleanup and package-integrity failures are classified separately as executor faults.

    \paragraph{Freezing the reconstruction.}
    The selected candidate fixes both $\eta$ and $\pi$: scene assets, physical parameters, initial configuration, camera estimates, controls, and their timing are preserved together. Subsequent workspace edits and construction rollouts cannot alter that candidate. Evaluation and any infrastructure rerun use the same package under the specified evaluator. A timeout without an eligible submission or protected checkpoint yields a build failure; it does not create a candidate from unfinished workspace files. Candidate selection uses the delivery protocol, without comparing hidden scores across intermediate reconstructions.

    \paragraph{Timeouts and infrastructure failures.}
    Model-service connection failures receive up to three request retries. Persistent API failures enter a recovery queue with a 600 s cooldown and at most three additional retests; the conversation, workspace, and cumulative construction budgets are preserved. Episodes that exhaust service recovery are held for inspection rather than assigned a task outcome. A validator infrastructure failure does not consume a submission or refinement opportunity.

    Evaluation has a separate cumulative allowance of 5,400 s for reference execution, candidate simulation, and metric computation. A pre-evaluation GPU probe has a separate 90 s timeout. A failed probe requeues the unchanged package without consuming its evaluation allowance and removes that GPU from admission; consecutive failures on two distinct GPUs pause the evaluation pool. Evaluator timeouts, unavailable dependencies, missing outputs, and corrupt execution records are recorded as infrastructure failures. They require a valid evaluation of the same frozen candidate before a task outcome is assigned. An infrastructure rerun does not regenerate the reconstruction or select a different workspace version. Candidate-caused invalidity and a physically executed failure remain method outcomes.


    \paragraph{Reproduction record.}
    Reproducing an interface run requires the source-video identity, assigned configuration, fully substituted prompt and public handout, model and adapter invocation, effective budgets, submitted package, construction transcript, and physical execution record. The construction record also identifies the candidate versions replayed, the rendered observations actually supplied to the agent, and the feedback preceding each revision. Versioned source and asset hashes identify the robot, simulator adapters, validation rules, and evaluation snapshot; recorded timesteps, initialization, seeds, and state timestamps identify the executed dynamics. The prompt supplement records the source snapshot used to prepare this section. Because simulator state, rendering, and solver behavior depend on the runtime, matching an interface specification alone is not evidence of identical trajectories across unverified installations.

    \subsection{Agent Initialization and Model Configurations}
    \label{app:interface_models}

    The experiment configuration identifies the model, agent framework, provider route, construction budget, public interface specification, and evaluation version. Seven model configurations use OpenCode 1.18.29 under the managed construction protocol. GPT-6 Astra uses Codex CLI 0.154.0-alpha.6.2, and Fable-5.1 uses Claude Code 2.1.276. Table~\ref{tab:agent_models} gives the exact model identifiers and inference settings. Provider fallback is disabled for the OpenCode configurations, and their per-response output limit is 32,000 tokens including reasoning. The native coding agents retain the output-length behavior of their pinned CLIs.

    \begin{table}[!htbp]
        \centering\small
        \caption{\textbf{Model configurations.} The first seven rows use OpenCode and the model-identifier prefix \texttt{openrouter/}. A dash denotes an unspecified reasoning-effort setting.}
        \label{tab:agent_models}
        \setlength{\tabcolsep}{3pt}\renewcommand{\arraystretch}{1.14}
        \begin{tabularx}{\linewidth}{@{}>{\raggedright\arraybackslash}p{0.18\linewidth}>{\raggedright\arraybackslash}p{0.40\linewidth}>{\raggedright\arraybackslash}p{0.22\linewidth}>{\raggedright\arraybackslash}X@{}}
        \toprule
        Model & Model identifier & Provider & Effort \\
        \midrule
        DeepSeek V4.1 Flash & \path{deepseek/deepseek-v4.1-flash} & \path{novita/fp8} & -- \\
        GLM-5.3 Flash & \path{z-ai/glm-5.3-flash} & \path{novita/fp8} & -- \\
        Kimi K3 & \path{moonshotai/kimi-k3} & \path{moonshotai/mxfp4} & -- \\
        GPT-5.6 & \path{openai/gpt-5.6-sol} & \path{openai} & -- \\
        Gemini 3.8 Flash & \path{google/gemini-3.8-flash} & \path{google-vertex/}\newline\path{global} & High \\
        Claude Opus 5 & \path{anthropic/claude-opus-5} & \path{anthropic} & -- \\
        Qwen3.8 Max & \path{qwen/qwen3.8-max-0902} & \path{alibaba} & -- \\
        \midrule
        GPT-6 Astra & \path{gpt-6-astra} & Codex & Ultra \\
        Fable-5.1 & \path{claude-fable-5-1} & Claude Code & Max \\
        \bottomrule
        \end{tabularx}
        \vspace{3pt}
    \end{table}

    \paragraph{Initialization of a single-video trial.}
    A trial is instantiated from the selected video, target embodiment, public interface specification, and model configuration. Before the first model call, the benchmark stages the permitted inputs, initializes empty analysis and output directories, and supplies the fully substituted task brief to a fresh agent session. The same conversation and workspace persist throughout construction. The selected candidate is subsequently transferred to the separate evaluation process, which has access to the hidden annotations. Reproductions use the same input and configuration versions and begin from a new session. Executable launch procedures and machine-specific setup are provided with the reproducibility material.

    \paragraph{Native coding-agent sessions.}
    Astra and Fable operate on exported public task packages containing the video, released resources, and instantiated brief. Each task package binds the source video, target embodiment, released resources, and prompt revision. Their external supervisors provide filesystem restrictions, API access, time accounting, and a local submission service. The native model is launched with the settings in Table~\ref{tab:agent_models}; framework-specific runtime settings accompany the experimental configurations. The completed package, submission receipt, session record, and transcript are imported into the benchmark evaluation queue.

    These sessions use a 180-minute hard limit. The selected result is the last successful submission, or a preserved checkpoint on timeout. Their local submission service does not by itself enforce the managed executor's complete staged-deadline and refinement policy. Accordingly, the 165-minute generation cutoff and executor-managed counters in Table~\ref{tab:delivery_clock} apply to the OpenCode runs. Agent-framework differences and Astra's per-instance reasoning settings remain part of the reported experimental configuration. The model configurations are associated with their own instance manifests and prompt revisions; the current interface inventory is not substituted for those experimental manifests.


    \subsection{Task Prompts}
\label{app:interface_prompts}

The task brief combines configuration-specific instructions, the applicable submission contract, and episode-specific information, including the video location, image dimensions, frame count, output location, and public resources. The panels below summarize these instructions in publication notation, using mathematical symbols and placeholders for episode-specific values. The submission schema is specified in Table~\ref{tab:submission_artifacts}. Panels labeled verbatim preserve the recorded wording apart from the indicated substitutions.


\paragraph{Construction feedback and session control.}
All evaluated model configurations provide access to candidate-only physical replay, rendered observations, and execution diagnostics. The shared construction instruction below directs agents to inspect their simulated interactions alongside the source video and revise the reconstruction as needed within the construction budget. Reference annotations and benchmark scores remain unavailable during construction.

For benchmark-managed OpenCode runs, candidate delivery requires explicit submission or selection of a protected checkpoint under the prescribed stopping rules; returning from an agent turn does not itself deliver a candidate. The validation-repair prompt applies to the two-round reference authoring driver. The managed-session continuation messages and executor-status template apply to the benchmark-managed executor.


    \begin{interfaceprompt}{Shared construction instruction: physical rollout feedback}
    Execute the reconstructed scene with your proposed robot controls and inspect the rendered physical rollout alongside the source video. Use the observed interaction and execution diagnostics to revise the scene and controls as needed within the construction budget. If execution fails, use the reported error to repair the candidate and replay it. Package validation alone does not replace physical execution and visual inspection. Feedback concerns your own reconstruction; reference annotations and benchmark scores are unavailable. Submit the completed candidate through the assigned delivery interface. The selected package is fixed for subsequent evaluation.
    \end{interfaceprompt}

    \begin{interfaceprompt}{Reconstruction brief: rigid scene and Cartesian control}
    You are given one RGB video $V$ with resolution $W\times H$ and $T$ frames, and the public specification of the assigned robot. Reconstruct the manipulated object, the receiving object or interacting surface, and the surrounding geometry needed to reproduce the observed task. Infer the task from the video. Source robot logs, depth, camera calibration, reference object geometry, and reference trajectories are not available.

    Estimate object shape, size, initial pose, and camera intrinsics from the images. Preserve hollows and openings that participate in contact. Represent static objects and supports separately from manipulated objects. Express geometry and motion in the prescribed camera frame; for a moving camera, use the first frame as the fixed coordinate system.

    Supply your own absolute TCP poses and gripper commands at 5 Hz. Positions are in metres, rotations follow $R_zR_yR_x$, and one denotes open while zero denotes closed. The first row specifies the initial TCP pose. Re-plan the motion for the assigned embodiment while preserving the demonstrated object interaction and outcome. Keep successive displacements within approximately 3 cm and rotations within 0.3 rad, and finish at rest.

    Deliver the scene, camera estimate, controls, object-motion estimates at the video timestamps, local geometry assets, annotated evidence frames, and a report of observations, assumptions, and uncertainty. Check that the grasp and release commands agree with the reconstructed object and goal locations. Run the assigned package validator and repair any reported errors. Write only in the designated output and temporary directories. Do not obtain the source dataset's ground truth or calibration.
    \end{interfaceprompt}

    \begin{interfaceprompt}[breakable=false]{Embodiment supplement: human video to paired hands}
    Reconstruct the manipulated object, visible support, and relevant receiving objects, and perform the interaction with a left and a right Wuji hand. Both hands are available even if only one human hand is visible. An unused hand must receive stationary commands.

    Supply synchronized palm poses and twenty finger angles for each hand. Both sides share one timestep, a positive multiple of $1/300$ s. Use the public palm axes, finger ordering, joint limits, and force-limited control specification. Finger angles determine posture; the palm stream's gripper indicator is informational. Express all poses and object-motion estimates in the first camera frame.

    Model a drawer with a passive prismatic joint and a door with a passive revolute joint. Preserve collision clearances and cavities. Do not attach objects to robot links or prescribe object motion during simulation. Include both hands in release reasoning and report uncertainty in the proposed task completion.
    \end{interfaceprompt}

    \begin{interfaceprompt}{Reconstruction brief: native dual-arm environments}
    You are given an RGB video and the public specifications of the assigned simulator and robot. Infer the demonstrated task, participating objects, and their roles from the video. Reconstruct the visible geometry, supports, scale, and initial poses, and provide your own joint commands. Make manipulated objects dynamic and preserve physical collisions. Use only the staged inputs, public resources, and feedback from your own reconstruction.

    Express scene poses in the OpenCV camera frame and provide a rigid transform from that frame to the public robot world. Joint commands are expressed in the robot's named joint frames. For RoboDojo, estimate the full optical pose and focal length; supply the assigned 14- or 16-value controls at 25 Hz, with zero meaning closed and one open. For the household robot, supply the 20-value arm, waist, head, and gripper controls at 20 Hz, with zero meaning open and one closed. Derive the first-frame camera transform from your estimated initial waist/head configuration and the public camera mount, as in Eq.~\ref{eq:appendix_camera_fk}.

    Deliver the manifest, scene description, finite action matrix, local geometry assets, and reconstruction report. Object motion must result from robot actuation and simulation. An expected object trajectory is neither required nor executed. Validate the package against the assigned submission requirements. Do not access hidden source layouts, source camera calibration, reference controls, or reference execution records.
    \end{interfaceprompt}

    \begin{interfaceprompt}{Reconstruction brief: rope and soft-toy environments}
    Reconstruct the video as a native MuJoCo scene using the supplied xArm7 robot and articulated gripper. Read the public robot description and preserve its internal body geometry, actuators, mechanical couplings, and adjacent-link collision exclusions. Estimate the base placement from the first camera frame.

    Model the manipulated object with passive bodies, an articulated chain, or deformable elements, and model the receiver with physical colliding surfaces. Set gravity consistently with the first camera frame. Supply eight joint controls in radians, with gripper angle zero for open and 0.85 for closed. Use the specified $50\,\mu$s physics timestep and preserve the gripper coupling and command-rate limits.

    Identify the robot, manipulated object, and receiving structure as disjoint physical roles. For rope routing, preserve the connected rope and the open channel between the receiver walls. Use no object actuators, artificial attachments, motion-capture drivers, or object-state overrides. Include the video, scene, controls, assets, and a report of uncertainties. Run structural validation and physical replay. A completed replay is not a benchmark task judgment. Read no hidden sample metadata, calibration, or reference trajectory.
    \end{interfaceprompt}

    \begin{interfaceprompt}{Validation-repair prompt (verbatim apart from substitutions)}
    Your package at \textit{[output location]} does not validate:\par
    -- \textit{[validation errors]}\par
    Fix it (same rules apply) and re-run the validate command.
    \end{interfaceprompt}

    \begin{interfaceprompt}{Managed-session continuation: no tool call or candidate (verbatim)}
    The previous turn ended without a tool call or delivery. Continue the task from the existing workspace and conversation; inspect existing artifacts, do not repeat completed steps; when a package passes public validate, call \texttt{`python public/v2\_tools.py submit <pkg\_dir>`}.
    \end{interfaceprompt}

    \begin{interfaceprompt}{Managed-session continuation: candidate or checkpoint present (verbatim)}
    No explicit submission received. Call python public/v2\_tools.py submit <pkg\_dir> to deliver within your remaining allowance.
    \end{interfaceprompt}

    The executor also appends a status message to the latest tool observation, or supplies it as a system prefix before the first tool result. The message reports elapsed time, remaining generation time, construction phase, the existence of a valid checkpoint, remaining submission and refinement opportunities, and media usage. The following template preserves the field order and fixed wording; angle-bracket fields are populated from the episode state.

    \noindent\begin{minipage}{\linewidth}
\begin{lstlisting}[style=interfacecode]
[executor-status/1]
{"revision":"executor-status/1","elapsed_s":<seconds>,"remaining_s":<seconds>,"stage":<phase>,"valid_checkpoint":<boolean>,"submit_remaining":<count>,"refinement_remaining":<count>,"media_remaining":<media-status>,"removed_images":<count>}
Removed images are visible placeholders. Re-read an image to attach it again within the remaining media budget.
[/executor-status/1]
    \end{lstlisting}
\end{minipage}

Because agent frameworks differ in their internal tool use, context management, and candidate-selection behavior, comparisons in this paper should be interpreted as comparisons among complete coding-agent configurations rather than controlled estimates of isolated backbone-model capability.

\section{Benchmark Construction and Reference Annotations}
\label{app:benchmark_construction}
\raggedbottom

Benchmark construction establishes the correspondence between a source demonstration and the quantities evaluated in simulation: task-relevant geometry, object motion, and interaction outcomes. Each video is paired with a target robot and simulator, together with hidden annotations defining these quantities. Annotation provenance distinguishes source measurements, estimates fitted to observations, and reconstructed physical targets. The human-assisted reference (HAR) is constructed using these resources and evaluated against the same applicable targets as coding-agent submissions.

\subsection{Data Sources and Video Selection}
\label{app:data_sources}

We collect videos from FurnitureBench, DROID, RoboDojo, HOI4D, HOT3D, DexYCB, OakInk2, an in-house household collection, and reconstructed digital-twin demonstrations of planar pushing, rope routing, and soft-toy packing. These sources vary in demonstrator embodiment, viewpoint, object properties, and interaction structure. RoboDojo supplies simulation-generated demonstrations; the other groups include recorded real-world interactions. The digital-twin group contains three planar-pushing, five rope-routing, and two soft-toy clips, using real videos with reconstructed physical targets.


The benchmark contains 189 distinct source video clips and 222 reconstruction instances across 39 task families (Table~\ref{tab:benchmark_statistics}). Each instance pairs a source video with a target simulator, robot embodiment, and control interface. The 33 human-demonstration clips each have an arm and a bimanual-hand configuration. A clip is a selected temporal window, not necessarily an independent acquisition episode: the 64 furniture clips come from 46 original episodes, and some human clips share a longer recording. All 222 instances are included in the main evaluation.

\begin{table}[!htbp]
    \centering\small
    \caption{\textbf{Composition of the construction inventory.} Clips count distinct staged video windows; instances count clip--configuration pairs. Embodiments include the end effector. Shared recordings and reuse across embodiments are retained in the provenance.}
    \label{tab:benchmark_statistics}
    \setlength{\tabcolsep}{4pt}\renewcommand{\arraystretch}{1.12}
    \begin{tabularx}{\linewidth}{@{}lrr>{\raggedright\arraybackslash}X@{}}
    \toprule
    Video source & Clips & Instances & Target embodiment \\
    \midrule
    FurnitureBench & 64 & 64 & Panda \\
    DROID & 20 & 20 & Panda/Robotiq (18); cloth gripper (2) \\
    RoboDojo & 25 & 25 & Dual X5 or dual xArm7 \\
    HOI4D & 9 & 18 & Panda; paired Wuji hands \\
    HOT3D & 7 & 14 & Panda; paired Wuji hands \\
    DexYCB & 11 & 22 & Panda; paired Wuji hands \\
    OakInk2 & 6 & 12 & Panda; paired Wuji hands \\
    In-house household & 37 & 37 & G1 with Omnipicker grippers \\
    Reconstructed twins & 10 & 10 & xArm7 with pusher or articulated gripper \\
    \midrule
    Total & 189 & 222 & Source-specific configurations above \\
    \bottomrule
    \end{tabularx}
\end{table}

\paragraph{Selection and coverage.}
Selection retains the initial configuration, the task-defining interaction, and the available terminal context. A visible task label alone is insufficient: the clip is reviewed to establish what interaction actually occurs, including whether it begins with an object already held or ends before release. Furniture windows are capped at eight per assembly step and two per leg-insertion step, covering 25 assembly steps. The retained-window ordering uses reference execution outcome, then agreement with the official terminal assembly pose, and finally a deterministic sample identifier. This construction choice favors reconstructable windows within the cap and is part of the selection policy. Unsuccessful selected references remain in the inventory. Human-video and household admissions follow documented clip-level review; clip boundaries, source identifiers, and exclusions are retained.

\paragraph{Source-specific annotation procedures.}
Auxiliary source information supports hidden annotations and is unavailable as episode evidence to the evaluated coding agent. Table~\ref{tab:source_construction} distinguishes information supplied by each source from the annotation work needed to make it usable for evaluation. Robot assets remain public as described in Section~\ref{app:agent_sandbox}. Geometric and trajectory coverage is source-dependent: a partial visible surface or an estimated orientation axis is not treated as a complete object model or a fully observed rigid pose.

\begin{table}[!t]
    \centering\small
    \caption{\textbf{Evidence used to construct the hidden annotations.} The middle column identifies available source information; the last identifies the processing and modeling needed for benchmark evaluation. Reconstructed twins supply modeled targets.}
    \label{tab:source_construction}
    \setlength{\tabcolsep}{4pt}\renewcommand{\arraystretch}{1.14}
    \begin{tabularx}{\linewidth}{@{}>{\raggedright\arraybackslash}p{0.15\linewidth}>{\raggedright\arraybackslash}p{0.32\linewidth}>{\raggedright\arraybackslash}X@{}}
    \toprule
    Source & Available source evidence & Benchmark annotation work \\
    \midrule
    FurnitureBench & CAD, tags, calibrated cameras, and robot logs. & Register part poses; fit grasp offsets to bridge occlusions during verified holding. \\
    DROID & Calibrated RGB-D, multiple views, and robot kinematics. & Fit object geometry and poses; reconstruct supports and cavities; track visible cloth surfaces. \\
    RoboDojo & Native assets and synchronized simulation states. & Export geometry, cameras, and trajectories from the same execution as the video. \\
    HOI4D & Instance CAD, object boxes, object poses, and camera trajectories. & Compose object and camera poses in a gravity-aligned frame; model supporting surfaces. \\
    HOT3D & Object CAD and tracked object/camera poses. & Register trajectories; identify supports and static/dynamic roles. \\
    DexYCB & YCB meshes, calibrated views, and object/hand poses. & Convert pose units and frames; estimate support height from initial mesh bottoms. \\
    OakInk2 & Task programs, part scans, motion capture, and camera calibration. & Select primitive segments; infer articulation from relative part motion; construct collision approximations. \\
    Household & Registered assets, RGB-D observations, and source recordings. & Estimate geometry and poses or visible surfaces; retain view dependence and uncertain intervals. \\
    Reconstructed twins & Reconstructed geometry and rigid or particle trajectories. & Freeze the modeled target and its provenance; adapt the simulation to the target robot. \\
    \bottomrule
    \end{tabularx}
\end{table}

\subsection{Task Organization and Dataset Composition}
\label{app:task_organization}

The benchmark comprises 39 task families used as fixed evaluation
groups. Family assignments reflect source-specific task categories,
target embodiments, and annotation configurations, as listed in
Figure~\ref{tab:sandbox_resources}. A reconstruction instance pairs a video with a target simulator, robot embodiment, and control interface; family assignments provide the grouping for task-level reporting. The manifest links each instance to its source sequence, clip boundaries, participating entities, and annotation version, including shared recordings across windows and embodiments. Family-specific conditions are given in Section~\ref{app:functional_correctness}.

\begin{figure}[!t]
    \centering
    \includegraphics[width=\columnwidth]{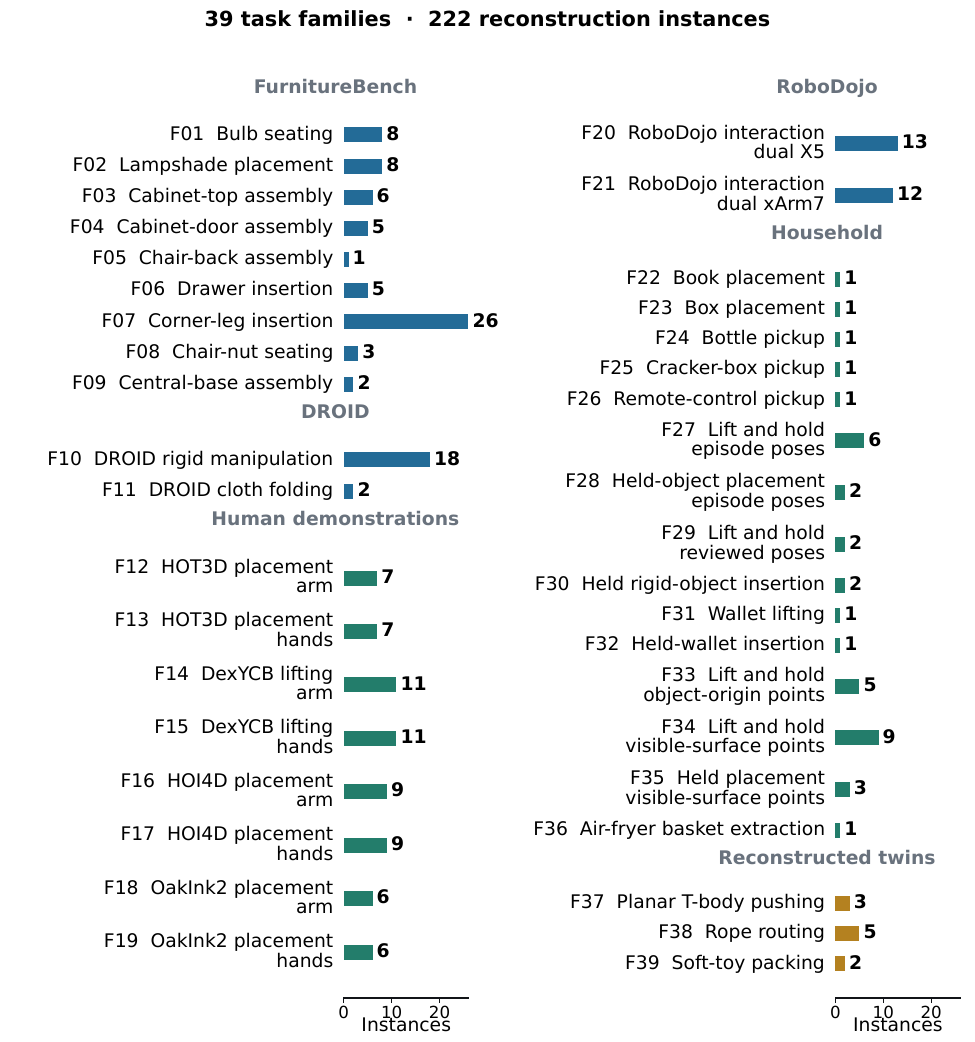}
    \caption{\textbf{Complete task-family inventory.} Each bar is one family in the fixed construction manifest, and its label gives the number of video--configuration pairs. The 39 counts sum to 222 instances. Arm and hand variants count separately; their underlying source clips are shared. Family identifiers F01--F39 index the accompanying catalogue. Counts describe the construction inventory rather than a model's successful builds or the metric-specific evaluable subset.}
    \label{fig:task_family_counts}
\end{figure}

\paragraph{Family inventory.}
Figure~\ref{fig:task_family_counts} lists every family and its number of reconstruction instances. The labels describe the interaction and retain the embodiment or annotation variant used in the fixed family assignments. F01--F09 cover furniture seating and assembly; F10--F11 cover DROID rigid manipulation and cloth folding; F12--F19 cover human-demonstrated placement or lifting under the two target embodiments. F20--F21 contain the native RoboDojo interactions for dual X5 and dual xArm7. These include pickup, tube and charger insertion, block and bowl stacking, tower construction, digit arrangement, category and size sorting, object exchange, and ordered covering/uncovering. Tower construction has one X5 instance; the other twelve native tasks have one instance per embodiment.

F22--F36 cover household placement, pickup with retained holding, initially held insertion, and basket extraction. Their labels distinguish the retained pose, object-origin, visible-surface, and wallet annotation variants. F37--F39 cover planar pushing, rope routing, and soft-toy packing. The accompanying family catalogue records the objective and instance membership for each label, preserving the manifest's assignments without merging families that share a broad verb such as ``place''. Metric-specific eligibility is specified in Section~\ref{app:aggregation}.

\subsection{Hidden Evaluation Specifications}
\label{app:hidden_specifications}

Each instance has a source-derived or explicitly reconstructed evaluation specification. The specification links persistent object identities to geometry, demonstrated motion, and task requirements. Provenance, uncertainty, and visibility masks identify the quantities and intervals that support evaluation. Human-assisted reference outputs are experimental submissions and do not replace source observations.

\paragraph{Entities and scene geometry.}
Entities include manipulated objects, receivers, supports, articulated structures, and relevant obstacles. Persistent identifiers link each entity's geometry, trajectory, and task annotations. The scene records metric dimensions, initial poses, and passive articulation where present. Evaluation geometry and simulator geometry are distinguished when collision decomposition, primitive fitting, or an articulation conversion changes the representation. A support or cavity inferred from incomplete observations retains that status; a visually plausible reconstruction is not relabeled as a direct measurement.

\paragraph{Demonstrated object motion.}
Rigid annotations contain positions and orientations; articulated annotations additionally record the relevant joint coordinate when available. Rope and soft-toy targets retain time-indexed particles or surfaces. Cloth uses camera-visible surface points from stereo reconstruction with object masks and robot-occlusion removal. Household point and wallet annotations retain their view and visibility conditions. Such surface records do not automatically establish material-point correspondence or a unique rigid rotation. Coordinate transforms, metres, seconds, scalar-first quaternions, source frame indices, and validity masks are recorded explicitly.

Interpolation and occlusion handling follow the source procedure rather than a universal smoother. Furniture tracking combines tag observations with an object--tool transform and tag-based residual corrections during verified held intervals; offset residuals beyond 3\,cm or $20^\circ$ from the segment median are excluded before temporal smoothing. Gaps in released static segments require compatible bracketing observations (within 1.5\,cm and $10^\circ$), an admissible gripper state, and the configured gap limit. These are annotation-construction checks, not task-success tolerances. DROID combines visible pose fits with measured tool motion during carrying. Released human-dataset poses undergo the recorded coordinate conversions; interpolated and estimated intervals remain identifiable. Household visibility masks preserve unsupported intervals instead of treating them as observations. Validity masks are shared across methods and are not selected from candidate errors.

\begin{figure}[!t]
    \centering
    \includegraphics[width=\columnwidth]{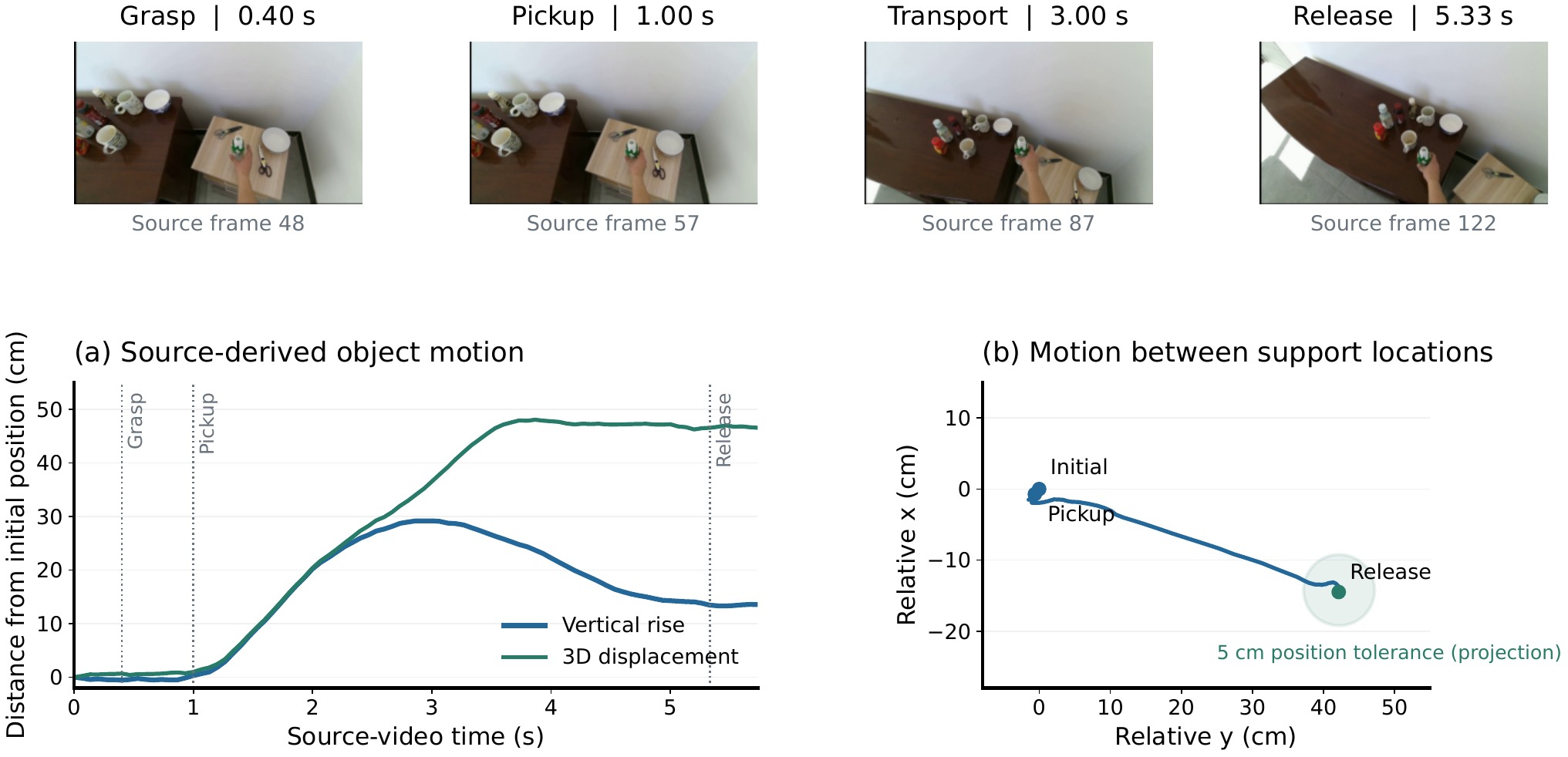}
    \caption{\textbf{From a video demonstration to an object-level evaluation specification.}
    The manually drawn outlines identify the same bottle in frames A--C; they are visual guides, not segmentation annotations. Corresponding points on its annotated height trajectory identify the initial support, transport, and receiving support; dashed lines mark the annotated pickup and release. The conditions at right define acceptable terminal placement, supplementing the required transfer. The curve is computed from source object poses, while the conditions are the evaluation targets; neither is a robot-execution result.}
    \label{fig:annotation_example}
\end{figure}

\paragraph{Illustrative annotation.}
Figure~\ref{fig:annotation_example} follows a HOI4D bottle transfer from the initial support to the receiving table. The clip contains 87 frames at 15\,Hz (source frames 42--128), with released bottle geometry of approximately $7.07\!\times\!7.00\!\times\!22.13$\,cm. Object and camera poses are composed in a fixed annotation frame, associating the same object with its motion and the receiving surface despite camera movement. The bottle rises by a maximum of 29.2\,cm and ends 13.6\,cm above its initial height. The annotation specifies pickup and release at clip indices 15 and 80, while the terminal predicate additionally requires the goal pose, support, release, and rest conditions shown in the figure. The receiving support's lateral extent is estimated from the available observations.

\paragraph{Task requirements and initial conditions.}
The task specification identifies the initial object relations, the events that define the observed interaction, and the terminal goal. A clip beginning with an object already held requires the subsequent transport and, where demonstrated, release; it does not require an unobserved pickup. A clip beginning with an object supported requires the transfer appropriate to that starting relation. These conditions constrain the interaction without requiring exact agreement with every source pose.

For placement goals, a favorable initial arrangement alone earns no completion. If the submitted object initially satisfies the goal relation, it must leave that relation and complete a new placement within the same manipulation attempt, including any required release and stability conditions. An initially satisfied goal is therefore not an automatic rejection of the entire reconstruction, but it cannot receive credit for remaining unchanged. The initial relation is evaluated before the first action; a post-action state cannot substitute for it. For a moving receiver, object--receiver relations use both poses at the same timestamp.

Required events are associated with the manipulated object and a single attempt. Release and stability require recorded evidence when the task calls for them. A release of another object or a nominal settling delay does not establish the required event. Incidental grasp choice and exact source robot motion are generally unrestricted. Native task rules can additionally require robot return to the home configuration; these explicit completion conditions are retained. These requirements are recorded with the task annotation and instantiated by the evaluation rules in Section~\ref{app:functional_correctness}.

\paragraph{From observations to task conditions.}
We identify the object relation that changes in the source clip and the events needed to establish that change. A placement annotation links transport to the receiving surface and, when demonstrated, release; removal requires the object to leave its original container; assembly identifies the part and its intended interface. Clip boundaries determine whether acquisition or release is observed and therefore required. The resulting condition is checked against the source frames and the annotated geometry before being used to assess an execution.

Physical trials and inspection of alternative solutions then test whether the condition captures the intended interaction. For example, support alone does not establish removal from a container, and an apparently seated furniture part may still correspond to the wrong interface. The family-specific predicates and numerical tolerances are collected in Section~\ref{app:functional_correctness}. Here we document their source evidence and the review process through which ambiguities and errors are resolved.

\subsection{Worked Example: From Video Input to Scored Execution}
\label{app:worked_example}

\begin{figure}[!t]
    \centering
    \includegraphics[width=\columnwidth]{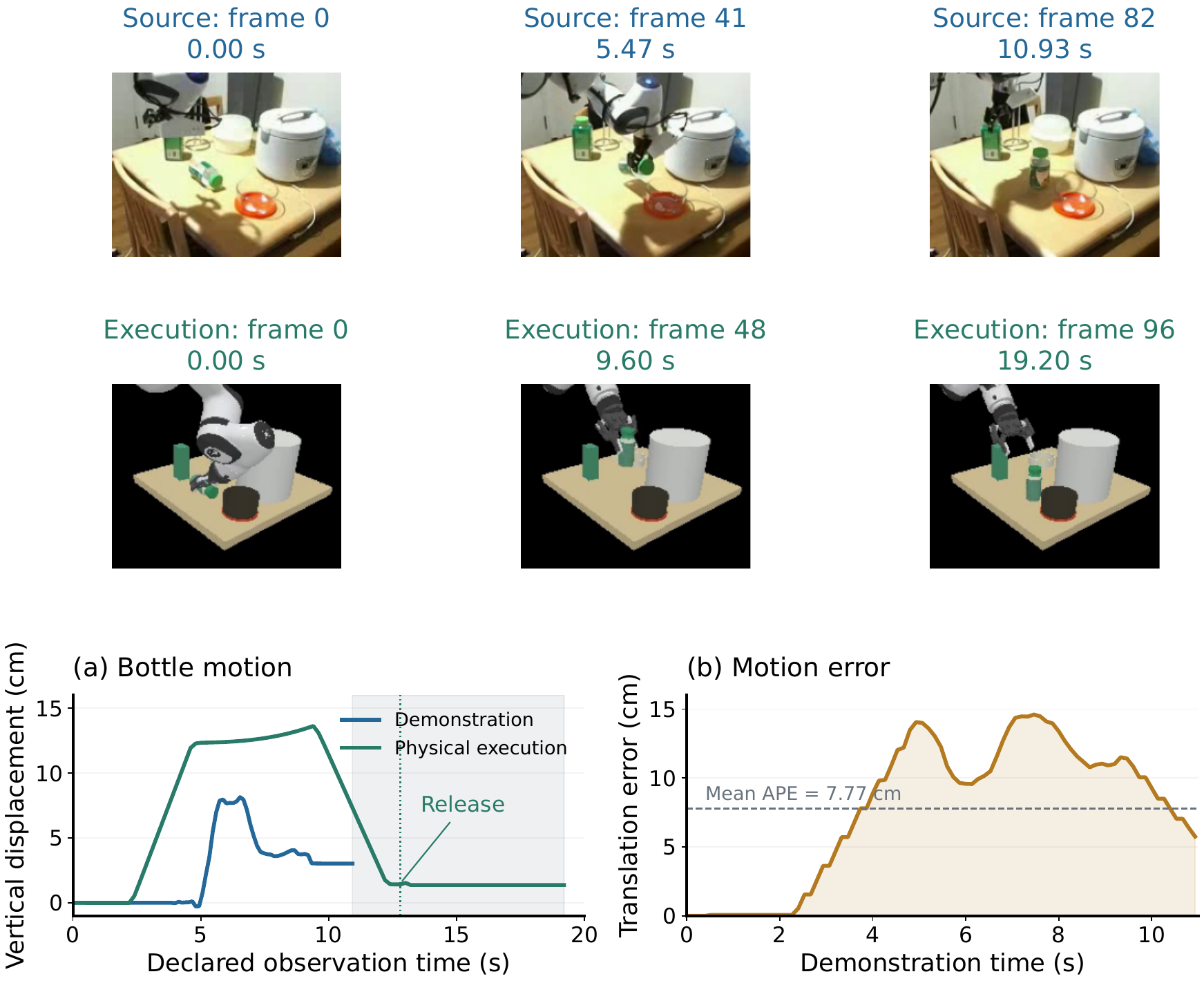}
    \caption{\textbf{A complete evaluated reconstruction.} Top: three frames from the agent's RGB input. Middle: three frames of its submitted controls executed in its reconstructed scene; the displayed times use the recorded observation clock. Bottom left: annotated and executed bottle-height changes after the object-frame correspondence used for scoring. The shaded tail lies beyond the demonstration's motion-comparison window, while functional evaluation still uses its release and terminal state. Bottom right: per-frame translation discrepancy on the demonstration clock, whose mean is 7.77\,cm. The recorded interaction succeeds despite temporal and geometric discrepancies. Full videos, annotations, controls, execution records, and scoring inputs accompany this example.}
    \label{fig:worked_evaluation_example}
\end{figure}

We trace one DROID upright-bottle instance through the complete interface, using the saved GPT-6 Astra submission from the visual-refinement experiment. Figure~\ref{fig:worked_evaluation_example} shows its input and physical execution. The example uses the unchanged submitted package and recorded rollout, scored under the current definitions in Section~\ref{app:evaluation}; no new agent construction or simulation was performed. It is an illustrative instance, not an aggregate model result.

\paragraph{Video, target configuration, and visible resources.}
The input is an 83-frame, $640\!\times\!640$ RGB clip at 7.5\,Hz from a fixed external camera, covering original episode indices 64--228 at stride two. Its first-to-last observation interval is 10.93\,s. The visible interaction is to pick up the smaller bottle lying on the table, turn it upright, and release it on the support. The assigned target is a Panda arm with a Robotiq 2F-85 gripper in SAPIEN, controlled by absolute TCP pose and gripper commands at 5\,Hz.

The agent receives the video, a contact sheet extracted from it, the public robot dimensions and TCP conventions, and the submission specification. The refinement round additionally supplies its first-pass package, a candidate-only initial render, a montage of that package's executed actions, and diagnostics reporting execution completion, inverse-kinematics error, and rest. The controller produces these simulator observations between rounds. The agent may inspect source frames, edit the package, and check its geometry and schema; this experiment does not give it direct simulator execution. In this worked example, the controller executes the candidate between rounds and supplies rendered observations and execution diagnostics to the agent for refinement. Hidden calibration, source object geometry and motion, task annotations, and scores are withheld. The archived full briefs and delivered feedback distinguish the original construction inputs from the subsequent refinement inputs.

\paragraph{Entity correspondence and hidden annotations.}
The manipulated entity is the smaller bottle, matched to the submission's moving green capped bottle; the receiving support is matched to its tabletop. The stationary bottle, glass, bowl, and cooker are contextual candidate geometry, not additional motion targets. The reference bottle is a human-reviewed fit to calibrated source observations, with full oriented-bounding-box dimensions of approximately $7.79\!\times\!7.79\!\times\!14.36$\,cm. Its 83 annotated poses, the support model, timestamps, and withheld camera registration establish geometry and motion targets in the common evaluation frame. The reference bottle begins nearly horizontal, with its axis approximately $87.53^\circ$ from vertical.

The active task annotation requires an upright bottle supported by the submitted tabletop, followed by release and rest. The bottle's neck determines the directed upright axis when the geometry resolves it. The terminal tilt must be at most $30^\circ$ from the support normal; support uses a 1\,cm clearance tolerance and 5\,mm footprint margin. Initial satisfaction alone cannot establish a new placement. The saved brief and original annotation predate the revised bottle rule and retain an absolute-endpoint condition; the current semantic predicate above governs this example's score. Absolute placement discrepancy remains visible in the geometric and motion diagnostics.

\paragraph{Submission and recorded execution.}
The final package contains the reconstructed bottle mesh, tabletop and contextual geometry, estimated camera and robot placement, and a $97\!\times\!7$ control array with interval 0.2\,s. It also includes an $83\!\times\!7$ estimated object-pose sequence, which is a declaration rather than an animation command. For scoring, the evaluator converts the camera-frame delivery into the evaluation frame using the withheld registration. Physical execution applies the submitted controls and records the bottle's states, gripper state, grasp/release events, and final rest condition.

The saved rollout contains 97 object states. On its declared observation clock, grasp is detected at 0.4\,s, transport at 2.6\,s, release at 12.8\,s, and terminal completion at 19.2\,s. These times identify the scoring record; the exported video's playback rate is not a physics clock. Motion fidelity uses the 83 demonstration timestamps, with nearest-state correspondence and full recording coverage. Forty-one execution states lie beyond that comparison window. They remain available for the functional decision, which includes the later release and terminal state. Thus, a longer successful interaction can still incur temporal motion error.

\paragraph{Inputs to the decision procedure.}
The scoring procedure receives four linked records: the validated scene and control package; the build receipt and entity correspondence; the executed object poses, terminal poses, grasp/release events, gripper state, and rest flag; and the hidden scene, source poses, camera registration, and active task specification. Geometry is evaluated from the initialized package. Motion is evaluated from the recorded execution, not the submitted pose estimates. The task decision checks the same moving bottle throughout: the initial upright goal is false, the terminal bottle is released and at rest, its support projection is admissible, and its directed axis differs from the tabletop normal by approximately $0.009^\circ$. The recorded support gap has magnitude below 0.01\,mm, within the prescribed tolerance. These conditions yield success and all four progress stages under the assigned rigid profile. They do not constitute a separate certification of every possible contact or collision.

\paragraph{Final scores and retained evidence.}
Build and Task Success are both one, and Task Progress is $100\%$. The continuous scores are
\begin{equation}
    \begin{aligned}
    (\text{Scene CD},\text{Shape CD},\text{Size},\text{Centre})
        &= (2.51,\,3.58,\,5.76,\,24.80)\ \mathrm{cm},\\
    (\text{translation APE},\text{translation RPE})
        &= (7.77,\,0.68)\ \mathrm{cm},\\
    (\text{rotation APE},\text{terminal axis error})
        &= (9.27^\circ,\,1.78^\circ).
    \end{aligned}
    \label{eq:worked_example_scores}
\end{equation}
Terminal displacement error is 4.57\,cm. It is measured relative to each trajectory's initial object position, whereas the 24.80\,cm centre error retains the initialized placement discrepancy. The low registered Scene CD therefore does not imply accurate absolute placement, and successful upright placement does not imply accurate bottle dimensions. No composite score is formed. The example supplement retains the input and rollout videos, full public prompts and feedback, both submitted and evaluation-frame packages, hidden annotations, active scoring rules, per-frame arrays, full-precision scores, and file hashes. A record-scoring entry point reproduces the measurements from these saved inputs without rerunning the agent or simulator.

\subsection{Human-Assisted Reference}
\label{app:human_assisted}
\label{app:reference_construction}
\label{app:object_agreement}

The HAR is developed separately for each instance through scene construction, robot-control design, and iterative physical testing. Source geometry, motion annotations, robot logs, and existing simulation assets provide the initial reconstruction; inspection of grasping, transport, insertion, and release guides subsequent revisions. The selected reference comprises both the reconstructed environment and the robot behavior that was executed in it.

This workflow permits access to source annotations, simulator states, task predicates, and evaluation scores. Consequently, comparison with coding agents measures the performance of reconstruction with additional information and human engineering, with selected execution identities and the available construction histories retained.

\paragraph{Construction by source.}
Furniture references use source geometry and robot/object motion to initialize grasps and transport, followed by physical approach, contact verification, alignment, and release. The current 64 selected furniture executions use PD arm and gripper control with frictional holding; their execution conditions are detailed below.

DROID references combine estimated object geometry with recorded robot motion or object-guided control fitting, retaining the derivation used by the selected package. Human-video references use released hand observations where available, object trajectories, and geometric contact selection to initialize target robot poses. In a carried segment, an initial grasp transform maps object motion into tool targets; physical execution determines whether the grasp actually holds. Contact search and optimization can revise wrist/palm poses and finger controls using object-motion, hand-pose, contact, and validity feedback. RoboDojo uses native task assets and recorded joint-control executions. Household references retain the selected source simulation or documented native reconstruction, with its own state and video identity. Pushing, rope, and toy references adapt the frozen reconstructed twins to the target robot and native execution mechanics.

\paragraph{Additional information by source.}
Table~\ref{tab:har_resources} identifies the resources beyond the agent's RGB input that support each reference workflow. These resources can supply geometry, calibration, motion, or an existing physical execution, rather than merely assisting visual interpretation. Their reuse is part of the HAR method. In the furniture feedback controller, privileged simulator object state also guides execution. Accordingly, the reference measures source-assisted reconstruction and robot engineering; it does not estimate unaided human reconstruction from RGB.

\begin{table}[!t]
    \centering\small
    \caption{\textbf{Information available to human-assisted reconstruction beyond RGB.} Inputs vary by source and instance. The final column describes their use in constructing the selected reference; it does not claim that all assets were reconstructed from video. The evaluated coding agents do not receive these source-specific annotations or reference executions.}
    \label{tab:har_resources}
    \setlength{\tabcolsep}{4pt}\renewcommand{\arraystretch}{1.12}
    \begin{tabularx}{\linewidth}{@{}>{\raggedright\arraybackslash}p{0.15\linewidth}>{\raggedright\arraybackslash}p{0.36\linewidth}>{\raggedright\arraybackslash}X@{}}
    \toprule
    Source & Additional annotations and assets & Use in reference construction \\
    \midrule
    FurnitureBench & Part CAD, calibrated part/robot motion, assembly targets, simulator state. & Reuse part and collision assets; initialize grasps and transport; refine contact, alignment, and release. \\
    DROID & Fitted object/support geometry, calibration, robot kinematics, object motion; visible cloth surfaces. & Initialize scene and controls from source records; fit object-guided motion and refine execution. The two cloth references retain their declared assisted grasp mechanism. \\
    RoboDojo & Native scenes, robot and object assets, task rules, recorded controls and states. & Adapt the recorded native task execution to the benchmark interface. \\
    HOI4D & Instance CAD, object and camera trajectories, fitted support, available hand observations. & Retarget object transport to the arm or hands; optimize contact using annotated motion. \\
    HOT3D & Object CAD, tracked object/camera motion, available hand poses, support estimates. & Initialize grasp and transport, then refine wrist/palm and finger controls. \\
    DexYCB & YCB meshes, calibrated views, object/hand poses, estimated support height. & Reuse geometry and pose observations to construct and refine lifting behavior. \\
    OakInk2 & Part scans, calibrated motion capture, task segments, hand/wrist motion. & Prepare geometry and object/hand-guided control targets; refine contact execution. \\
    Household & Source assets, accepted RGB-D pose or surface estimates, paired source simulation or recovered execution records. & Retain the selected recorded simulation or documented native reconstruction, preserving its state/video identity. \\
    Reconstructed twins & Frozen reconstructed geometry and rigid/particle motion; robot and native simulation assets. & Adapt the reconstructed target to xArm7 pushing, rope routing, or toy packing. \\
    \bottomrule
    \end{tabularx}
\end{table}

\paragraph{Human effort and recorded iterations.}
The retained records do not provide systematically measured person-hours or a complete count of manual interventions by source. Human inspection, asset preparation, controller development, and automated search were interleaved. Process runtimes, file timestamps, and optimizer evaluations therefore cannot be converted into human labor. We report the coverage of identifiable trial ledgers and leave unrecorded human-time totals unspecified, rather than imputing a uniform budget.

\begin{figure}[!t]
    \centering
    \includegraphics[width=\columnwidth]{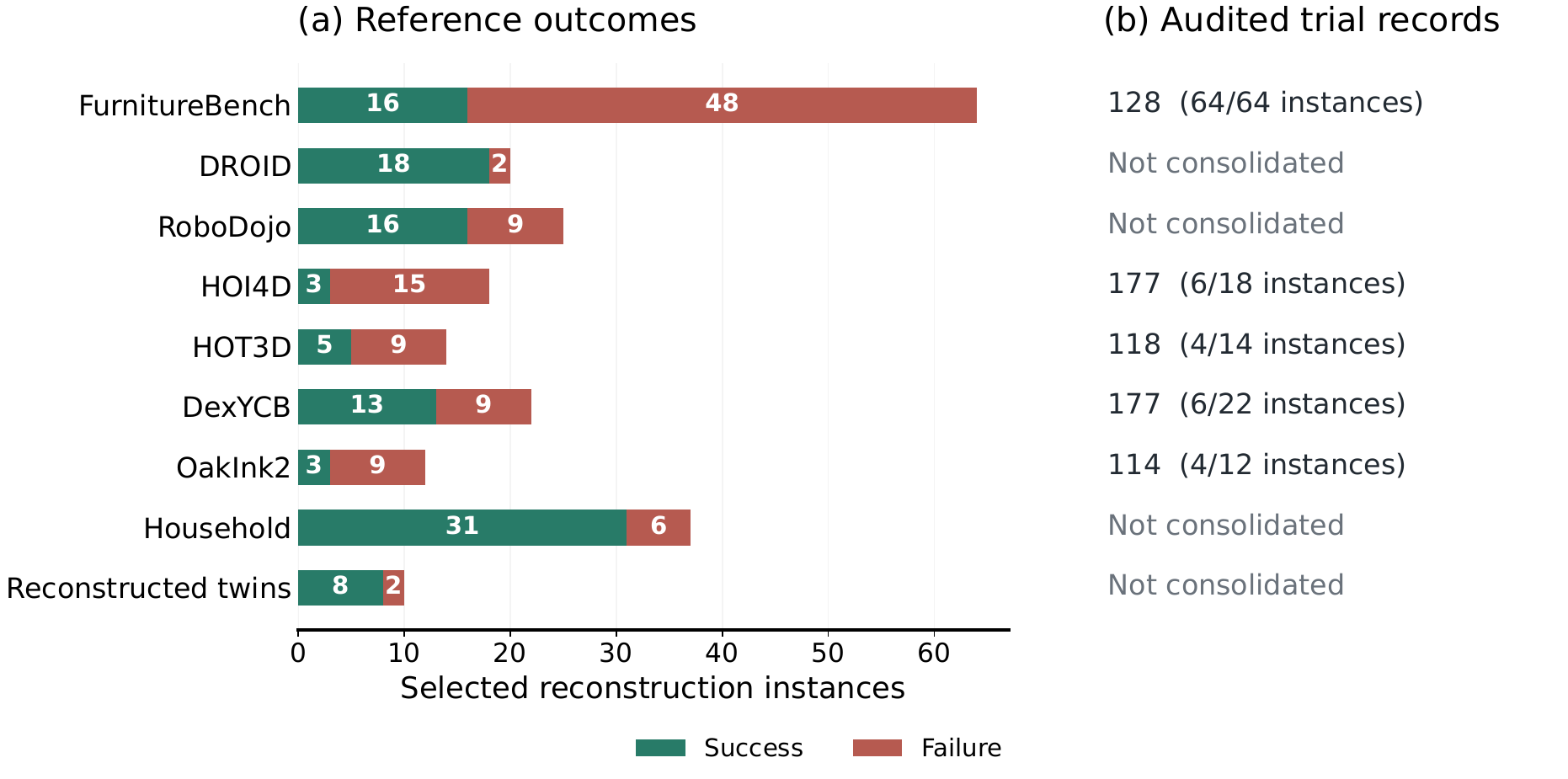}
    \caption{\textbf{Source-level reference outcomes and recorded construction effort.} Left: outcomes of the selected references under the current evaluation snapshot, including failures and suspended task targets. These are instance counts; the headline score instead gives equal weight to task families. Right: trial counts from the two audited campaigns, with the number of covered configurations relative to that source's inventory. Unconsolidated histories are explicitly marked. Trial records do not measure human person-hours or complete manual-iteration counts, and coverage differs across sources.}
    \label{fig:har_source_disclosure}
\end{figure}

Figure~\ref{fig:har_source_disclosure}(b) summarizes two documented campaigns. The furniture campaign covers 64 instances with 128 attempt records, including two pilots: 127 contain complete physical records and one records a rendering-initialization interruption. Attempts per instance range from one to six, with median two. A separately audited human-video contact-search batch covers 20 of the 66 arm/hand configurations: 177 physical trial records for six HOI4D configurations, 118 for four HOT3D configurations, 177 for six DexYCB configurations, and 114 for four OakInk2 configurations. Each covered configuration has 26--31 records, including baseline, seeded, and optimized trials. These counts describe saved simulator trials within those campaigns, not manual editing iterations or the total historical effort for each source. DROID, RoboDojo, household, and reconstructed-twin histories lack a comparably consolidated attempt count in this audit; their missing totals are not zero.

\paragraph{Evaluation feedback and unsuccessful references.}
Reference development permits access to task predicates, reference-based errors, physical diagnostics, and success/progress judgments. The furniture campaign uses scored retries and observed contact failures; the human-video search records use annotated object/hand errors and contact/validity feedback. These workflows are consequently informed by evaluation targets. For reused native or source executions, the retained evidence establishes the selected run and its outcome but does not provide a complete history of how often a person consulted scores. We make no claim of evaluation-blind development for any source.

Both successful and unsuccessful selected references are retained in the benchmark. All 222 instances have a selected reference execution and are included in functional evaluation. The selected execution records the retained reconstruction, not first-attempt success. Earlier failures remain in the available construction histories but do not become additional test instances. Figure~\ref{fig:har_source_disclosure}(a) gives the source-level counts. All 222 instances have a selected execution receipt; this records the retained reconstruction, not first-attempt success. Earlier failures are retained in the available construction histories but do not become additional test instances. The furniture source-selection cap described in Section~\ref{app:data_sources} remains a separate, outcome-informed admission choice.

\paragraph{Interpretation of the reference score.}
HAR achieves 58.8\% Task Success, macro-averaged with equal weight across the 39 task families covering all 222 benchmark instances, consistent with Table 2. The score characterizes the selected outcomes of source-assisted reconstruction and human engineering. It does not isolate human visual-reconstruction ability or performance under the coding agents’ information and construction budgets. Together, the resource disclosure, trial coverage, feedback policy, and retained failures specify the scope of this comparison.

For instance $i$, the reference and its execution are
\begin{equation}
    P_i^h=(\eta_i^h,\pi_i^h),
    \qquad
    \tau_i^h=\mathcal{R}_{c_i^h}(\eta_i^h,\pi_i^h).
\end{equation}
Here $c_i^h$ records the simulator, embodiment, controller,
initialization, and observation settings used for the selected
HAR execution. The HAR scene $\eta_i^h$ is evaluated against the
annotated reference geometry $\eta_i^*$, although the two may share
source assets. Shared assets and conversions are documented.
When HAR directly reuses evaluation geometry, its geometric score
measures correspondence to that geometry and does not independently
establish reconstruction ability from RGB. Demonstrated object
trajectories provide the motion-evaluation targets, while
$\tau_i^h$ records the execution of the selected HAR program.

\paragraph{Execution conditions and comparison.}
HAR and coding-agent reconstructions are evaluated against the same
applicable geometry, motion, and task-completion targets using common
metric definitions. Each method executes its own robot behavior in
its own reconstructed scene. Execution settings are recorded for
each method.

For the furniture configurations, both HAR and the coding-agent
executor use 100\,Hz physics integration and 20\,Hz joint control.
HAR observes the terminal state for 10\,s, while the coding-agent
executor uses 1\,s of terminal settling. HAR uses PD arm and gripper
control, a friction coefficient of 1, and a 70\,N finger-force limit,
without grasp attachments, pose snapping, or assembly latches.
The coding-agent executor additionally permits the bounded initial
grasp-alignment correction and, for applicable insertion tasks,
the seating-retention approximation described in Appendix~\ref{app:simulation_environment}.
These differences can affect contact outcomes and terminal stability.

The reported comparison therefore evaluates the complete
reconstruction workflows, including differences in information
access, human refinement, and recorded execution conditions.
Isolating the effect of the construction method alone would
require matched execution conditions.



\paragraph{Human trials and alternative solutions.}
Construction involved author-directed attempts, inspection of alternative solutions, and explicit review of failed interactions. The furniture campaign covers 64 instances from 46 original episodes. It records 126 launches for the main campaign: 125 completed physical records and one rendering-initialization interruption, plus two earlier pilot records. Completion of a record includes unsuccessful or early-aborted task execution. Revisions included sustained bilateral-contact confirmation, gripper-width feedback, lift verification, measured in-hand pose, grasp-axis search, slower release, and stopping after loss of hold.

A subsequent two-task study tested two alternatives each for a desk leg and a cabinet door. All four completed physical trials failed; four initialization diagnostics and two early stops were retained separately. The leg trials compared height-aligned insertion with additional small contact searches. Door trials examined 47 grasp candidates and an expanded set of 576 candidates, with geometric and kinematic rejection reasons retained. These attempts exposed a missing door-axis/interface annotation: the annotated endpoint itself was rejected by the old target rule. The response was to repair the task specification and re-evaluate saved execution, rather than interpreting every failed attempt as evidence that the controller or embodiment could not solve the task.

\paragraph{Stopping, feedback, and selection.}
The comparison does not assume a common construction budget. Available campaign records retain search parameters, attempt limits, and selection decisions, but they do not establish complete human-effort totals for every reference. In the current furniture campaign, the selected execution is the latest successful trial if one exists, otherwise the latest complete trial. Earlier reference campaigns used their declared variant ordering and retained a successful selected execution or the last actual execution when all attempts failed. Subsequent scoring revisions preserve the selected execution and recompute its scores; they do not select a new winner from history. Failed builds, physical failures, and planning attempts that never reached execution remain distinguishable.

Reference success is neither an annotation-acceptance condition nor a guaranteed upper bound. The main comparison executes each method's own behavior in its own scene, without transferring controls between reference and coding-agent scenes. Differences reflect the complete construction workflows, including privileged information, engineering effort, and any recorded execution-condition differences.

\subsection{Human Evaluation}
\label{app:human_evaluation}

\paragraph{Evaluation target and stimuli.}
We assess overall scene-and-interaction faithfulness through pairwise comparisons of the nine coding-agent systems and HAR. Each comparison presents the source demonstration, denoted R, alongside two reconstructions of the same case, denoted A and B. A case identifies a source demonstration and target configuration; different embodiments of the same source video can form distinct cases. Each coding-agent video shows its own generated behavior executing in its own reconstructed scene. HAR participates as an ordinary candidate, rather than as the source video or a guaranteed correct answer.

The rating question was: ``Which result more faithfully reconstructs the scene and interaction in the real video?'' Raters were instructed to consider objects and layout, task semantics, motion, and visible physical plausibility. They were also instructed that geometric simplification or a different robot embodiment does not automatically imply failure, and that visual detail alone should not determine preference. This question measures holistic faithfulness rather than a separate task-success judgment.

Videos were displayed without audio at $640\times480$ resolution using aspect-preserving resizing and padding. We retained the complete recorded interaction, including prelude, waiting, and terminal settling, at its original duration and $1\times$ playback speed. Videos could be played independently or together; neither camera viewpoints nor trajectory timing were forced to match. A preference submission required at least 90\% playback coverage of each playable video. ``Unable to judge/playback issue'' could be submitted without meeting that threshold. All retained preference judgments met the playback requirement, which checks playback rather than sustained attention.

\paragraph{Case sampling and comparison coverage.}
The study used a frozen pool of 195 cases from 30 task families, preceding the final 222-instance benchmark inventory. We sampled 54 cases using task-family stratification and seed 20260924. Each family received at least one case; the remaining slots were allocated by the largest deficit relative to the family's proportional share of 54, subject to availability. Cases were then sampled within families using the fixed seed.

We planned two agent--agent comparisons per sampled case, giving 108 comparisons, with each of the 36 agent-pair types assigned to three cases. We then added three HAR comparisons per agent across 27 distinct families, retaining the original sampled cases and agent comparisons. Thus, the complete plan contained 135 case--candidate-pair comparisons, with each of the 45 system-pair types represented three times. Pair allocation used identifiers, family membership, and assignment counts, without consulting automatic scores, task-success labels, or human preferences.

Media and provenance checks admitted 83 comparisons from 41 cases: 70 agent--agent and 13 agent--HAR comparisons. The remaining 52 comparisons were deferred, without replacement by higher-quality or successful examples. Admission required verifiable source correspondence and complete execution footage, or a confirmed agent-caused build failure represented by an anonymous failure panel. Missing media or infrastructure failures were deferred rather than treated as candidate losses. The released comparisons covered 43 of the 45 system-pair types; the effective comparison graph connected all ten systems, and HAR was compared with every agent. Actual coverage was therefore less balanced than the planned design. Playable executions were retained regardless of task success.

\paragraph{Participants, masking, and assignment.}
Twelve invited internal raters, comprising paper authors and company employees, participated voluntarily with informed consent and received no additional compensation. They were not informed of the specific experimental purpose. Each released comparison was assigned to three distinct raters using a seeded allocation that balanced workload. Nine raters completed 21 primary comparisons each and three completed 20 each. Raters were instructed to answer independently.

The interface, media addresses, and personal exports concealed system names, HAR identity, automatic scores, and other raters' answers. The organizers retained the identity mapping. A/B positions were randomized independently for each assignment, and question order was shuffled within each rater. R remained the fixed source reference; the interface did not prescribe a randomized playback sequence. Masking candidate labels does not establish that internally recruited raters had no prior familiarity with the outputs.

\paragraph{Responses and quality checks.}
The five responses were A better, B better, equivalent, both poor, and unable to judge/playback issue. The 83 comparisons yielded 249 primary judgments: 79 A-better, 82 B-better, 14 equivalent, 68 both-poor, and six unable-to-judge responses. The six excluded responses came from two comparisons for which all three raters selected unable to judge; both candidates in each comparison were confirmed build failures. The remaining 243 judgments covered 81 comparisons across 39 cases and 19 task families, with three valid judgments from distinct raters per comparison. Seven cases contributed three judgments each, 22 contributed six, and ten contributed nine. Raters could overlap across different comparisons of the same case. Valid preferences covered seven of the nine source categories in the sampling pool; no Poway or RoboDojo comparison passed admission in this study.

Each rater also received one unannounced repeat comparison, separated from its original presentation by at least eight primary questions and with independently randomized A/B positions. These 12 repeats were excluded from ratings and confidence intervals. After aligning candidate identities, repeat-answer consistency was $8/12$; no rater was excluded on this basis. Across the 81 valid comparisons, all three raters agreed in preference direction on 31, counting equivalent and both-poor responses as ties. These are descriptive consistency measures.

\paragraph{Human Preference Rating.}
We fit a Bradley--Terry model to all 243 valid judgments, rather than to majority votes. For judgment $t$ comparing candidates $i_t$ and $j_t$, let $y_t=1$ when $i_t$ is preferred, $y_t=0$ when $j_t$ is preferred, and $y_t=0.5$ for either equivalent or both-poor responses. Unable-to-judge responses are excluded. With $d_t=\beta_{i_t}-\beta_{j_t}$, we minimize
\begin{equation}
    \mathcal{L}(\boldsymbol{\beta})
    = \sum_t \left[\log(1+\exp(d_t))-y_t d_t\right]
      + \frac{0.001}{2}\sum_{m=1}^{10}\beta_m^2.
    \label{eq:human_bt}
\end{equation}
Each valid judgment has unit weight; we do not additionally equalize case or family weights. Equivalent and both-poor responses each contribute half a win and half a loss, while their counts remain separately reported. Both-poor responses account for $68/243=28.0\%$ of valid judgments. No separate tie-probability parameter is fitted.

We optimize using L-BFGS-B from zero initialization, with \texttt{ftol=1e-12}, \texttt{gtol=1e-8}, and at most 1,000 iterations. After checking convergence, we center the ten fitted parameters and convert them to the Elo scale:
\begin{equation}
    \mathrm{HPR}_m
    = 1000+\frac{400}{\ln 10}(\beta_m-\bar\beta),
    \qquad \bar\beta=\frac{1}{10}\sum_{m=1}^{10}\beta_m.
    \label{eq:human_hpr}
\end{equation}
The mean Human Preference Rating (HPR) is therefore 1000. HPR is a batch estimate of relative preference, not an online Elo update or a task-success percentage.

\paragraph{Confidence intervals.}
We use 2,000 source-video-cluster bootstrap replicates with seed 20260924. Clusters are defined by the SHA-256 hash of the original demonstration, keeping all associated embodiments, comparisons, and raters together. The 41 cases with primary responses form 39 source-video clusters. Two clusters contain only unable-to-judge responses and remain in the resampling pool; the 243 valid judgments originate from 37 nonempty clusters. Each replicate samples 39 clusters with replacement, carries their valid judgments with the corresponding multiplicities, and refits all ten ratings with the same regularization and centering. Sampling an empty cluster adds no judgments. All 2,000 fits were valid. The 95\% intervals in Figure~\ref{fig:human_comparison}(a) are the 2.5th and 97.5th percentiles of each system's bootstrap ratings.

\paragraph{Interpretation and scope.}
The intervals quantify uncertainty under source-video resampling for this rater group; raters are not independently resampled. We do not infer significant pairwise differences from the point-estimate ranking. The study measures relative holistic faithfulness under its observed case coverage and internal participant pool. HAR uses privileged information and human refinement, and its videos came from an independently versioned reference release; participation in the same preference study does not imply matched construction budgets or identical automatic-evaluation snapshots. Automatic scores do not enter the preference model or determine its outcomes. The study compares complete reconstruction workflows and complements automatic evaluation with an assessment of their visible outputs.

\subsection{Annotation Verification and Evaluator Refinement}
\label{app:annotation_verification}

Verification examines whether the annotations represent the demonstrated interaction and whether the evaluator recognizes valid executions of that interaction. Source videos, geometric overlays, object trajectories, and task semantics provide the evidence used to define and adjudicate evaluation criteria. Reference and agent-generated executions are used only to expose ambiguous or incomplete predicates; they do not determine the intended task outcome or method-specific scoring rules. Any identified ambiguity is resolved against the source demonstration and task specification, after which the revised evaluator is applied uniformly to all submitted methods. A different grasp or transport path is therefore admissible when it preserves the task-defining events and outcome.

\paragraph{Author review and adjudication.}
Authors inspect the source video, annotated object motion, reference execution, and task predicates together. An accepted annotation can accompany an unsuccessful HAR execution, and a geometric modeling concern need not invalidate an otherwise successful interaction. Accordingly, these counts describe review coverage, not agreement with binary task-success labels.

Review identified two erroneous success decisions, in bulb insertion and drawer insertion, and three additional progress disagreements, in cabinet-top and drawer assembly. In these cases, the scoring criteria were corrected by referring back to the demonstrated interaction and its task-defining events. Other review notes concern interpenetration, inaccurate geometry, or uncertain endpoints. Corrections were incorporated into the shared evaluator and the affected submissions were rescored under the same revised criteria.

All results reported in the paper are computed using the same final evaluation snapshot; no reported method is scored under a method-specific or earlier predicate version.

\paragraph{Human adjudication of scoring errors.}
Saved executions also document two complementary scoring failures and their corrections. First, an endpoint-only rule accepted a lampshade placement whose initial state already satisfied the broad goal condition. In the saved 36-execution control study spanning 12 source clips, the original action sequence and two modified sequences for this instance all received full success and progress. Requiring departure from the initial goal and a new placement changed all three judgments to failure, with progress reduced to 0.25; the other 33 judgments were unchanged. These are erroneous acceptances of an interaction that was not completed.

\begin{figure}[!t]
    \centering
    \includegraphics[width=\columnwidth]{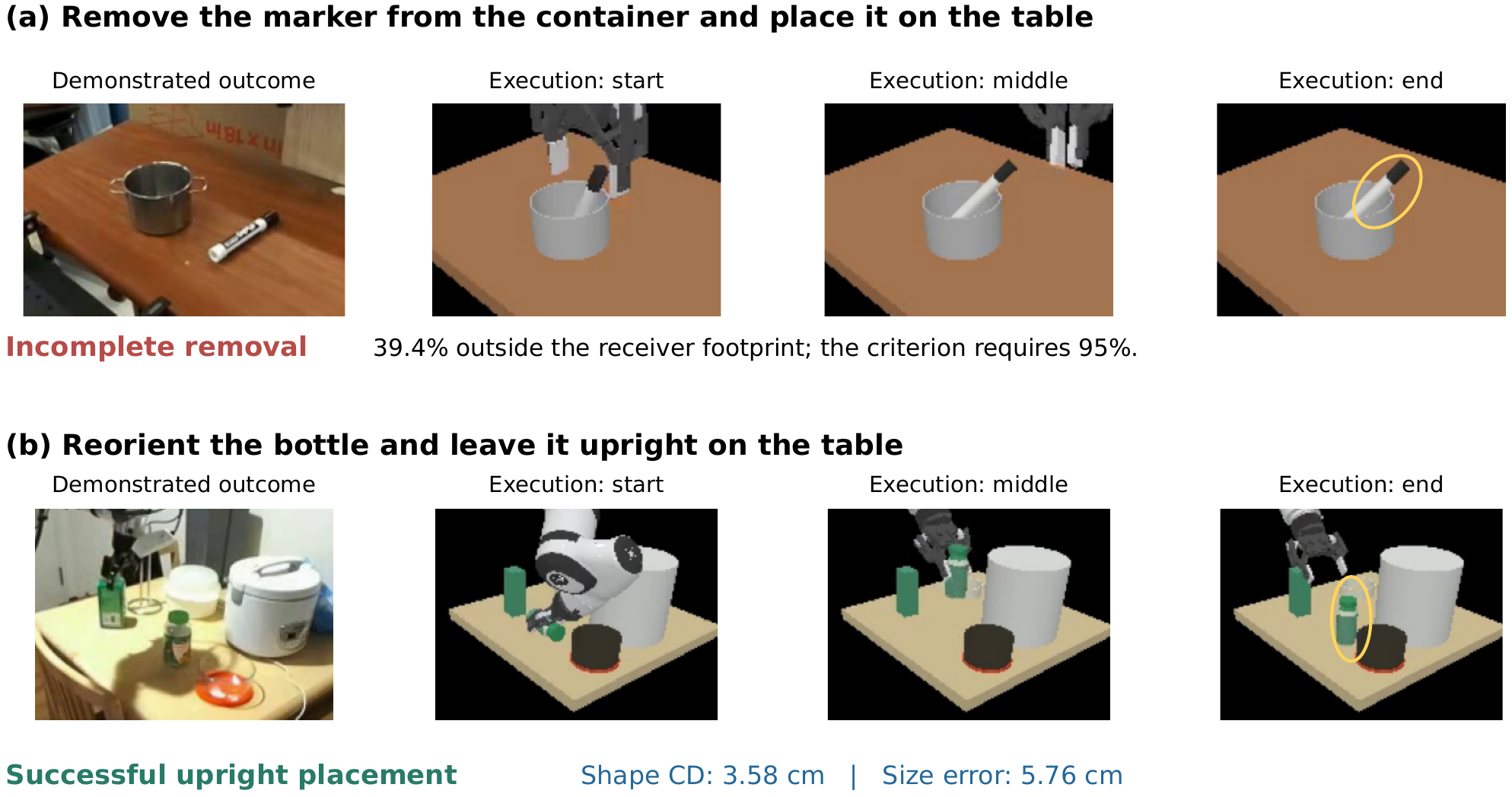}
    \caption{\textbf{Evaluating the requested interaction separately from geometric fidelity.}
    Each row shows the demonstrated outcome and three frames from the scored Astra visual-feedback execution, with manually drawn endpoint outlines identifying the manipulated object. (a) The marker remains partly in the container; robot withdrawal does not establish object extraction. (b) The bottle is reoriented and placed upright, satisfying the task predicate despite the reported shape and size errors. Execution frames are 0, 36, and 72 for (a), and 0, 48, and 96 for (b). Judgments use the complete saved rollout; frames are cropped for visibility and are not temporally aligned with the source endpoint.}
    \label{fig:construction_cases}
\end{figure}

Second, two Astra bottle reconstructions were rejected despite an upright, released, supported endpoint. The earlier comparison paired opposite ends of the reference bottle axis and also imposed proximity to an unnecessary absolute source endpoint. Adjudication instead required the candidate bottle's directed axis to point upward, together with support, release, rest, and an actual reorientation. Both reconstructions then satisfied the revised predicate in a review covering six executions of this source clip. Figure~\ref{fig:construction_cases}(b) shows the visual-feedback reconstruction. These erroneous rejections concern task interpretation; geometric discrepancies remain measurable under the separate reconstruction metrics. The documented cases are examples of evaluator refinement, rather than an estimate of error prevalence across the benchmark.

\paragraph{Counterexamples and admissible alternatives.}
Figure~\ref{fig:construction_cases} contrasts incomplete extraction with successful reorientation. In the extraction case, the marker remains partly inside its receiver after the robot withdraws: only 39.4\% of its vertices lie outside the receiver footprint, below the 95\% removal criterion. In the reorientation case, the bottle is placed upright, but its Shape CD and Size error remain 3.58\,cm and 5.76\,cm. Thus, successful interaction and geometric reconstruction accuracy require separate measurements. A further cup-placement reconstruction satisfies its relative-placement goal with Scene CD 3.64\,cm, Shape CD 3.05\,cm, and Size error 6.23\,cm.

Counterexamples also test whether the rules reject an already solved initial arrangement without excluding legitimate alternative solutions. Native solved-start probes require a new object-level completion, rather than robot return home alone. Conversely, two shortened Push-T executions retain success because the object reaches the goal before truncation: their position/orientation errors are 1.09\,cm/$5.90^\circ$ and 2.27\,cm/$7.80^\circ$, within the 4\,cm/$15^\circ$ tolerance. A modified or shorter action sequence is therefore evaluated by its physical outcome, not assigned a failure label by construction.

\paragraph{Revision and consistent rescoring.}
The recorded revisions address initial-goal shortcuts, object-specific release and stability, central versus corner leg interfaces, door-axis annotations, and directed bottle orientation. Container and support rules are defined on the receiver's free cavity and bearing surface. A corrected condition applies to every method under the same task contract. Changes confined to scoring are evaluated on the retained executions; changes to physical execution require a new rollout. Annotation, evaluator, and execution versions preserve the association between a reported judgment and the evidence on which it depends.

\section{Evaluation Details}
\label{app:evaluation}
\raggedbottom

Video2World evaluates geometric fidelity, fidelity of executed object motion, and functional correctness separately. Geometry describes the initialized reconstruction; motion metrics compare its physical execution with the demonstration; task metrics assess whether the requested interaction is completed. These quantities characterize a complete reconstruction, including both its environment and its robot behavior. Robot trajectories provide execution evidence but are not motion-matching targets.

For method $m$ and instance $i$, the evaluated reconstruction and rollout are
\begin{equation}
    P_i^m=(\eta_i^m,\pi_i^m),\qquad
    \tau_i^m=\mathcal{R}_{c_i^m}(\eta_i^m,\pi_i^m).
\end{equation}
The record $c_i^m$ identifies the assigned simulator and embodiment and the controller, initialization, sampling, and settling conditions actually used. The HAR execution settings are specified in Section~\ref{app:human_assisted}. Each method executes its own behavior in its own scene; fixed-reference-action replays, where retained, are separate diagnostics. Superscript $*$ denotes the evaluation annotations described in Section~\ref{app:hidden_specifications}, including explicitly modeled targets for reconstructed twins. The method superscript is suppressed below. Metric applicability follows the available annotation and the assigned evaluation profile.

\subsection{Geometric Fidelity}
\label{app:geometric_fidelity}

Rigid-scene evaluation compares the initialized geometry in the prescribed common frame. Scene Chamfer Distance measures visible scene correspondence after bounded registration. Object-level measurements separately retain physical size and initial placement. Distances in the definitions below are in metres and are reported in centimetres.

\paragraph{Surface extraction and visibility.}
Task-relevant entities are associated through the instance's semantic roles. Where an object is delivered as several parts, the registered assembly procedure combines the corresponding physical parts before surface sampling. Supports are distinguished from manipulated and receiving entities. Mesh surfaces are sampled uniformly by area at $10^4$ points/m$^2$, with 300--40,000 points per entity and random seed zero. Canonical triangle ordering makes the scene samples independent of vertex indices, face ordering, and winding. The dense reference surfaces used as distance targets use 64 times this density and upper sample limit. Candidate samples are deduplicated into 4\,mm voxel centres.

Visibility is determined with the per-instance evaluation camera after the configuration's coordinate conversion. Samples lie within a 0.9\,m horizontal radius of the robot base, in front of the camera by more than 5\,cm, and within the calibrated image's valid content rows, excluding letterbox padding. A rasterized depth buffer removes occluded samples using a 1\,cm tolerance plus the depth range in a $3\!\times\!3$ pixel neighbourhood. The candidate's identified table is excluded from the depth buffer used for its object surfaces, so a misplaced support cannot hide those objects from comparison. A required reference manipulated or receiving entity with no visible samples retains its unoccluded surface within the workspace. Profiles without a calibrated evaluation camera use the workspace surface directly.

\paragraph{Interaction weights.}
Let $\mathcal{E}_i$ contain the reference non-support entities with surface samples. A manipulated object or its designated receiver receives weight two. For another entity $e$, let $d_{i,e}^{\mathrm{ctx}}$ be its minimum distance to the demonstrated TCP path and the manipulated object's swept geometry. Its weight is
\begin{equation}
    w_{i,e}=\operatorname{clip}_{[0,1]}
    \left(\frac{0.25\,\mathrm m-d_{i,e}^{\mathrm{ctx}}}{0.20\,\mathrm m}\right).
    \label{eq:eval_context_weight}
\end{equation}
Thus, nearby context receives unit weight within 5\,cm and zero weight beyond 25\,cm. The sweep uses at most 300 distinct object vertices at regularly spaced source frames, with stride $\max(1,\lfloor T/15\rfloor)$ for a $T$-frame trajectory; context entities use at most 1,500 vertices. Vertex subsampling uses seed zero, and weights are rounded to three decimal places. Nonfinite source TCP observations are excluded. An entity whose context distance cannot be determined receives unit weight; if no demonstrated motion is available, all context entities receive unit weight. These weights depend on the evaluation annotations rather than the candidate.

\paragraph{Interaction-weighted surface distance.}
Write $\mathcal{S}_{i,e}$ for an entity's reference samples, $\mathcal{B}_{i,e}$ for its dense surface, and $\mathcal{B}_{i}^{\mathrm{tab}}$ for the dense reference support. Let $\mathcal{C}_i$ be the candidate non-support voxel centres and $\widehat{\mathcal{C}}_i=g_i(\mathcal{C}_i)$ their registered positions. With
\begin{equation}
    d(p,\mathcal Q)=\min_{q\in\mathcal Q}\|p-q\|_2,\qquad
    \mathcal{B}_i=\mathcal{B}_{i}^{\mathrm{tab}}\cup\bigcup_e\mathcal{B}_{i,e},
\end{equation}
the forward entity residual is
\begin{equation}
    a_{i,e}=\operatorname{med}_{p\in\mathcal{S}_{i,e}}d(p,\widehat{\mathcal{C}}_i),
    \qquad \delta_{i,e}=\min\{a_{i,e},c_{\mathrm{sc}}\},
    \qquad c_{\mathrm{sc}}=0.10\,\mathrm m.
\end{equation}
This residual compares each annotated entity with the complete candidate non-support surface. Semantic correspondence determines roles and assembly membership; it does not impose a separate binary presence test on this nearest-surface distance. For a candidate point $p$, let $e_i(p)$ be the entity containing its nearest dense reference non-support sample, and set $\omega_i(p)=w_{i,e_i(p)}$. Scene Chamfer Distance is
\begin{equation}
    d_i^{\mathrm{sc}}=
    \frac{\sum_e w_{i,e}\delta_{i,e}}{\sum_e w_{i,e}}
    +\operatorname{wmed}_{p\in\widehat{\mathcal{C}}_i;\,\omega_i}
        \min\{d(p,\mathcal{B}_i),c_{\mathrm{sc}}\}.
    \label{eq:eval_scene_cd}
\end{equation}
The weighted median is the first sorted residual whose cumulative weight reaches half the total weight. If all weights in either term are zero, that term uses its unweighted counterpart. Support is included as a reverse-distance target, although candidate support samples are scored separately. If the candidate has no non-support samples, both terms receive the 10\,cm cap, giving a 20\,cm Scene CD. Missing measurable reference surfaces constitute an evaluation error.

\paragraph{Bounded registration.}
Scene registration uses
\begin{equation}
    g(p)=sRp+t,\qquad s\in[0.95,1.05],\qquad
    \theta(R)\leq15^\circ,
    \label{eq:eval_scene_registration}
\end{equation}
with free translation. A deterministic finite search initializes translation from the identity, scene centroids, coordinate-wise medians, bounding-box centres, and available corresponding-entity centroids. Each initialization undergoes at most 30 similarity iterative closest point (ICP) iterations, retaining the closest 70\% of point correspondences. Scale is bounded at each update; a rotation beyond $15^\circ$ is clamped about its estimated axis and translation is refitted. The two best hypotheses undergo additional translation-only Powell refinement within $\pm10$\,cm of their fitted translations, using at most 180 objective evaluations and position/objective tolerances $10^{-5}$/$10^{-6}$. The final hypothesis minimizes Eq.~\ref{eq:eval_scene_cd} among the fitted candidates and the identity transform. This finite search does not assert a global optimum.

Supports do not anchor object registration. Their classification is checked again in the registered frame, with refitting if the classification changes. Native multi-object scenes additionally use the declared initial-object anchor to establish their common frame before surface comparison. This coordinate preparation and the bounded scene fit are recorded separately. The fitted scene transform is not applied to object-placement or trajectory errors.

\paragraph{Missing geometry, extra geometry, and supports.}
With threshold $\rho=0.05$\,m, the coverage diagnostics are
\begin{equation}
    f_i^{\mathrm{miss}}=
    \frac{\sum_e w_{i,e}\mathbf{1}[a_{i,e}>\rho]}{\sum_e w_{i,e}},
    \qquad
    f_i^{\mathrm{extra}}=
    \frac{\sum_{p\in\widehat{\mathcal{C}}_i}\mathbf{1}[d(p,\mathcal{B}_i)>\rho]}
         {|\widehat{\mathcal{C}}_i|}.
\end{equation}
Missing geometry is therefore a weighted fraction of entities, whereas extra geometry is a fraction of candidate voxels. The former uses an unweighted mean if all entity weights are zero. Empty candidate geometry gives missing fraction one and extra fraction zero, with the empty denominator recorded. An entity classified as missing may have a residual between 5 and 10\,cm; its distance contribution is still the capped measured residual.

Explicit supports and identified table parts are evaluated separately. Otherwise, a candidate part is table-like when its horizontal span is at least 45\,cm, its median distance to the reference table plane or surface is at most 10\,cm, and fewer than half its samples lie within 3\,cm of a reference non-support entity. Declared physical assemblies retain their object role. Table error is the median distance from the registered candidate support samples to the annotated plane or dense support surface, capped at 10\,cm. A missing candidate support receives this cap; absent reference support makes the measure inapplicable. For instances with source-annotated partial tabletop boundaries, an additional image-space diagnostic measures distance to the projected candidate boundary using the delivered camera and no fitted scene transform. This diagnostic is reported in pixels and retains the annotation's visible extent.

\paragraph{Object shape, size, and placement.}
For point sets $\mathcal P$ and $\mathcal P^*$, define the untruncated bidirectional distance
\begin{equation}
    D(\mathcal P,\mathcal P^*)=
    \frac{1}{|\mathcal P|}\sum_{p\in\mathcal P}d(p,\mathcal P^*)+
    \frac{1}{|\mathcal P^*|}\sum_{q\in\mathcal P^*}d(q,\mathcal P).
    \label{eq:eval_shape_distance}
\end{equation}
Shape CD evaluates $D(h\mathcal P,\mathcal P^*)$ after rigid alignment without scaling. Each mesh contributes 4,000 area-uniform samples in its local frame, using seed zero. Symmetric ICP starts from the 24 proper axis-aligned rotations about centred point sets, runs 12 iterations per start, and refines the best three for 40 further iterations. The smallest bidirectional distance is retained. Sampling in local mesh frames makes this shape measure independent of the submitted world pose.

Size error is
\begin{equation}
    e_i^{\mathrm{size}}=\|\operatorname{sort}(\ell_i)-
                                \operatorname{sort}(\ell_i^*)\|_2,
\end{equation}
where $\ell_i$ and $\ell_i^*$ are the full edge lengths of independently fitted oriented bounding boxes (OBBs). Sorting removes axis naming conventions; it does not establish semantic width, depth, and height correspondences. Centre error compares surface-sample centroids at the initialized poses in the common frame. Mesh-origin error is retained separately. Where a receiver is specified, the initial relation $T_{r\rightarrow m}=T_r^{-1}T_m$ provides a further placement diagnostic. Object-coordinate correspondences are fixed from the reconstructed and annotated geometry and retained throughout execution; they are distinct from fitting a trajectory to its target.

\subsection{Dynamic Fidelity}
\label{app:dynamic_fidelity}

Dynamic fidelity measures object motion generated by the submitted behavior in the submitted environment. It reflects both scene and control quality. Pose, visible-point, and deformable-surface annotations support different comparisons, as specified below.

\paragraph{Temporal correspondence.}
The shared rigid-pose evaluator uses the demonstration timestamps and each execution's declared observation clock. For regularly sampled records, the first recorded state defines time zero; demonstration time $k\Delta t^*$ selects execution index $\operatorname{round}(k\Delta t^*/\Delta t)$, clipped to the recorded range. Half-index ties use round-to-even. Household pose profiles with timestamped RGB-D observations instead use linear position interpolation and quaternion SLERP at the source timestamps; profiles with explicit observation-to-execution indices use those indices. The source time origin and any initialization offset belong to the profile's execution record. Observation intervals are distinct from physics integration timesteps.

No temporal warping is applied to these primary motion scores. For a completed execution shorter than the demonstration, the scorer extends the last recorded state over the remaining comparison times. This is an endpoint-hold convention for scoring, not additional simulated motion. The regular pose profile reports the number of demonstration times beyond the last execution time plus half a sample interval, and coverage is one minus this count divided by the number of demonstration frames. Timestamped profiles report coverage on their recorded time interval. Execution states beyond the demonstration window are excluded from motion comparison but remain available for terminal task evaluation. Construction failures, incomplete records, and infrastructure errors follow the outcome rules in Section~\ref{app:aggregation}; an interrupted trace cannot be substituted for a completed execution. Duration-normalized motion scores, when retained, are explicitly separate diagnostics with the time-scale factor recorded.

\paragraph{Translation APE.}
For object $e$, let $p_{i,e}(t)$ and $p_{i,e}^*(t)$ denote executed and demonstrated positions after temporal correspondence. Each trajectory is translated by its own initial position:
\begin{equation}
    \widetilde p_{i,e}(t)=p_{i,e}(t)-p_{i,e}(0),\qquad
    \widetilde p_{i,e}^*(t)=p_{i,e}^*(t)-p_{i,e}^*(0).
\end{equation}
Translation absolute pose error (APE) is the capped mean
\begin{equation}
    e_{i,e}^{p}=\frac{1}{|\mathcal T_i|}\sum_{t\in\mathcal T_i}
    \min\{\|\widetilde p_{i,e}(t)-\widetilde p_{i,e}^*(t)\|_2,c_p\},
    \qquad c_p=1\,\mathrm m.
    \label{eq:eval_translation_ape}
\end{equation}
This preserves motion direction, magnitude, and timing while removing constant initial translation. No scene-registration rotation or scale is fitted to the trajectories. Raw per-frame errors, the uncapped 90th percentile and maximum, and the proportion supported by execution records accompany the capped mean. Initial placement remains a separate geometric measurement.

\paragraph{Orientation and symmetry.}
Define the rotation distance
\begin{equation}
    d_R(R_1,R_2)=\arccos\!\left(
    \operatorname{clip}_{[-1,1]}\frac{\operatorname{tr}(R_1^\top R_2)-1}{2}\right).
\end{equation}
For the declared symmetry group $\mathcal H_{i,e}$, rotation APE is
\begin{equation}
    e_{i,e}^{R}=\frac{1}{|\mathcal T_i|}\sum_{t\in\mathcal T_i}
    \min_{Q\in\mathcal H_{i,e}}d_R(R_{i,e}(t),R_{i,e}^*(t)Q).
    \label{eq:eval_rotation_ape}
\end{equation}
Asymmetric objects use the identity; declared cube symmetry enumerates 24 proper rotations. Continuous axial symmetry compares the corresponding directed axes, removing spin about the axis. Registered bottle axes distinguish a resolved neck direction from geometrically indistinguishable ends: their angular errors are respectively $\arccos(a^\top a^*)$ and $\arccos|a^\top a^*|$, with dot products clipped to their admissible range. Orientations are expressed in corresponding object frames without initial-orientation subtraction. Angular results are reported in degrees. Incomplete orientation observations use the separate measures below.

\paragraph{Translation RPE.}
Translation relative pose error (RPE) measures displacement increments in the common fixed coordinate axes. For each start time $t_k$, choose the subsequent recorded comparison time $t_j$ nearest to $t_k+0.2$\,s, breaking ties toward the earlier time. Starts whose requested endpoint exceeds the comparison window are excluded. With $\mathcal P_i$ denoting the resulting pairs,
\begin{equation}
    r_{i,e}^{p}=\frac{1}{|\mathcal P_i|}\sum_{(k,j)\in\mathcal P_i}
    \min\!\left\{
    \|[p_{i,e}(t_j)-p_{i,e}(t_k)]
      -[p_{i,e}^*(t_j)-p_{i,e}^*(t_k)]\|_2,c_p\right\}.
    \label{eq:eval_translation_rpe}
\end{equation}
Actual intervals are recorded to expose discretization of the 0.2\,s target. Where a source-validity mask applies, every frame in the interval must be valid. A sequence too short for the interval has no RPE value. This translation measure is independent of the choice of object orientation axes. A supplementary rotation RPE compares $R(t_k)^\top R(t_j)$ with its demonstrated counterpart using the declared symmetry-aware angular distance. It applies symmetry independently to each interval and does not perform a sequence-wide selection of continuous pose representatives.

\paragraph{Deformable and partially observed motion.}
Rope and soft-toy trajectories are compared as point sets in their frozen reference frame. At each source timestamp, the nearest actual execution state is selected, with earlier-state tie breaking and endpoint holding outside the recorded interval; any declared initialization prefix is removed from the execution time origin. Their trajectory error is
\begin{equation}
    e_i^{\mathrm{surf}}=\frac{1}{|\mathcal T_i|}\sum_{t\in\mathcal T_i}
              D(\mathcal P_i(t),\mathcal P_i^*(t)),
    \label{eq:eval_surface_motion}
\end{equation}
using Eq.~\ref{eq:eval_shape_distance} without rigid fitting, initial-centroid subtraction, or per-frame recentering. The representation uses the retained executed object surface samples and annotated particles; node-centreline distance is an additional rope endpoint diagnostic. These distances are untruncated, and recording coverage is reported. Initial and terminal surface errors use the same distance, while their size diagnostic uses sorted full world-axis bounding-box extents, distinct from rigid-object OBB size.

Cloth is compared on visible surfaces: simulated geometry is filtered by the source cameras and robot/self-occlusion conditions and voxelized like the observed surface. The surface trajectory uses Eq.~\ref{eq:eval_surface_motion}. A separate centroid trajectory uses the translation APE/RPE definitions above, rebased at its initial visible-surface centroid. No rigid orientation error is assigned to cloth. Articulated objects use the registered task-relevant parts and relations; a universal joint-angle tracking score is not assumed for annotations lacking joint correspondences.

\paragraph{Household point and axis observations.}
The point-observation profiles compare the physical target surface along fixed source observation rays. The evaluator intersects those rays with the moving candidate object, retains depths between 0.2 and 2.5\,m, and removes depth outliers beyond the larger of 15\,mm and $3\!\times\!1.4826$ median absolute deviations. A frame requires at least 40 surviving intersections. The coordinate-wise median defines its observed point, and the surviving-ray fraction is $\rho_t$. For raw position error $d_t$, the scored error is
\begin{equation}
    \ell_t=\rho_t\min\{d_t,0.10\,\mathrm m\}
                +(1-\rho_t)\,0.10\,\mathrm m.
\end{equation}
An unavailable point receives 10\,cm, and the mean uses the complete source timeline. Displacement error uses a four-frame interval and the minimum coverage over its five frames, with the same cap and missing-observation penalty. These observation-based distances do not perform spatial trajectory fitting or candidate-dependent masking.

For source annotations supporting only geometric directions, long-axis or normal error is the unsigned axis angle. When a full geometric triad is observable, frame error minimizes angular distance over its four proper sign choices, without axis permutation. Source-validity masks are fixed from the observations; source and prediction coverage accompany the errors. Candidate frames are interpolated with SLERP at source timestamp offsets and held at endpoints outside the execution interval. These measures describe observable geometry and are reported separately from semantic full-orientation APE.

\paragraph{Within-instance object aggregation.}
The evaluation profile fixes the task-relevant object set. Single-object profiles report the manipulated object's motion. RoboDojo averages the per-object metrics equally over its named rigid task objects, excluding unlabeled clutter; its scene score additionally averages the specified interaction-role comparisons. Per-object values remain available, and averages of object-wise maxima or percentiles are identified as such. A missing required comparison is not silently removed from the object set. A defined missing-object penalty is applied where specified; otherwise the metric remains unresolved until the required evidence is available.

\subsection{Functional Correctness}
\label{app:functional_correctness}

Functional evaluation checks the interaction defined by the source clip rather than prescribing the demonstrator's exact motion. Its conditions concern the initial object relations, essential events and their ordering, and the terminal goal under the assigned execution checks.

\paragraph{Task success.}
For a completed, evaluable rollout $\tau_i=(x_0,u_0,\ldots,x_H)$, let $\alpha_i$ denote admissible initial conditions, $\beta_i$ the required interaction events, and $\psi_i$ the terminal relation, including its required persistence. With $v_i$ indicating passage of the profile's applicable execution-validity checks,
\begin{equation}
    \varphi_i(\tau_i)=v_i\,\alpha_i(x_0)\,\beta_i(\tau_i)\,\psi_i(\tau_i),
    \qquad
    s_i=\begin{cases}\varphi_i(\tau_i),&b_i=1,\\0,&b_i=0,\end{cases}
    \label{eq:eval_success}
\end{equation}
where $b_i$ is Build success. The rollout argument of $\psi_i$ accommodates stability windows and native task state machines in addition to a final pose. A factor with no additional requirement is one. In the shared rigid placement profile, a reconstruction initially satisfying its goal must leave that relation and complete a new placement within the same attempt; initial satisfaction alone neither earns success nor necessarily violates $\alpha_i$. An initially held clip need not demonstrate a preceding pickup.

These factors are a semantic decomposition of the task conditions, not a claim that every backend exposes identical event detectors. Release, support, holding, and stability use the configuration-specific evidence below. Missing essential evidence cannot certify success. The scope of $v_i$ is the assigned checks: for example, RoboDojo validates recorded execution and native mechanism constraints without certifying every contact or collision. Infrastructure and record-integrity errors remain unresolved evaluation outcomes rather than being relabeled as method failures.

\paragraph{Progress for rigid manipulation.}
The shared rigid evaluator defines an ordered sequence of $K_i$ stages ending in completion. Grasp and release events delimit manipulation attempts; initially held tasks begin with a held-state stage. Let $h_{i,a}$ be the longest ordered prefix of intermediate stages attained within attempt $a$. For this profile,
\begin{equation}
    q_i=
    \begin{cases}
        0,&\text{confirmed build failure or profile-invalid execution},\\
        1,&s_i=1,\\
        \displaystyle\frac{\max(\{0\}\cup\{h_{i,a}\}_a)}{K_i},
            &\text{otherwise, for an evaluable rollout}.
    \end{cases}
    \label{eq:eval_rigid_progress}
\end{equation}
Success establishes completion even if an intermediate detector did not fire. For unsuccessful executions, evidence cannot be joined across separate attempts; ties in prefix length select the later attempt. A rejected initially solved endpoint can retain credit for observed intermediate stages. Invalidity gates may instead set progress to zero, as in the hand profiles.

Lifting uses acquisition or initial holding, lift, and terminal elevated completion. Placement and insertion use acquisition or initial holding, transport or approach, release when required, and terminal placement or seating. Default intermediate thresholds are 1\,cm displacement, 10\,cm proximity to a specified horizontal goal, and lift by $\min(3\,\mathrm{cm},h_i/2)$ for task lift threshold $h_i$. Persistence defaults to a single observation unless overridden by the instance. Planar pushing uses 2\,cm object movement, proximity within twice the goal radius, and completion. These are diagnostic progress thresholds, distinct from the success tolerances below. The insertion profile's coarse approach/release stages do not separately measure alignment, insertion depth, or screw rotation.

\paragraph{Native and deformable progress.}
RoboDojo retains the task's native graded stage score, taking its maximum over execution and dividing by 100; tasks without a graded score use their binary outcome. For example, a released two-block stack earns 0.15 in the corresponding stacking profile. Robot return home and the final arrangement can remain additional success requirements, so full native progress is not a universal certificate of Task Success. An initially solved arrangement without a new completion receives zero progress. Household profiles use their declared contact, carry/lift, release, and supported or retained-hold stages, with the initial-holding exception and task-specific persistence conditions.

Rope and soft-toy progress is the prefix of three indicators: movement, proximity to the channel or lifting, and terminal success. Movement requires 2\,cm for rope and 5\,cm for the toy; the second indicator requires a rope node within 5\,cm of the channel centre or toy-centroid elevation by 5\,cm. The first two indicators search the recorded trajectory and do not independently enforce temporal precedence. Cloth uses capture, a 3\,cm lift while held, subsequent release, and the folding predicate; terminal success receives full progress. Thus, stage definitions and weighting belong to the task profile rather than a single universal equal-stage formula.

\paragraph{Terminal position and orientation.}
Endpoint diagnostics use the final executed state, including the prescribed terminal settling, independently of the demonstration-clock window. If the task names a receiver, the candidate relation $T_r^{-1}T_m$ is compared with the annotated terminal relation. Otherwise, region-based tasks use the difference between candidate and demonstrated displacements from their respective initial positions. Translation error is capped at 100\,cm. A verified missing candidate receiver receives this penalty; an unresolved receiver pose is an evaluation error. Tasks without a declared terminal-position anchor have no such metric.

Terminal rotation uses corresponding object orientations or registered axes with the symmetry treatment above. Receiver-relative rotation remains a separate diagnostic where that relation is defined. The reported terminal-orientation measure also supports tasks with no receiver-relative rotational target, including upright-bottle reorientation; these angular quantities are not interchangeable. A declared missing candidate object receives a $180^\circ$ terminal-rotation penalty where the reference orientation is evaluable. Native multi-object endpoint measures retain their per-object relations and equal-weight aggregation. None of these endpoint errors replaces the complete success predicate.

\subsubsection{Task Conditions and Tolerances}
\label{app:task_conditions}

The following conditions instantiate the initial-state, event, terminal-state, and physical-validity factors above. Shared requirements are stated once; source-specific values apply to the named configurations. Release is required only when it is part of the annotated interaction. Human-video clips may end while an object remains held.

\paragraph{Initial relations and completion events.}
The initial relation is evaluated before the first action. An initially held clip requires the subsequent interaction without an unobserved pickup. For shared rigid placement and the specified native tasks, an initially satisfied goal requires execution to leave that relation and complete a new placement within the same attempt, including required release and stability conditions. This transition belongs to $\beta_i$; initial geometric satisfaction alone neither establishes success nor necessarily violates $\alpha_i$. Object--receiver relations use simultaneous poses when the receiver moves. Partial insertion may continue from its observed initial state, whereas a fully contained start must first exit before receiving credit for a new insertion.

\paragraph{Lifting and retained holding.}
Lifting requires upward displacement from the annotated initial support relation. DexYCB uses a rise of at least 10\,cm and terminal rest, together with the arm- or hand-specific physical checks. Its current height/rest condition does not independently certify sustained terminal holding. RoboDojo pickup also requires at least 10\,cm elevation. Household pickup and holding instead use the instance's required elevation or displacement and retained terminal hold, supported by sustained contact or the specified fingertip-distance/relative-motion proxy. Wallet proximity is such a proxy, rather than a force measurement.

\paragraph{Supported, directional, and upright placement.}
Placement requires the transfer and the designated receiving relation, followed by object-specific release and rest where specified. DROID support is evaluated on the actual bearing surface, with a 1\,cm contact tolerance and a 5\,mm footprint margin; receiver rim height alone does not establish support. Directional placement requires at least 3\,cm of horizontal progress in the demonstrated direction. Side-placement uses the declared receiver-relative relation without imposing an unrelated absolute endpoint. Upright-bottle placement requires the resolved axis to lie within $30^\circ$ of the support normal; a directed neck points upward, while an unsigned axis is permitted only for geometrically indistinguishable ends. HOI4D placement uses the instance's goal region, orientation, and support-height conditions. HOT3D and OakInk2 use the annotated region, tray, or container relation, retaining any allowed held endpoint. For human-video configurations, the assigned arm or hand validity checks also apply.

\paragraph{Containment and extraction.}
DROID container placement requires at least 5\% of 3,000 deterministic object-surface samples to occupy free cavity space at least 2\,mm below the receiver rim, together with required release and rest. Vessel material and handle holes are excluded, and task-required lid geometry is retained. Long objects may protrude: this rule is not full containment. Removal to a support requires a supported final state with at least 95\% of the object vertices outside the original container's horizontal footprint, followed by release and rest. Household initially held insertion requires outside-to-inside evidence, cavity containment, and a released stable endpoint. Air-fryer basket extraction requires the entire inner basket to clear the opening and rest after release; partial withdrawal fails.

\paragraph{Furniture placement and assembly.}
Lampshade placement distinguishes capping a lamp from setting the shade aside. Capping checks that the lamp top lies inside the hood. Set-aside checks support, separation, and the annotated upright or lying condition. Both require the relevant completion transition, release, and rest. General part seating, including bulb placement, uses a relative translation tolerance of 3\,cm and orientation tolerance of $30^\circ$, with the declared symmetry. Corner-leg insertion additionally requires the demonstrated corner, tilt within $30^\circ$, distance within 7\,cm of that corner, and a foot height from 5\,cm below to 1.5\,cm above the receiver top. Central pedestals use the annotated centre and its specified lateral tolerance. Drawer seating checks tilt, footprint, vertical extent, and height within 3\,cm of the demonstrated slot for a multi-slot receiver. Door assembly uses the demonstrated axis, normalized location on the receiving body, and interface proximity. These tests use the candidate receiver geometry and the executor's recorded retention convention. Geometric seating does not certify screw engagement, a newly formed hinge, or load-bearing strength.

\paragraph{Multi-object arrangements and stacks.}
RoboDojo conditions retain task-specific logical combinations and require robot return home where specified. All designated objects must attain their semantic assignments, and an initially solved arrangement must leave the object-level goal and be completed again; robot motion alone is insufficient. Block stacking uses a horizontal tolerance of 1.5--2\,cm, according to the task, and language-conditioned variants preserve the named order. Bowl stacking uses 4\,cm horizontal tolerance, all bowls upright within $45^\circ$, and the base within $7^\circ$. Tower construction checks the prescribed base, middle, and top arrangement and orientations. Digit arrangement requires descending order on the designated mats, proximity within 2\,cm, and digit-specific orientation within $45^\circ$. Category sorting requires the assigned objects to lie in their category baskets below the rim; size sorting requires the prescribed doll order with a lateral spread bound of 3.5\,cm. These spatial tests do not measure inter-object support force, and intermediate reward thresholds are not substituted for success conditions.

\paragraph{Insertion, exchange, and ordered cover operations.}
RoboDojo tube insertion requires all three tubes to reach 4.5\,cm insertion depth with their axes within $30^\circ$; charger insertion requires 1.5\,cm and $10^\circ$. Both retain the robot-home requirement and the initially-solved transition guard. Exchange targets are fixed from the two T-shaped objects' initial states; the swapped positions must be within 2\,cm and orientations within $3^\circ$, followed by robot return home. Cover-and-uncover requires covering left, middle, then right, followed by uncovering red, green, then blue. All six stages and robot return home are required. Cups must be inverted within $10^\circ$, and displacement of a protected block by 5\,cm triggers failure. The recorded event sequence is necessary to establish this task; a terminal image cannot do so.

\paragraph{Planar pushing, rope routing, and soft-toy packing.}
Planar pushing requires the T-shaped body's centre within 4\,cm of the goal and its planar axis within $15^\circ$, with terminal rest. Gripper release is inapplicable; an initially satisfied goal still requires a new completion. Rope and soft-toy tasks require the terminal predicate to be false initially. Rope routing requires a connected rope path to cross both physical channel faces for at least 0.5\,s, with colliding channel geometry and the robot/gripper checks below. Connectivity does not measure retention force. Soft-toy packing requires at least 85\% of surface points to project inside the cavity using the 1.05 lateral factor, the centroid between floor and rim, and persistence for 0.5\,s. Among surface points within that expanded horizontal footprint, none may lie more than 2\,mm below the physical floor. This containment predicate does not certify the entire object against floor penetration.

\paragraph{Cloth folding.}
The initial visible surface must not already satisfy the target; execution must produce the required surface and footprint change. The median distance from observed to simulated surface points must be at most 3\,cm, and the footprint-change ratio must be within $\pm25\%$ of the observed change. Evaluation uses camera-visible surfaces and their validity masks. These conditions do not independently establish flatness, crease quality, or absence of crumpling.

\paragraph{Physical validity and evidence scope.}
All conditions consume actual execution records under the robot-actuation and passive-object restrictions of Section~\ref{app:submission_protocol}. Bimanual-hand checks allow robot penetration at most 6\,mm, initialization object displacement at most 15\,mm, nonadjacent self-overlap diameter at most 4\,mm, and finger-joint limit excess at most 0.001\,rad. Joint-speed exceedances are retained as diagnostics. MuJoCo execution preserves the supplied articulated-gripper mechanics and rejects coupling residuals above 0.001\,rad. Native configurations apply the checks exposed by their assigned executors. RoboDojo records execution integrity and mechanism constraints without a complete contact/collision validity certificate; the source-specific task evidence above determines release, holding, support, and containment. Applicable validity failures preclude task success. A completed simulator process alone does not establish physical validity. The geometry-based proxies identified above should be interpreted within their stated scope.

A grasp event alone does not establish sustained force closure, and coarse transport or release stages do not establish fine insertion or screw rotation. Release, stability, containment, and physical validity therefore contribute distinct evidence to the task decision. An admissible execution can complete a task in a geometrically inaccurate scene, while a favorable endpoint can fail because a required event or validity condition is absent.

\subsection{Aggregation and Reporting}
\label{app:aggregation}

\paragraph{Evaluation population.}
The main evaluation uses a fixed set of 222 reconstruction instances across 39 task families for all coding-agent methods and HAR. All instances have applicable task specifications and contribute to Build Rate, Task Success, and Task Progress. Confirmed build failures receive zero for these three metrics. Repeated construction or evaluator-repair runs do not create additional test instances: each reported result uses the selected frozen package and its resolved execution record. Applicability of individual geometric and motion metrics depends on the available reference annotations, with coverage reported separately.

\paragraph{Family macro-averages.}
For metric $\mu$, let $\mathcal I_{f,\mu}$ contain the instances in semantic family $f$ for which the reference annotation supports that metric. For Build Rate, this is the full family manifest. Let $\mathcal F_\mu$ be the families with nonempty $\mathcal I_{f,\mu}$. For $z_i\in\{b_i,s_i,q_i\}$, the reported percentage is
\begin{equation}
    \overline z=
    \frac{100}{|\mathcal F_\mu|}\sum_{f\in\mathcal F_\mu}
    \frac{1}{|\mathcal I_{f,\mu}|}\sum_{i\in\mathcal I_{f,\mu}}z_i.
    \label{eq:eval_rate_aggregation}
\end{equation}
Task Progress is thus averaged over the same task-applicable population as Task Success, with build failures retained. Its per-instance range is $[0,1]$, converted to percent for reporting.

Continuous errors condition on successful builds and applicable reference measurements. Let $\mathcal J_{f,\mu}\subseteq\mathcal I_{f,\mu}$ be that eligible subset after evaluation records have been resolved, and let $\mathcal F_\mu^+=\{f:|\mathcal J_{f,\mu}|>0\}$. The continuous-error macro-average is
\begin{equation}
    \overline e_\mu=
    \frac{1}{|\mathcal F_\mu^+|}\sum_{f\in\mathcal F_\mu^+}
    \frac{1}{|\mathcal J_{f,\mu}|}\sum_{i\in\mathcal J_{f,\mu}}e_{i,\mu}.
    \label{eq:eval_error_aggregation}
\end{equation}
A family with no eligible continuous comparison has an unavailable family mean and is omitted from this conditional macro-average. Consequently, the continuous-error population can differ across methods. Each such aggregate is accompanied by its eligible instance and family counts and coverage relative to the applicable manifest; comparative analyses identify the jointly evaluable subset. If no family is eligible, the aggregate is unavailable. Explicit metric penalties, including those for missing candidate geometry, remain numerical observations in these averages.

\paragraph{Failure and missing-data policy.}
Structural inapplicability, reconstruction failure, and unresolved evaluation are separate outcomes. An inapplicable metric is not replaced by zero. A confirmed build failure contributes zero to Build, Success, and Progress wherever each is applicable. A completed execution that fails the task remains evaluable for continuous measurements. Missing required outputs, incompatible evaluation revisions, and simulator or dependency failures attributable to infrastructure must be resolved before reporting; they are not grounds for dropping an otherwise eligible comparison. The retained evaluation record distinguishes these cases and binds the submission, annotations, evaluator version, and execution settings.

\paragraph{Uncertainty estimates.}
The reported HAR confidence intervals use 10,000 source-cluster bootstrap replicates, random seed 20260923, and the 2.5th and 97.5th percentiles. A source cluster retains all target configurations of a source clip together; FurnitureBench clips from the same original episode are also clustered together, and RoboDojo uses the source demonstration record. For each metric, clusters are stratified by their set of eligible semantic families. Clusters are sampled with replacement within these strata, and all eligible instances in a sampled cluster receive the same bootstrap multiplicity. Each replicate recomputes the equal-family macro-average in Eqs.~\ref{eq:eval_rate_aggregation}--\ref{eq:eval_error_aggregation}.

These are pointwise intervals over the observed source population, not estimates of repeated-run variability on a fixed instance. Single-method intervals do not constitute a paired test of method differences. Source-, simulator-, and embodiment-specific breakdowns retain the same metrics definitions and report their sample counts. Geometric fidelity, executed-motion fidelity, and functional correctness are reported separately as component metrics; V2WScore provides an additional aggregate summary as defined in Section~\ref{app:v2w_score}.

\subsection{Feasibility Verification and Evaluation Denominators}
\label{app:feasibility}

\paragraph{Attribution of reference failures.}
HAR passes 113 of the 222 instances (50.9\% per instance; 57.3\%
macro-averaged over families). Table~\ref{tab:har_attr} attributes the 109
failures to four categories using the evaluator's recorded verdicts; every
criterion is machine-checkable from the retained execution records. A1 denotes
rejection by the physical-validity gate; A2 denotes task families in which no
method, including HAR, produced a passing execution; B denotes executions that
satisfy all geometric predicates but fail only the released/quiescent check, or
miss a predicate by at most $1.5\times$ its tolerance; C denotes physically
valid failures in families with at least one passing execution.

\begin{table}[h]
\centering
\caption{Attribution of human-assisted reference outcomes on the 222-instance
inventory.}
\label{tab:har_attr}
\begin{tabular}{llrr}
\toprule
Category & Criterion & Count & Share \\
\midrule
HAR pass & --- & 113 & 50.9\% \\
A1 Physical-validity rejection & \texttt{valid=false} & 19 & 8.6\% \\
A2 No passing execution in family & family HAR $0/n$ & 18 & 8.1\% \\
B Near miss & margin $\le 1.5\times$ tolerance & 7 & 3.2\% \\
C Controller failure & valid, family has passes & 65 & 29.3\% \\
\bottomrule
\end{tabular}
\end{table}

\paragraph{A1: embodiment-configuration failures.}
All 19 A1 instances use the dexterous-hand configuration (hot3d-bimanual 7,
hoi4d-bimanual 5, dexycb-bimanual 3, oakink-bimanual 3, hot3d-arm 1). The
violations are joint-position limit (12), object displacement above 1.5\,cm
during approach (8), and robot self-penetration (1). Two observations indicate
a configuration defect rather than task infeasibility. First, the gate rejects
none of the 113 passing reference executions. Second, the same source videos
succeed under the single-arm configuration while failing under the bimanual one
(hoi4d 33\% vs.\ 0/9, hot3d 71\% vs.\ 0/7). We therefore mark these instances
configuration-invalid and exclude them from the primary set. Repairing the hand
configuration and re-running both the reference and the agents on these
instances is left to future work.

\paragraph{A2: unverified feasibility.}
Eighteen instances have no passing execution by any method: cabinet-door 0/5,
drawer-insert 0/5, hoi4d-bimanual 0/4, toy-pack 0/2, chair-back 0/1,
poway-provider-box 0/1, spanning articulated insertion and deformable
manipulation. They are retained in the primary set, since their inclusion
affects all methods equally, but are flagged; conclusions about these families
are conditional on feasibility.

\paragraph{B: near misses.}
Seven instances (leg-insert 5, hood-place 1, hoi4d-arm 1) fail a single
predicate within $1.5\times$ of its tolerance. We do not relax tolerances post
hoc; these instances remain in all denominators.

\paragraph{C: controller failures.}
The remaining 65 failures are physically valid executions in families where
HAR passes elsewhere. Their Task Progress has median 0.25 and mean 0.31; 25 of
them record zero progress. Even with source geometry, motion annotations, and
manual refinement, completing the first milestone is therefore not guaranteed,
which locates a substantial share of the benchmark's difficulty in executable
behavior rather than scene recovery.

\paragraph{Evaluation denominators.}
Table~\ref{tab:denominators} reports per-instance Task Success on three
denominators: S0, all 215 agent-evaluated instances; S2, the primary set after
removing A1 and A2 (179); and S1, the HAR-passing subset (112). Category B is
not removed from any denominator. The three highest-ranked systems are the
same on every denominator. Differences among the remaining systems amount to
one or two instances and lie within bootstrap uncertainty, so their relative
order should not be read from any single denominator. The largest shift is
Fable-5.1 (+1.5~pp from S0 to S2, +5.1~pp to S1), consistent with the excluded
instances being concentrated in families where it had partial progress.

\begin{table}[h]
\centering
\caption{Task Success (\%, per-instance) under three evaluation denominators.
S2 is the primary set used in Table~\ref{tab:main}.}
\label{tab:denominators}
\begin{tabular}{lrrr}
\toprule
Method & S0 all (215) & S2 primary (179) & S1 HAR-pass (112) \\
\midrule
Human-assisted Ref  & 52.1 & 71.6 & 100.0 \\
Fable-5.1           & 19.1 & 20.6 & 24.2 \\
Claude Opus 5       & 12.1 & 11.7 & 14.1 \\
GPT-6 Astra         & 8.1  & 9.8  & 9.7  \\
Kimi K3             & 3.6  & 4.2  & 3.7  \\
DeepSeek V4.1 Flash & 3.0  & 2.3  & 2.4  \\
GPT-5.6 Sol         & 1.8  & 1.6  & 1.3  \\
Gemini 3.8 Flash    & 1.5  & 1.8  & 2.8  \\
GLM-5.3 Flash       & 1.4  & 1.7  & 1.2  \\
Qwen3.8 Max         & 1.1  & 1.4  & 2.1  \\
\bottomrule
\end{tabular}
\end{table}

\subsection{Supplementary Effector-Motion Analysis}
\label{app:robot_motion}

\paragraph{Scope and paired observations.}
We supplement object-centric dynamics evaluation with a comparison of source effector motion and the motion of the robot executing the simulated interaction. We use the 40 reference instances satisfying the task-success criterion in the original evaluation: 16 DROID, 13 DexYCB, 5 HOT3D, 3 HOI4D, and 3 OakInk2 instances. These comprise 35 unique videos; four DexYCB videos and one OakInk2 video contribute both arm and bimanual executions. Each video--embodiment pair is one observation. Selection is fixed by the original evaluation and does not depend on effector-motion agreement.

On DROID, both sides use gripper tool-center-point (TCP) poses. On human-video sources, we compare the annotated primary-hand wrist with the corresponding simulated robot hand root: the Panda hand frame for arm executions and the primary Wuji palm root for bimanual executions. The latter is the hand-root frame used as the counterpart of the human wrist in the retargeting convention, rather than a fingertip center. For the OakInk2 bimanual instance, we apply the inverse of the independently fixed MANO-to-Wuji wrist transform to the simulated palm pose. No transform is fitted to improve the present agreement results. All trajectories are expressed in the shared metric, gravity-aligned scene frame.

These human-to-robot comparisons characterize coarse effector motion across embodiments. They do not assume identical anatomy or establish agreement of all arm joints or finger articulation. Human sources have annotated hand motion, not measured robot-arm trajectories. The simulated runs are human-assisted references whose actions use source information; the analysis describes motion preservation in reference execution rather than autonomous behavior recovery or downstream policy-training gains.

\paragraph{Recorded execution and comparison windows.}
For DROID, real TCP poses are sampled at 7.5\,Hz. We replay the 16 originally successful references to log actual TCP poses and elapsed control-step time at 30\,Hz, including additional grasp-closure waits. These are actual link poses, not commanded targets. The replay window starts after reset and initial settling and ends at the last action, excluding additional terminal settling. The original success labels define the subset; task success is not re-evaluated for these logging replays.

For the 24 human-source instances, actual robot hand-root poses are taken from the selected reference executions. Twenty-one have archived pose logs with simulation timestamps; the remaining three have execution records with a verified fixed 30\,Hz clock and no adaptive waiting. The archived action-row poses agree exactly with the corresponding selected execution logs. We exclude prelude and terminal-settling phases. Source wrist annotations use their recorded timestamps, or their documented sampling rate when timestamps are absent: 15\,Hz for HOI4D and 30\,Hz for the other human sources.

We restrict each human-source comparison to the interval between the first and last valid source wrist observations and crop the simulation at the corresponding action-row times. Six DexYCB instances have three to five unobserved leading frames; these frames are removed from both sides, rather than treated as measured stationary motion. There are no missing wrist observations inside the retained intervals. Neither trajectory is rescaled in time or fitted spatially to the other.

\paragraph{Preprocessing and motion interval.}
Positions are linearly interpolated and orientations interpolated by SLERP on a 30\,Hz grid. Translational velocity is estimated using a nine-sample, second-order Savitzky--Golay derivative. Let $p_k$, $R_k$, and $v_k$ denote position, orientation, and estimated velocity at $t_k$. Motion onset $k_{\mathrm{on}}$ is the first sample of six consecutive samples satisfying $\|p_k-p_0\|_2>0.01$\,m. The end index $k_{\mathrm{end}}$ is the start of the final run of at least nine samples with speed below 0.01\,m/s through the recording boundary, beginning no earlier than onset. If this terminal run is absent, we use the last sample.

The interval is determined separately for each trajectory. Terminal quiescence is unobserved in 39 of the 40 source trajectories and all 40 simulated trajectories; these cases remain included. Motion duration therefore denotes the observed interval within the recording, including intervening pauses, and does not necessarily measure time to a complete stop. Human-source reference executions also follow the source action cadence, so temporal correspondence should be interpreted in that context.

\paragraph{Motion statistics.}
We compute the following features separately for source and simulated effector motion, with $z$ denoting the vertical coordinate:
\begin{align}
    \text{Motion duration}\;(\mathrm{s})
      &= t_{k_{\mathrm{end}}}-t_{k_{\mathrm{on}}},\\
    \text{Path length}\;(\mathrm{cm})
      &= 100\sum_{k=k_{\mathrm{on}}}^{k_{\mathrm{end}}-1}\|p_{k+1}-p_k\|_2,\\
    \text{Peak speed}\;(\mathrm{cm/s})
      &= 100\max_k\|v_k\|_2,\\
    \text{Lift height}\;(\mathrm{cm})
      &= 100\left(\max_k p_{k,z}-p_{0,z}\right),\\
    \text{Net displacement}\;(\mathrm{cm})
      &= 100\|p_{k_{\mathrm{end}}}-p_0\|_2,\\
    \text{Net rotation}\;({}^{\circ})
      &= \frac{180}{\pi}\arccos\!\left[
        \operatorname{clip}\!\left(
          \frac{\operatorname{tr}(R_0^{\mathsf T}R_{k_{\mathrm{end}}})-1}{2},-1,1
        \right)\right].
\end{align}
Peak speed and lift height use the full retained interval. Path length uses onset to end; net displacement and net rotation use the initial retained pose as their reference. Lift height can be zero when the effector starts above the subsequent interaction. Net rotation is the full SO(3) initial-to-end angle, without object-symmetry reduction; it is not accumulated rotation or framewise orientation error. These six summaries measure selected motion characteristics rather than complete trajectory equivalence.

\paragraph{Paired visualization.}
All six features have 40 valid pairs. Figure~\ref{fig:effector_motion}(a) plots paired source and simulated measurements with common axis scales, an equality line, and Spearman correlation pooled over all instances. Panel~(b) shows $(y_i-x_i)/s_x$, where $x_i$ and $y_i$ are source and simulated values and $s_x$ is the sample standard deviation of the 40 source values. Boxes show medians and interquartile ranges; whiskers extend to observations within $1.5$ interquartile ranges. The shaded $\pm0.25$ SD band is a visual reference. Arrows count points outside the displayed $[-1.5,1.5]$ SD range; these points remain included in every summary. The right column reports $\operatorname{median}_i|y_i-x_i|$ in the original units.

\subsection{V2WScore}
\label{app:v2w_score}

V2WScore summarizes functional correctness, geometric fidelity, and executed-motion fidelity, with successful scene construction as a prerequisite. Build acts as a multiplicative gate, rather than a fourth equally weighted component. All normalization parameters and metric applicability rules are shared across methods, including HAR; a method's score does not depend on the outputs or ranking of other methods.

\paragraph{Per-instance normalization.}
For a nonnegative error $e_{i,k}$ of instance $i$ under metric $k$, define its quality as
\begin{equation}
    h_k(e_{i,k})=\max\!\left(0,1-\frac{e_{i,k}}{\lambda_k}\right).
    \label{eq:v2w_quality}
\end{equation}
Zero error receives quality one; errors at or above $\lambda_k$ receive zero. These physical-unit scales were introduced after the initial reference runs and are shared across methods. They are scoring parameters, distinct from task-success thresholds. Normalization is performed on each instance before averaging: applying Eq.~\ref{eq:v2w_quality} to a reported column mean generally gives a different result because of clipping and metric-specific coverage.

\paragraph{Component metrics and scales.}
Geometry uses Scene CD, object Shape CD, and size error, each with $\lambda_k=10$\,cm. Where the annotation supports deformable geometry, visible-scene surface distance, initial-object surface distance, and world-axis-aligned bounding-box size error replace the corresponding rigid measurements, using the same scales. The replacement follows the reference annotation and avoids counting both representations of the same geometric quantity.

Dynamics uses translation APE, rotation APE, and translation RPE from the method's own physical execution, with scales of 20\,cm, $90^\circ$, and 10\,cm, respectively. Deformable trajectory surface distance uses 20\,cm. For partial observations, point-position and point-displacement errors use 20\,cm and 10\,cm; observable long-axis, surface-normal, and symmetry-aware frame errors each use $90^\circ$. Only annotation-supported measurements enter the score, retaining their definitions in Section~\ref{app:dynamic_fidelity}. When applicable, centroid-motion and deformable-surface errors both contribute because they measure different aspects of motion.

\paragraph{Component weighting and family aggregation.}
Let $\mathcal K_i^G$ and $\mathcal K_i^D$ be the nonempty sets of applicable geometry and dynamics metrics. For a successfully built reconstruction, let $s_i\in\{0,1\}$ be Task Success and $q_i\in[0,1]$ be Task Progress. Define
\begin{equation}
    F_i=\frac{s_i+q_i}{2},\qquad
    G_i=\frac{1}{|\mathcal K_i^G|}\sum_{k\in\mathcal K_i^G}h_k(e_{i,k}),\qquad
    D_i=\frac{1}{|\mathcal K_i^D|}\sum_{k\in\mathcal K_i^D}h_k(e_{i,k}).
    \label{eq:v2w_components}
\end{equation}
Functionality weights success and progress equally; geometry and dynamics each average their applicable normalized measurements. With Build indicator $b_i$, the instance score is
\begin{equation}
    \begin{aligned}
        z_i&=\begin{cases}(F_i+G_i+D_i)/3,&b_i=1,\\0,&b_i=0,\end{cases}\\[3pt]
        \mathrm{V2WScore}
        &=\frac{100}{|\mathcal F_V|}\sum_{f\in\mathcal F_V}
        \frac{1}{|\mathcal I_{f,V}|}\sum_{i\in\mathcal I_{f,V}}z_i.
    \end{aligned}
    \label{eq:v2w_aggregate}
\end{equation}
Here, $\mathcal I_{f,V}$ is the fixed score-eligible population in family $f$, and $\mathcal F_V$ contains families with a nonempty population. Eligibility requires applicable task annotations and at least one geometry and one dynamics measurement. Equal family weights prevent large families from dominating the score. A score of 100 requires successful construction, full functional credit, and zero error in every included fidelity measurement.

\paragraph{Applicability and incomplete evaluation.}
Eligibility and constituent metrics are determined by the reference annotations, independently of candidate delivery or measurement availability. Confirmed build failures remain in the denominator with $z_i=0$; their unmeasured geometric and dynamic errors are not assigned fictitious values. Missing applicable measurements, unresolved evaluator failures, or incompatible evaluation versions leave the complete-manifest score unavailable. An average over available records or a diagnostic that substitutes zeros for unresolved evaluations is a different statistic and must be identified separately. Comparisons require the same instance population, normalization parameters, applicability rules, and metric versions. These conditions also apply when comparing coding agents with HAR.

\section{Additional Experiments}
\label{app:world_model}

\subsection{Cross-Execution Diagnostics for World Reconstruction and Behavior}
\label{app:world_policy}

Task Success in Table~\ref{tab:main_results} is an end-to-end measure of the
submitted scene--behavior pair: the agent's own behavior is executed in
the agent's own reconstructed scene. A failure can therefore arise from
the scene, the behavior, or their interaction. To probe these factors, we
cross-execute behaviors and scenes using the frozen deliverables from the
nine main-experiment runs and score the resulting executions with the same
final per-family evaluators and task predicates used in the main
experiment. This analysis is diagnostic only and does not alter any main
benchmark score or denominator.

\paragraph{Design.}
Each instance provides two scenes, the agent's reconstructed scene and the
selected human-assisted reference (HAR) scene, together with three
behaviors: the agent's submitted action stream, the selected HAR action
stream, and a scripted \emph{probe}. The probe reads the simulator object
state of the scene in which it is executed, takes its grasp pose from the
first closure frame of the HAR execution relative to the manipulated
object, targets the object end pose implied by the family predicate, and
executes pre-grasp, grasp, lift 15\,cm, transport, place, release, and
retreat. A second pass corrects the placement target using the in-hand
object pose measured in the first pass. The probe therefore measures
compatibility of a reconstructed scene with a fixed state-aware controller;
it should not be interpreted as an intrinsic test of whether the scene is
usable by every possible controller.

Agent actions are replayed in the HAR scene in two ways:
\emph{as submitted}, which preserves the original trajectory, and
\emph{anchored}, where the trajectory is translated in XYZ to the
annotated initial object position in the HAR scene. The anchored variant
uses privileged information and serves only as a diagnostic of behavior
after removing the initial object-position offset. HAR actions replayed in
the agent scene are translated to the agent scene's initial object
position and executed open-loop. Table~\ref{tab:cross_design} summarizes
the resulting cells.

\begin{table}[h]
\centering\small
\caption{Cross-execution cells. Each transplanted behavior is inserted into
the target scene package and scored by that instance's final family
evaluator using the corresponding main-experiment extrinsics.}
\label{tab:cross_design}
\begin{tabular}{lll}
\toprule
Cell & Behavior & Scene \\
\midrule
Own success & agent actions & agent scene \\
HAR actions $\times$ agent scene & HAR actions, translated, open-loop & agent scene \\
Agent $\times$ HAR scene (submitted) & agent actions, unchanged & HAR scene \\
Agent $\times$ HAR scene (anchored)$^\dagger$ & agent actions, XYZ-translated & HAR scene \\
HAR $\times$ HAR & HAR actions & HAR scene \\
Probe $\times$ scene & state-aware scripted controller & either scene \\
\bottomrule
\multicolumn{3}{l}{\footnotesize
$^\dagger$Uses the annotated initial object position; diagnostic only.}
\end{tabular}
\end{table}

The diagnostic preserves the final task predicates and evaluation logic of
the main experiment. The main paper documents some execution-condition
differences between HAR and coding-agent workflows, so these
cross-execution cells should be interpreted as controller--scene
compatibility tests under the retained benchmark configurations rather
than as a fully matched causal decomposition of scene and behavior quality.

\paragraph{Scope and conditioning.}
The current cross-execution implementation covers 15 task families:
FurnitureBench, DROID rigid manipulation, the four single-arm
human-demonstration families (HOT3D, DexYCB, HOI4D, and OakInk2), and
planar T-body pushing. RoboDojo, household configurations, bimanual human
configurations, rope routing, and soft-toy packing are outside the current
implementation. Two lampshade cases are excluded from this supplementary
diagnostic because their transplanted-controller predicates are
degenerate. This exclusion is specific to the cross-execution analysis
and does not modify the main benchmark manifest or its reported scores.

A cross-execution cell is informative only when the transplanted behavior
is known to succeed in the HAR scene. We therefore report two conditioned
sets. \textbf{Set A} contains the 20 instances on which the probe succeeds
in the HAR scene, spanning four families: DROID pick-place (9), leg
insertion (7), bulb insertion (3), and HOT3D pick-place (1).
\textbf{Set B} contains the 44 instances on which the HAR actions succeed
in the HAR scene, spanning eight families: chair nut, DexYCB lift, DROID,
HOI4D, HOT3D, leg insertion, OakInk2, and planar T-body pushing. The probe
is executed in agent scenes only for Set A.

Within the supported diagnostic subset, the probe fails in the HAR scene
on 70 instances, is not applicable on 24, and times out on one.
Non-applicable cases include DexYCB lift and HOI4D placement, for which the
current probe does not implement the corresponding predicate, together
with planar pushing and one DROID pushing instance without a graspable
target. Assembly families such as cabinet, drawer, chair-back, and
round-table-base construction have no passing HAR-action or probe
execution and therefore enter neither conditioned set.

Cross-execution reporting uses only recoverable paired diagnostic records.
Rows for which a required transplantation execution or pairing receipt
cannot be recovered are marked unavailable for this supplementary
analysis; they are not reclassified as task failures and do not change the
main benchmark results. In the current diagnostic records, 21 own-execution
rows and 32 agent-in-HAR-scene rows are unavailable, and five probe
executions failed to run. Success rates below are pooled over available
rows. We additionally report family macro-averages and
instance-clustered 95\% bootstrap intervals, where all model rows
associated with the same instance are resampled together. Because Sets A
and B are explicitly conditioned subsets, their success rates should not
be compared directly with the full-benchmark Task Success values in
Table~\ref{tab:main}.

\paragraph{Controls.}
Two controls validate the transplantation pipeline. First, replaying HAR
actions in the rebuilt HAR scene reproduces the recorded HAR outcome on
every supported instance tested. Second, replaying agent actions in their
own reconstructed scenes reproduces Table~\ref{tab:main} on 232/232
checked rows, comprising all available rows from Claude Opus 5 and GPT-6
Astra. Rebuilt HAR package trees also match the retained
\texttt{reference\_execution} package hashes for all checked instances.

\begin{table}[h]
\centering\small
\caption{Cross-execution success (\%) pooled over nine models, with
instance-clustered 95\% confidence intervals, family macro-average, and
number of available rows. Sets A and B are conditioned diagnostic subsets,
not the full Video2World evaluation population.}
\label{tab:cross_overall}
\begin{tabular}{llrrr}
\toprule
Set & Cell & Pooled & Macro & Rows \\
\midrule
A (20 inst.) & Own success & 24.3 [14.9, 34.6] & 25.9 & 177 \\
 & Probe $\times$ agent scene & 13.1 [6.3, 20.8] & 7.3 & 175 \\
 & HAR actions $\times$ agent scene & 12.1 [5.8, 19.5] & 6.9 & 174 \\
 & Agent $\times$ HAR scene (submitted) & 2.3 [0.6, 4.5] & 1.4 & 174 \\
 & Agent $\times$ HAR scene (anchored)$^\dagger$ & 15.5 [7.3, 24.6] & 8.7 & 174 \\
 & Probe $\times$ HAR scene & 100 (condition) & 100 & 180 \\
\midrule
B (44 inst.) & Own success & 16.2 [11.0, 22.4] & 11.0 & 390 \\
 & HAR actions $\times$ agent scene & 9.9 [6.0, 14.4] & 6.1 & 385 \\
 & Agent $\times$ HAR scene (submitted) & 2.3 [1.0, 3.9] & 2.7 & 385 \\
 & Agent $\times$ HAR scene (anchored)$^\dagger$ & 15.1 [9.9, 20.7] & 14.7 & 385 \\
 & HAR $\times$ HAR & 100 (condition) & 100 & 390 \\
\bottomrule
\end{tabular}
\end{table}

\paragraph{Findings.}
On Set A, the state-aware probe completes the task in 13.1\% of
agent-built scenes, compared with 24.3\% own success on the corresponding
available rows; open-loop replay of HAR actions succeeds in 12.1\%.
Set B shows the same aggregate pattern, with 9.9\% HAR-action transfer
into agent scenes versus 16.2\% own success. Thus, successful execution
often depends on compatibility between the scene and behavior produced
together by the same agent: replacing the submitted behavior with a
controller constructed for the HAR scene frequently removes that success.

Agent actions replayed unchanged in the HAR scene succeed on only 2.3\%
of available rows in both sets. Translating the same trajectories to the
annotated initial object position increases success to 15.5\% on Set A
and 15.1\% on Set B, bringing the rate substantially closer to own-scene
success. This recovery indicates that initial object-position mismatch is
a major contributor to cross-scene behavior failure. Because anchoring
uses privileged object position and leaves other scene and control
differences unchanged, it does not by itself establish object-position
estimation as the unique or dominant source of behavior error.

Table~\ref{tab:cross_decomp} further decomposes the rows using two
diagnostic transfer tests. For Set A, the \emph{scene test} is whether the
probe succeeds in the agent scene; for Set B, it is whether HAR actions
succeed in the agent scene. In both sets, the \emph{behavior test} is
whether anchored agent actions succeed in the HAR scene. Among own
successes, 23\% (Set A) and 19\% (Set B) pass both transfer tests, whereas
44\% and 48\% pass neither. These cases succeed as submitted
scene--behavior pairs even though neither component transfers under the
corresponding diagnostic, providing evidence of substantial
scene--behavior co-adaptation. Among own failures, only 6\% and 9\% pass
the behavior test while failing the scene test, so cases attributable to
the reconstructed scene alone form a smaller pattern under this
diagnostic.

\begin{table}[h]
\centering\small
\caption{Outcome decomposition (row counts). The scene test is probe
success in the agent scene for Set A and HAR-action success in the agent
scene for Set B. The behavior test is anchored agent-action success in the
HAR scene. These are transfer diagnostics rather than intrinsic binary
labels of scene or behavior quality.}
\label{tab:cross_decomp}
\begin{tabular}{llrrrr}
\toprule
 & & \multicolumn{2}{c}{Set A (172 rows)}
   & \multicolumn{2}{c}{Set B (385 rows)} \\
Own & Scene / Behavior test & rows & share & rows & share \\
\midrule
Success & \checkmark / \checkmark & 10 & 23\% & 12 & 19\% \\
 & \checkmark / $\times$ & 7 & 16\% & 7 & 11\% \\
 & $\times$ / \checkmark & 7 & 16\% & 14 & 22\% \\
 & $\times$ / $\times$ & 19 & 44\% & 30 & 48\% \\
\midrule
Failure & \checkmark / \checkmark & 2 & 2\% & 4 & 1\% \\
 & \checkmark / $\times$ & 4 & 3\% & 15 & 5\% \\
 & $\times$ / \checkmark & 8 & 6\% & 28 & 9\% \\
 & $\times$ / $\times$ & 115 & 89\% & 275 & 85\% \\
\bottomrule
\end{tabular}
\end{table}

\paragraph{Per-model and per-family results.}
Table~\ref{tab:cross_model} reports Set-A results by model. Fable-5.1
has 70.0\% own success and 40.0\% probe-in-scene success on this
conditioned subset. GPT-6 Astra has 35.0\% own success, while its scenes
support the probe on 5.0\% of rows and open-loop HAR-action replay on
30.0\%. This is the largest disagreement between the two scene-transfer
diagnostics among the evaluated models, and we have not traced its cause
at the instance level. With roughly 20 rows per model and confidence
intervals spanning approximately 20--40 percentage points, these
per-model differences should be treated as descriptive.

At the family level, DROID pick-place is the only Set-A family whose
agent scenes support the probe at an appreciable rate; leg insertion,
bulb insertion, and HOT3D placement rarely do. Drawer insertion reaches
23.3\% own success (10/43; Opus 5 4/5, Fable-5.1 3/5, DeepSeek 2/5,
GPT-5.6 1/5), but neither the HAR actions nor the probe produce a passing
HAR-scene execution for that family. Its own successes therefore cannot be
attributed by the present cross-execution diagnostic. Full
model$\times$family results are released with the benchmark.

\begin{table}[h]
\centering
\footnotesize
\setlength{\tabcolsep}{3.5pt} 
\renewcommand{\arraystretch}{0.95} 
\caption{Set-A cross-execution success (\%) per model; 20 valid rows per
model unless noted. Instance-clustered 95\% intervals span approximately
20--40 percentage points, so per-model differences are descriptive.}
\label{tab:cross_model}
\begin{tabular}{lrrrrr}
\toprule
Model & Own & Probe $\times$ agent
& HAR $\times$ agent
& Agent $\times$ HAR (subm.)
& Agent $\times$ HAR (anch.)$^\dagger$ \\
\midrule
Fable-5.1            & 70.0 & 40.0 & 25.0 & 5.0 & 35.0 \\
Claude Opus 5        & 42.1 & 21.1 & 21.1 & 5.3 & 10.5 \\
GPT-6 Astra          & 35.0 & 5.0  & 30.0 & 5.0 & 25.0 \\
GPT-5.6 Sol          & 15.0 & 15.8 & 10.0 & 0.0 & 15.0 \\
DeepSeek V4.1 Flash  & 15.0 & 5.0  & 5.3  & 0.0 & 15.8 \\
Gemini 3.8 Flash     & 15.0 & 15.0 & 10.0 & 5.0 & 0.0 \\
Kimi K3              & 10.0 & 5.0  & 5.0  & 0.0 & 15.0 \\
Qwen3.8 Max          & 10.0 & 5.0  & 0.0  & 0.0 & 5.6 \\
GLM-5.3 Flash        & 5.6  & 5.9  & 0.0  & 0.0 & 16.7 \\
\bottomrule
\end{tabular}
\end{table}

\paragraph{Limitations.}
Coverage is the main limitation. Set A contains only 20 instances across
four families, with 16 belonging to DROID or leg insertion, while several
assembly families have no passing HAR-action or probe execution and
therefore cannot enter the conditioned analysis. Set B provides broader
coverage but remains a selected subset of the full benchmark.

The probe is a fixed state-aware controller whose grasp pose is derived
from the HAR execution. Failure of the probe in an agent scene therefore
shows that the submitted scene is insufficiently compatible with this
specific transferred controller; it does not show that no controller
could successfully act in that scene. HAR-action replay is open-loop and
imposes a different, generally stricter transfer test, which explains why
the two scene diagnostics agree in aggregate direction but can diverge on
individual models. The anchored agent-action cell uses privileged initial
object position and should be interpreted only as a diagnostic of
trajectory compatibility after removing translational initialization
error.

Finally, the main benchmark documents execution-condition differences
between HAR and coding-agent workflows. The present study retains the
available benchmark execution configurations and therefore does not
constitute a perfectly matched causal intervention on scene and behavior
alone. Each reported row also comes from a single frozen generation from
the main experiment, with no repeated reconstruction runs.

\paragraph{Can agent-built worlds support experience beyond the source video?}
We test whether successful reconstructions support new configurations using nine source videos. For each evaluated model, we use its successful initial reconstruction of each source video. Each reconstruction is tested on 20 layouts formed by translating either the manipulated object or the target in two predefined table directions at five magnitudes (2, 5, 10, 15, and 20\,cm), yielding 180 paired layout evaluations per model. For each source, both directions are verified across all five magnitudes using the human-assisted reconstruction before evaluating candidate layouts. The same videos and displacements are used across models. This experiment therefore measures reuse conditional on successful initial reconstruction and the prescribed layout selection.

\emph{Direct reuse} executes the original robot behavior without code changes, retaining any existing feedback or planning logic. \emph{Agent adaptation} starts from the same original submission and displaced layout, and allows the corresponding coding agent to revise robot behavior using simulation feedback. The agent receives the source video, original submission, and layout change. The evaluator fixes the displaced scene, assets, physical parameters, and task criteria throughout adaptation. Both settings use the same runtime observation interface. Adaptation starts with a fresh context for each layout and stops after 10 rollout attempts or 10 minutes; the final behavior is evaluated from a clean reset.

\begin{figure}[t]
    \centering
    \includegraphics[width=\columnwidth]{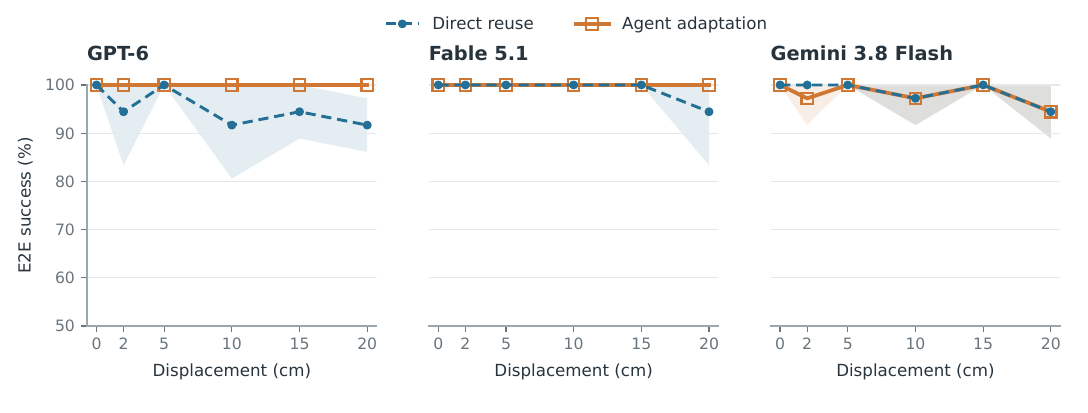}
    \caption{
        \textbf{Robot Experience Beyond the Source Video.} Direct reuse and agent adaptation across translated object and target layouts. The same nine source videos (three per task family) contribute at every displacement, with 180 paired layout evaluations per model. Each source uses the same two table directions and both moved-object roles at every magnitude. Both directions were verified using the human-assisted reconstruction jointly across all five magnitudes before new candidate evaluations. Success is averaged within video and then across task families. Shading shows 95\% paired source-video bootstrap intervals within family. The 0\,cm points are verified original baselines without adaptation.
    }
    \label{fig:simulation_reuse}
\end{figure}

Figure~\ref{fig:simulation_reuse} shows that successful reconstructions can support executions beyond the observed layout. Across the three models and 540 paired layout evaluations, direct reuse succeeds in 97.22\% of cases (525/540), increasing to 99.26\% (536/540) after adaptation. Adaptation recovers 86.67\% of direct-reuse failures (13/15), while turning 0.38\% of initially successful cases into failures (2/525), yielding a net gain of 11 successful executions. At the largest displacement of 20\,cm, success increases from 93.52\% to 98.15\%, although performance is not monotonic in displacement.

The benefit varies across models. GPT-6 improves from 94.44\% to 100\%, and Fable~5.1 from 98.89\% to 100\%, recovering all ten and two of their respective direct-reuse failures. Gemini~3.8 Flash decreases slightly from 98.33\% to 97.78\%: it recovers one failure but introduces two new failures. Adaptation therefore provides a means of repairing behavior under changed layouts, but does not guarantee improvement. Across the 540 evaluated adaptation sessions, including unsuccessful outcomes, the mean cost is 1.06 rollout attempts and 3.26 minutes per layout, excluding additional infrastructure and recovery launches.

These results demonstrate substantial layout reuse within the selected cohort, with adaptation providing a modest aggregate gain over an already strong direct-reuse baseline. The evidence is conditional on successful initial reconstruction and reference-verified translations of the same nine source tasks; it does not establish generalization to new task families or improved downstream policy learning.

\section{Cost Analysis}
\label{app:cost}

All methods were assigned the same nominal per-episode budgets: a 3-hour wall-clock limit, a \$60 API-spend threshold, and at most three package submissions (Appendix~\ref{app:interface_budget}). The benchmark-managed executor directly enforces these limits for the seven OpenCode configurations. Fable-5.1 and GPT-6 Astra run through their native agent harnesses under the corresponding limits, while retaining framework-specific checkpoint and refinement semantics. Table~\ref{tab:cost} reports per-episode cost. For the OpenCode configurations, API cost is the provider-billed amount reported through OpenRouter. For Fable-5.1 and GPT-6 Astra, it is the list-price equivalent computed from the native harness token counts; comparable API-call counts were not recorded. Evaluation uses one NVIDIA L40S per episode for reference execution, simulation of the delivered package, and metric computation.

\begin{table}[!htbp]
\centering
\small
\setlength{\tabcolsep}{4pt}
\caption{\textbf{Per-episode cost of each method.} API cost is in USD; agent time is wall-clock time from the first model call to termination; evaluation time is GPU-minutes on one L40S. Estimated totals extrapolate the reported per-episode means over all 222 benchmark instances.}
\label{tab:cost}
\begin{tabular}{@{}lrrrrrrr@{}}
\toprule
& \multicolumn{4}{c}{API cost (USD)} & Agent time & API calls & Eval. GPU \\
\cmidrule(lr){2-5}
Method & mean & median & max & est. total & mean (min) & mean & mean (min) \\
\midrule
Fable-5.1$^{\dagger}$   & 14.03 & 12.90 & 45.19 & 3{,}115 & 37.5 & -- & 10.1 \\
GPT-6 Astra$^{\dagger}$ & 6.40 & 3.91 & 24.58 & 1{,}421 & 14.6 & -- & 9.7 \\
Claude Opus 5           & 13.83 & 10.03 & 61.07 & 3{,}070 & 32.6 & 59 & 8.7 \\
Kimi K3                 & 5.51 & 4.81 & 13.27 & 1{,}223 & 82.2 & 72 & 7.4 \\
GPT-5.6 Sol             & 1.39 & 1.00 & 7.89 & 309 & 3.7 & 17 & 8.2 \\
DeepSeek V4.1 Flash     & 0.23 & 0.21 & 0.62 & 51 & 13.9 & 53 & 8.3 \\
Gemini 3.8 Flash        & 1.53 & 1.02 & 8.15 & 340 & 18.6 & 100 & 7.5 \\
Qwen3.8 Max             & 0.67 & 0.60 & 2.26 & 149 & 26.0 & 34 & 8.4 \\
GLM-5.3 Flash           & 0.31 & 0.23 & 1.46 & 69 & 53.0 & 60 & 8.9 \\
\midrule
All methods             & 4.88 & -- & 61.07 & 9{,}746 & 31.3 & -- & 8.6 \\
\bottomrule
\end{tabular}
\vspace{2pt}

{\footnotesize $^{\dagger}$List-price equivalent computed from token counts recorded by the native harness; API-call counts were not recorded. Estimated totals are the displayed means multiplied by 222 and rounded to the nearest dollar.}
\end{table}

Running all nine methods over the 222-instance benchmark corresponds to approximately \$9.7k in API spend, 1{,}044 agent-hours, and 286 L40S GPU-hours of evaluation at the reported per-episode averages. API spend is concentrated in Fable-5.1 and Claude Opus 5, which each cost about \$14 per episode and together account for approximately 63\% of the estimated total. The flash-tier configurations average less than \$2 per episode. Cost does not track score monotonically: GPT-6 Astra achieves the second-highest V2WScore at about \$6 per episode, while Kimi K3 uses the most agent time, averaging 82 minutes per episode, without a corresponding score advantage.

No episode exceeds the 3-hour wall-clock allowance; the longest recorded episode lasts 164 minutes. One Claude Opus 5 episode records \$61.07 because delayed provider reporting can overshoot the \$60 observed-spend threshold. Evaluation cost varies comparatively little across methods, from 7.4 to 10.1 GPU-minutes per episode. At the reported average, adding a method over all 222 instances requires approximately 32 L40S GPU-hours of evaluation, in addition to its generation spend and agent time.

\section{Case Studies Across Demonstration Sources}
\label{app:case_studies}

We examine six demonstrations to explain how geometric fidelity, executed object motion, and task completion can disagree. The cases cover real robot demonstrations, egocentric human video, and simulation-generated robot demonstrations. Three detailed reconstructions use Claude Opus~5; two further demonstrations permit a comparison of all nine coding models on the same instance; and a native stacking task is examined under two target embodiments. Cases were selected for their diagnostic value, rather than to estimate the prevalence of particular errors or rank models.

Each result refers to a particular submitted scene, its own robot behavior, and the recorded physical execution. Scores are taken from that submission's retained evaluation record. The accompanying evidence identifies the instance, model, package, scoring record, and execution used for each analysis. Image sequences are arranged by task phase unless an explicit time axis is shown; phase correspondence does not imply equal execution speed. Scene Chamfer Distance (Scene CD), translation absolute pose error (APE), and Task Success retain their distinct meanings from Appendix~\ref{app:evaluation}.

\subsection{Robot Demonstrations: Pose and Task-Critical Geometry}
\label{app:case_robot}

The first two cases reconstruct DROID demonstrations with a Panda arm and Robotiq gripper. They illustrate two different limitations of interpreting a favorable geometric or endpoint measurement as complete reconstruction quality.

\paragraph{Task completion despite pose disagreement.}
In the first demonstration, an orange block is placed on a green block. The Opus reconstruction completes this interaction and reaches all four recorded progress stages. Its terminal relative-position error is 1.08\,cm, with translation APE of 2.87\,cm and translation RPE of 1.80\,cm. Geometry also gives small surface errors: Scene CD is 1.23\,cm and object Shape CD is 0.77\,cm. Nevertheless, initial centre error is 5.92\,cm, size error is 1.31\,cm, and rotation APE is $144^\circ$.

\begin{figure}[!t]
    \centering
    \includegraphics[width=\columnwidth]{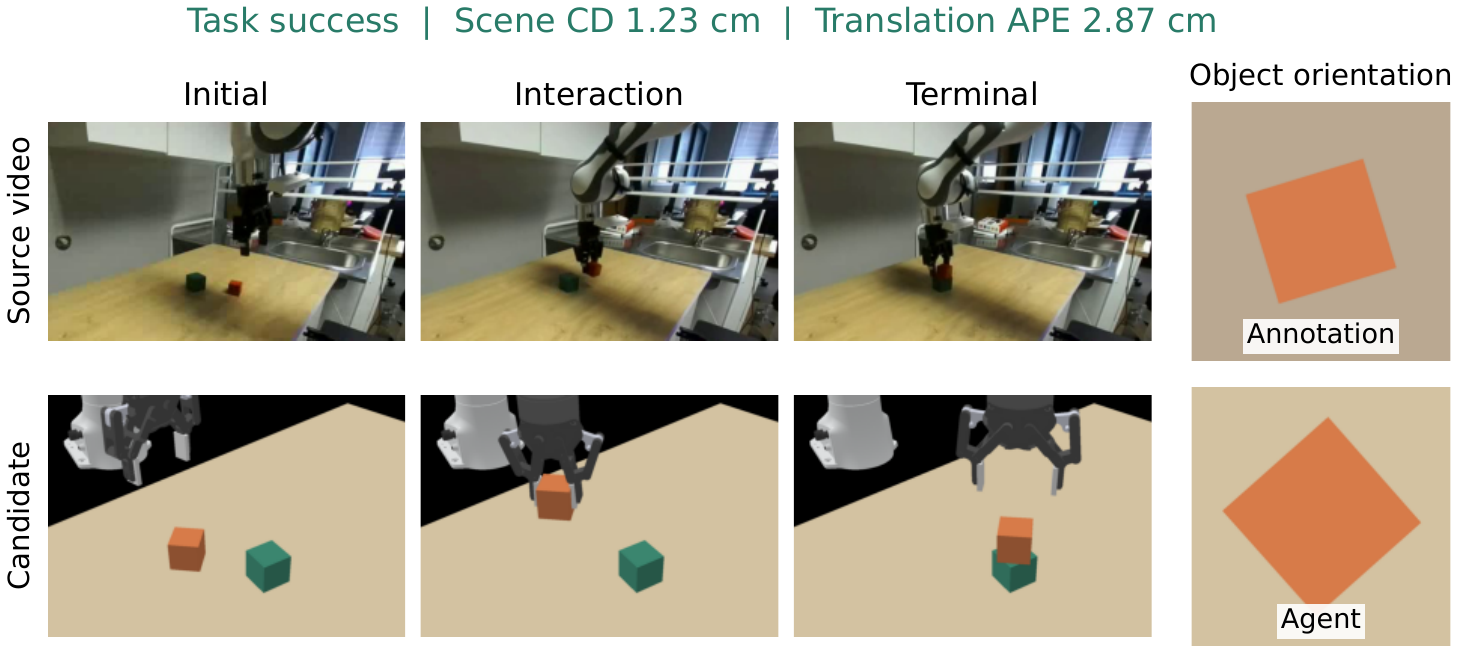}
    \caption{\textbf{Task completion despite pose disagreement.} DROID block stacking with a Panda--Robotiq embodiment. The source and candidate execution are shown at three task phases. The top-view inset compares the submitted and annotated block orientations. Successful stacking coexists with a large object-frame rotation error; the rigidly aligned shape score does not measure the submitted world orientation.}
    \label{fig:case_robot}
\end{figure}

Figure~\ref{fig:case_robot} shows why these quantities need separate interpretation. The submitted object is a 4.5\,cm cube, whereas the annotated dimensions are approximately $3.5\times3.8\times4.1$\,cm. Its submitted yaw is $-138.4^\circ$, compared with $17^\circ$ in the annotation. Reducing the cube's yaw modulo its quarter-turn symmetry gives $41.6^\circ$; even this surface-equivalent representative differs from the annotated orientation. Shape CD is evaluated after rigid alignment, so it does not measure this world-pose discrepancy. Scene CD measures registered visible surfaces and can remain small for a compact, nearly symmetric object. Rotation APE instead evaluates the corresponding object frames under the annotated symmetry, which does not remove the reported disagreement.

Successful stacking therefore establishes that the submitted behavior can manipulate the submitted block and satisfy the specified relation. It does not establish accurate object dimensions, initial placement, or orientation history. Conversely, the large angular discrepancy does not negate the observed completion of an orientation-tolerant stacking task.

\paragraph{A plausible endpoint without the required receiving geometry.}
In the second demonstration, a blue block is inserted through an opening in the top of a shape-sorting box. The reconstruction grasps, lifts, and releases the block, which rests only 1.12\,cm from the annotated receiver-relative goal. Translation APE is 6.75\,cm. Despite this favorable endpoint, Task Success is zero and Task Progress is $75\%$.

\begin{figure}[!t]
    \centering
    \includegraphics[width=\columnwidth]{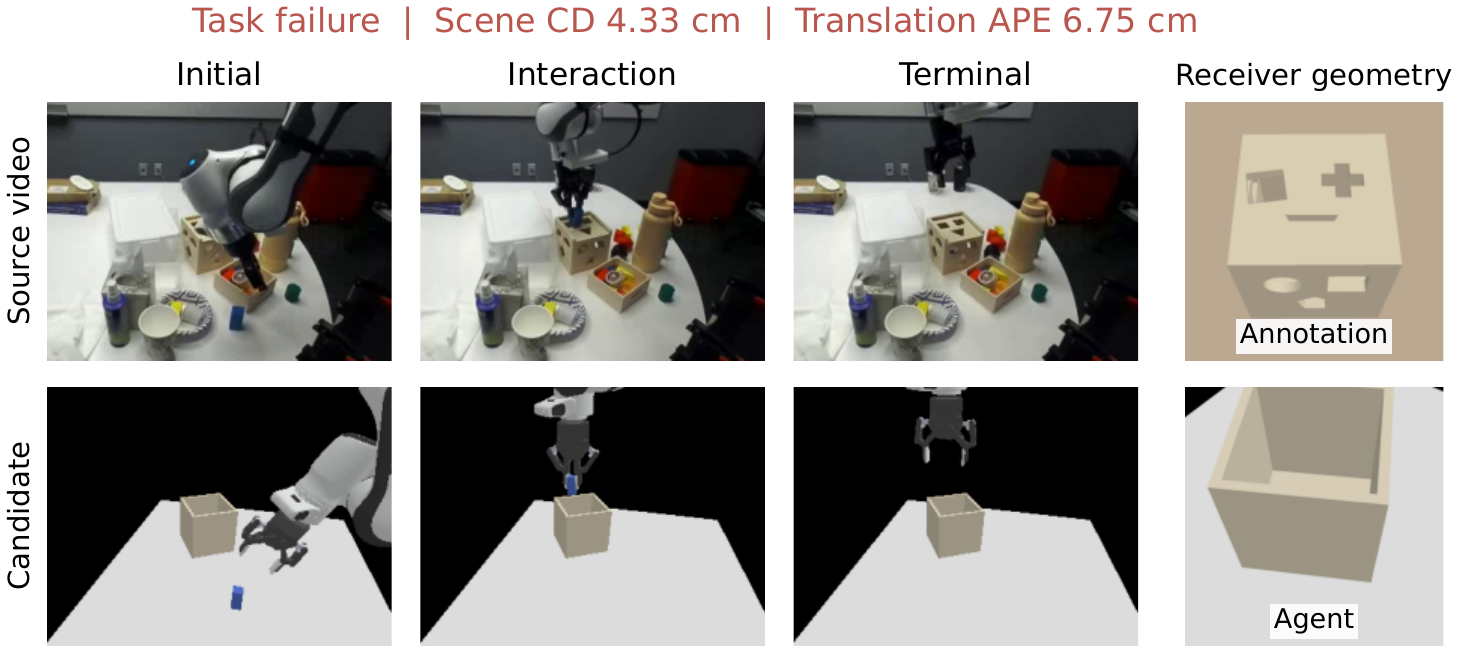}
    \caption{\textbf{Endpoint proximity does not establish the required interaction.} DROID shape-sorter insertion. The block is transferred into the submitted open container, but the receiver lacks the apertured top visible in the annotation. The final relative-position error passes its tolerance; top-face coverage fails its separate requirement. The source and candidate rows show task phases, rather than equal timestamps.}
    \label{fig:case_lid}
\end{figure}

The submitted receiver is an open container with 8\,mm walls, rather than a box with an apertured top (Figure~\ref{fig:case_lid}). Its top-face coverage is 0.23, below the required 0.50; the annotated box has coverage 0.85. The receiver therefore lacks a task-defining structure, even though the block reaches an admissible relative position. The coverage test verifies the presence of upper receiving geometry; it should not be interpreted as an independent measurement of passage through a particular aperture.

Scene CD is 4.33\,cm, while the manipulated block's Shape CD is 0.70\,cm. The weighted missing-geometry fraction is 0.50. Here that diagnostic does not mean that the block was omitted: the delivered block exceeds the surface-distance threshold after registration. These measurements describe different aspects of the reconstruction. Surface discrepancy characterizes geometric correspondence, whereas the task-specific receiver condition determines whether the intended insertion environment was constructed. An endpoint tolerance alone would miss this failure.

\subsection{Nine Models on the Same Instances}
\label{app:case_agents}

\begin{figure}[!p]
    \centering
    \includegraphics[width=\columnwidth]{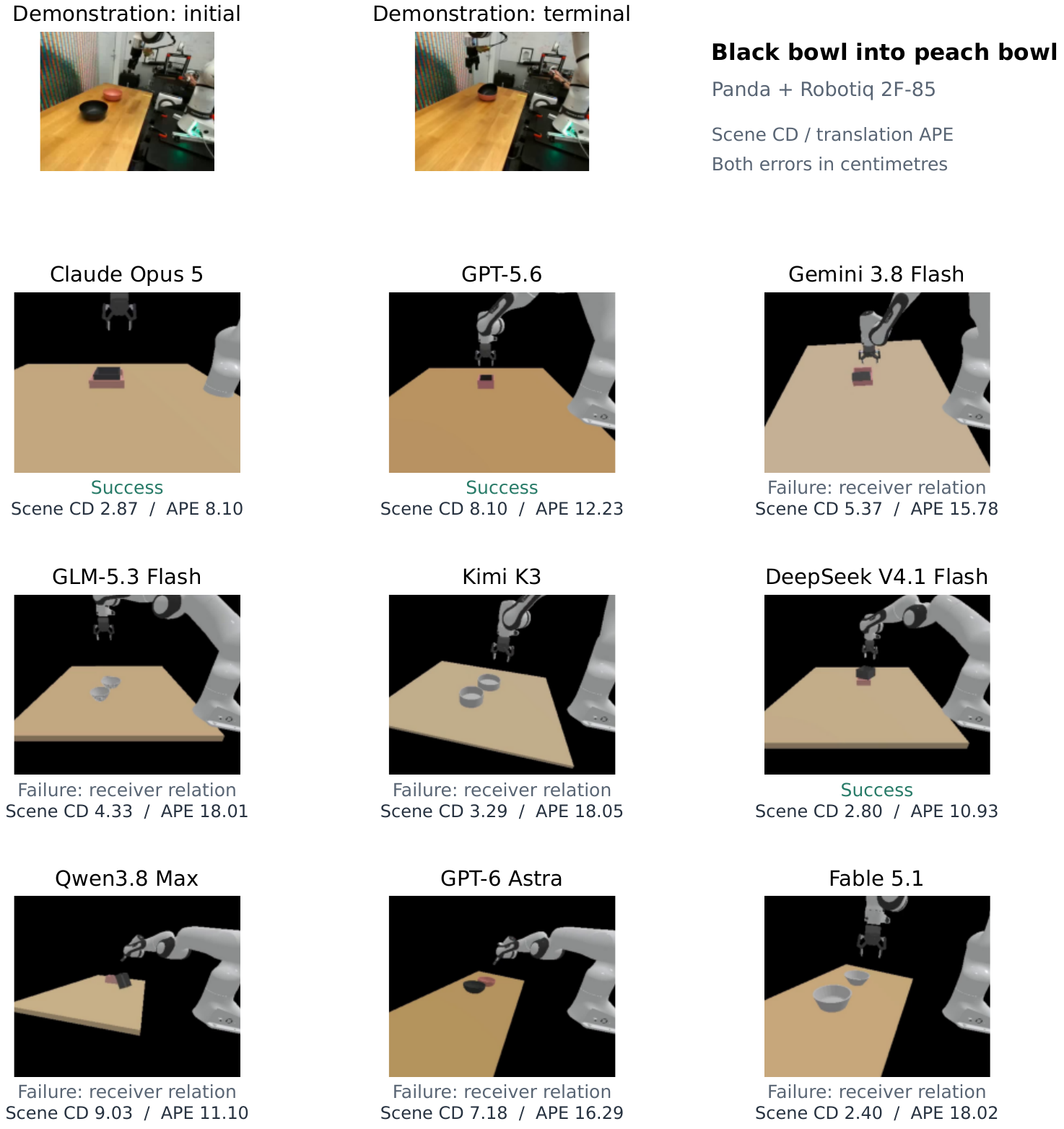}
    \caption{\textbf{Nine reconstructions of the same bowl-nesting demonstration.} Top: the source video's initial and terminal configurations. The nine model panels show the last frame of each submission's evaluation review video, using a common crop within this instance. Scene CD and translation APE are in centimetres. Opus, GPT-5.6, and DeepSeek succeed; the other six fail the receiving relation. Fable's smaller surface error than Opus does not imply successful placement. Scores and decisions come from the complete execution records, including terminal settling; the displayed video endpoints are qualitative views rather than substitutes for those records.}
    \label{fig:case_agents_stack}
\end{figure}

\begin{figure}[!p]
    \centering
    \includegraphics[width=\columnwidth]{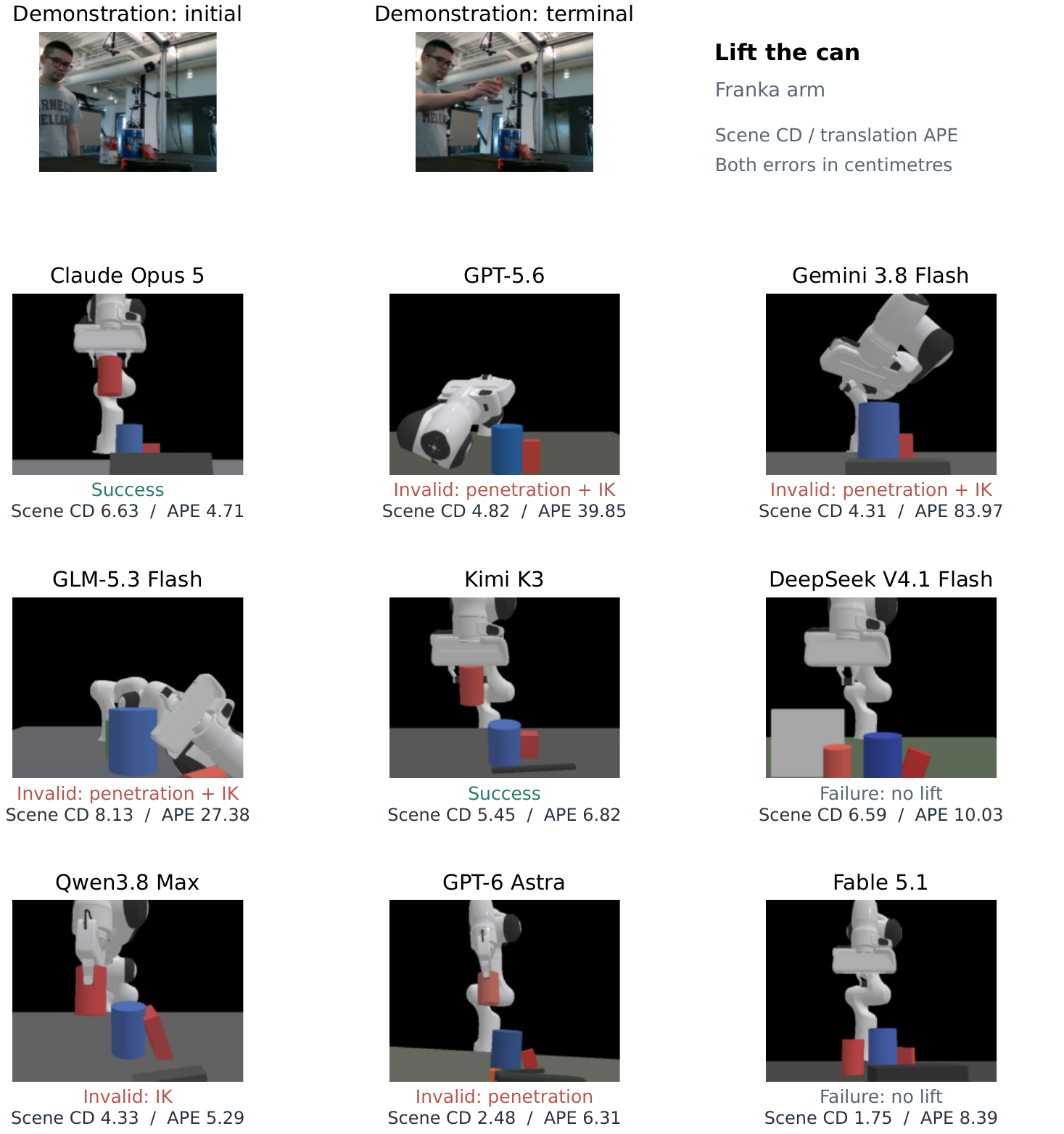}
    \caption{\textbf{Nine reconstructions of the same can-lifting demonstration.} Source frames and model panels follow Figure~\ref{fig:case_agents_stack}, with Scene CD and translation APE in centimetres. Opus and Kimi succeed. Qwen and Astra meet the ungated terminal lift condition but violate inverse-kinematics feasibility and robot-penetration checks, respectively. DeepSeek and Fable have valid executions without the required lift. The remaining three violate both checks. In particular, an elevated can in a rendered frame does not establish physical validity, which is determined from the execution record.}
    \label{fig:case_agents_lift}
\end{figure}

\begin{figure}[!t]
    \centering
    \includegraphics[width=\columnwidth]{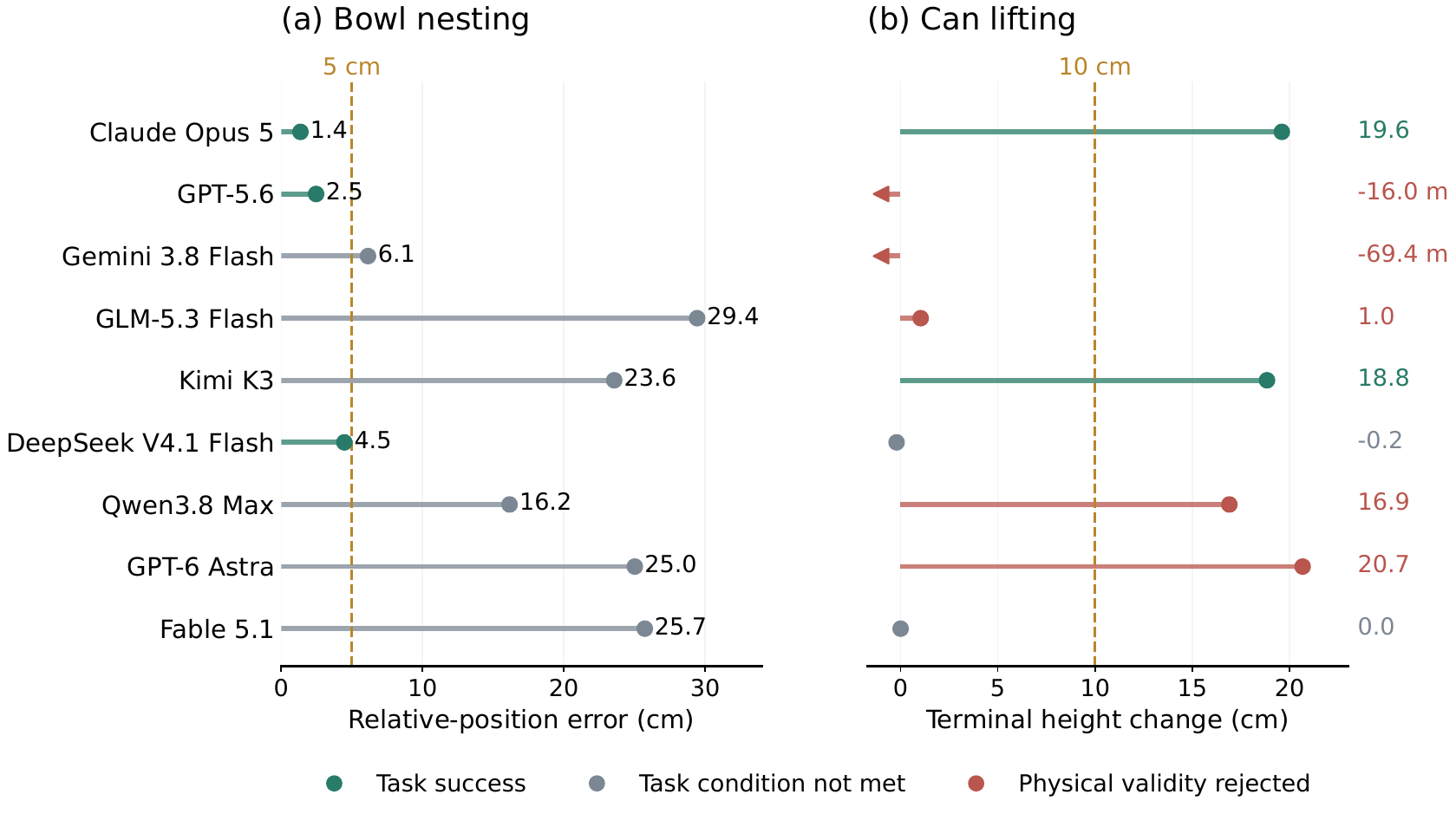}
    \caption{\textbf{Nine models, two tasks, and different reasons for failure.} (a) Terminal receiver-relative position error for black-bowl nesting. (b) Terminal can-height change for DexYCB lifting. Dashed lines mark the task-specific positional and height thresholds, which are components of the complete predicates. Green denotes Task Success, grey an unmet task condition, and red rejection by physical-validity checks. Qwen and Astra exceed the lift threshold but are rejected for inverse-kinematics infeasibility and robot penetration, respectively. Left-pointing markers denote out-of-range negative displacements, with their values stated explicitly. No aggregate model ranking is inferred from these two selected cases.}
    \label{fig:case_agents_comparison}
\end{figure}

We next compare Claude Opus~5, GPT-5.6, Gemini~3.8~Flash, GLM-5.3~Flash, Kimi~K3, DeepSeek~V4.1~Flash, Qwen3.8~Max, GPT-6 Astra, and Fable~5.1 on two shared instances. All nine have a completed physical execution and evaluation record for both cases. Comparisons below hold the demonstration, target embodiment, and task specification fixed within each instance. They concern these particular submissions, rather than model-level success rates. Figures~\ref{fig:case_agents_stack} and~\ref{fig:case_agents_lift} show every model's reconstruction, with the same model order in both grids. Each panel pairs the last recorded review-video frame with the submission's geometric error, motion error, and complete evaluation decision. Figure~\ref{fig:case_agents_comparison} complements these views with the task-specific decision variables.

\paragraph{Receiver-relative completion in bowl nesting.}
The DROID task is to place a black bowl \emph{inside a peach-coloured bowl}. The receiver is therefore another bowl, not the tabletop. Opus, GPT-5.6, and DeepSeek complete the task, with terminal relative-position errors of 1.36, 2.47, and 4.47\,cm, respectively. Gemini, Qwen, Kimi, GLM, Astra, and Fable fail, with errors of 6.14, 16.16, 23.57, 29.43, 25.02, and 25.73\,cm. Figure~\ref{fig:case_agents_comparison}(a) shows these errors alongside the 5\,cm positional tolerance. The plotted distance is one component of the decision; completion also uses the recorded interaction, release, and terminal-rest conditions.

Geometric correspondence does not determine these outcomes. Fable has a lower Scene CD than Opus, 2.40 versus 2.87\,cm, but fails to establish the receiving relation. Their translation APE values are 18.02 and 8.10\,cm, respectively. GPT-5.6 succeeds despite a Scene CD of 8.10\,cm and translation APE of 12.23\,cm. These examples show that geometric fidelity and task completion can disagree within a fixed instance; they do not establish statistical independence between the two dimensions.

\paragraph{Terminal lift versus physically valid execution.}
The second instance is a DexYCB demonstration of lifting a can, retargeted to a Franka arm. Opus and Kimi succeed, with terminal height increases of 19.60 and 18.84\,cm, above the 10\,cm requirement. Their executions also pass the applicable physical checks. Their translation APE values are 4.71 and 6.82\,cm, and their Scene CD values are 6.63 and 5.45\,cm.

Qwen and Astra satisfy the ungated terminal task condition: their height increases are 16.90 and 20.67\,cm, and the object is at rest. Both nevertheless receive zero Task Success. Qwen violates the inverse-kinematics feasibility check; Astra violates the robot-penetration check. These are distinct recorded violations. Astra's Scene CD is 2.48\,cm and translation APE is 6.31\,cm, illustrating why neither favorable surface correspondence nor a close object trajectory replaces physical validity.

DeepSeek and Fable instead pass the physical checks but do not complete the lift. The object's terminal height changes by approximately $-0.20$\,cm and $0.006$\,cm, respectively. Fable's Scene CD is 1.75\,cm, the lowest among these nine reconstructions, while translation APE is 8.39\,cm. Its geometry is therefore relatively accurate under the recorded surface metric, but its behavior does not achieve the required state change. GPT-5.6, Gemini, and GLM record both robot penetration and infeasible inverse kinematics. The first two also have extreme negative terminal displacements, approximately $-16.0$ and $-69.4$\,m; these are abnormal executed states, not plausible lifting motions. GLM's terminal rise is only 1.04\,cm and does not meet the height requirement.

Figure~\ref{fig:case_agents_comparison}(b) distinguishes valid but unsuccessful execution from execution rejected by physical checks. In particular, the Qwen and Astra outcomes demonstrate that the validity condition changes the success decision even when the ungated terminal task condition passes. The height and rest tests alone do not independently certify sustained grasp contact; claims about holding must rely on the corresponding execution evidence.

\subsection{Egocentric Human Demonstrations under a Robot Embodiment}
\label{app:case_human}

\paragraph{Task completion with spatial and temporal mismatch.}
A HOT3D head-camera recording shows a person moving a food package into a bowl. Opus reconstructs the scene and executes the interaction with a Franka arm. The execution passes the applicable physical checks and all five recorded progress stages. The package finishes 2.7\,cm from the axis of the submitted bowl, whose rim radius is 14.0\,cm. Its terminal relative-position error against the demonstrated relation is 6.88\,cm. This error is a diagnostic rather than the bowl-containment success threshold.

\begin{figure}[!t]
    \centering
    \includegraphics[width=\columnwidth]{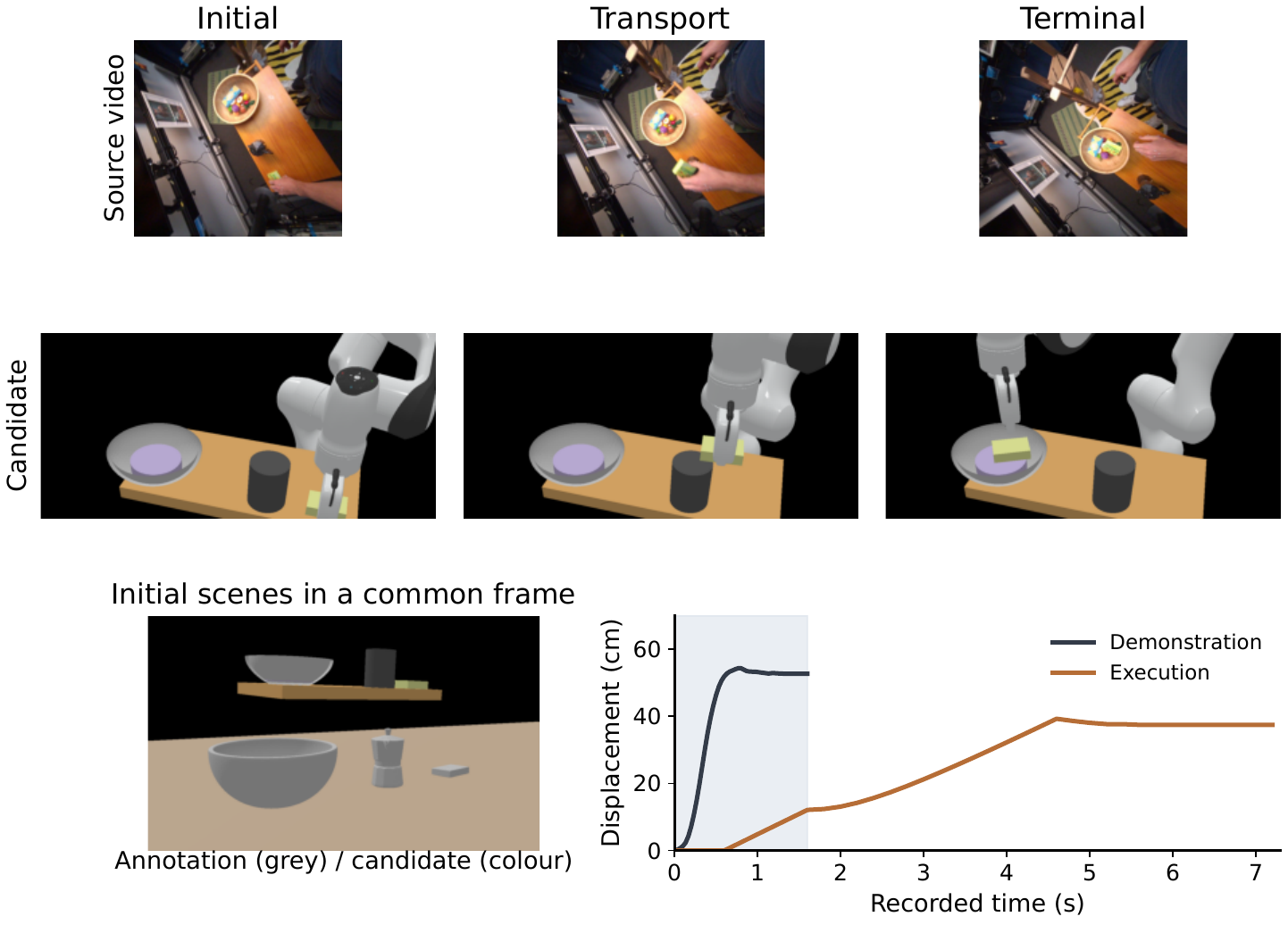}
    \caption{\textbf{Successful containment with spatial and temporal mismatch.} HOT3D food-package placement by a Franka arm. Top: source and candidate execution at selected task phases. Bottom left: the annotated scene in grey and the reconstruction in colour, viewed from the side in a common frame. Bottom right: displacement of each recorded object origin from its own initial position. The shaded interval is the 1.6\,s demonstration window; the candidate execution continues to 7.2\,s. This displacement plot illustrates timing and is not itself an APE curve.}
    \label{fig:case_human}
\end{figure}

The successful interaction occurs in a displaced reconstruction. Scene CD is 6.79\,cm, initial object-centre error is 28.2\,cm, and the recorded scene-registration translation has magnitude 28.4\,cm. The submitted tabletop lies approximately 26\,cm above the annotated support. The side view in Figure~\ref{fig:case_human} exposes this disagreement, which is less apparent in the original camera view. The submitted package is also initially 38.6\,cm from its bowl in the support plane, compared with 55.3\,cm in the annotation. A single global registration cannot remove this relative-layout discrepancy.

The weighted missing-geometry fraction is 0.49, although the package and kettle are present in the submission. Their registered median surface residuals are approximately 17.1 and 5.1\,cm, so they exceed the coverage threshold. As in the blue-block case, this diagnostic describes inadequate surface correspondence rather than necessarily absent semantic entities.

\paragraph{The execution clock explains an additional discrepancy.}
Translation APE is 41.2\,cm, rotation APE is $167^\circ$, and translation RPE is 10.6\,cm. The submitted behavior contains 37 command rows at 0.2\,s intervals, spanning 7.2\,s, whereas the demonstration spans 1.6\,s. On the demonstration clock, much of the demonstrated transport is compared with an execution that has not yet performed the corresponding motion. Figure~\ref{fig:case_human} plots displacement against the separate recorded clocks, without stretching either trajectory. The recorded duration-normalized translation error is lower, 18.1\,cm, but remains a diagnostic outside the reported APE.

The instance therefore demonstrates two independent limitations of task completion as a reconstruction criterion. The containment condition can be satisfied relative to the submitted bowl even when the scene is globally displaced, and the same interaction can be completed at a substantially different rate. Geometry and motion fidelity expose those discrepancies separately.

\subsection{Native Robot Demonstrations across Target Embodiments}
\label{app:case_twin}

\begin{figure}[!p]
    \centering
    \includegraphics[width=0.96\columnwidth]{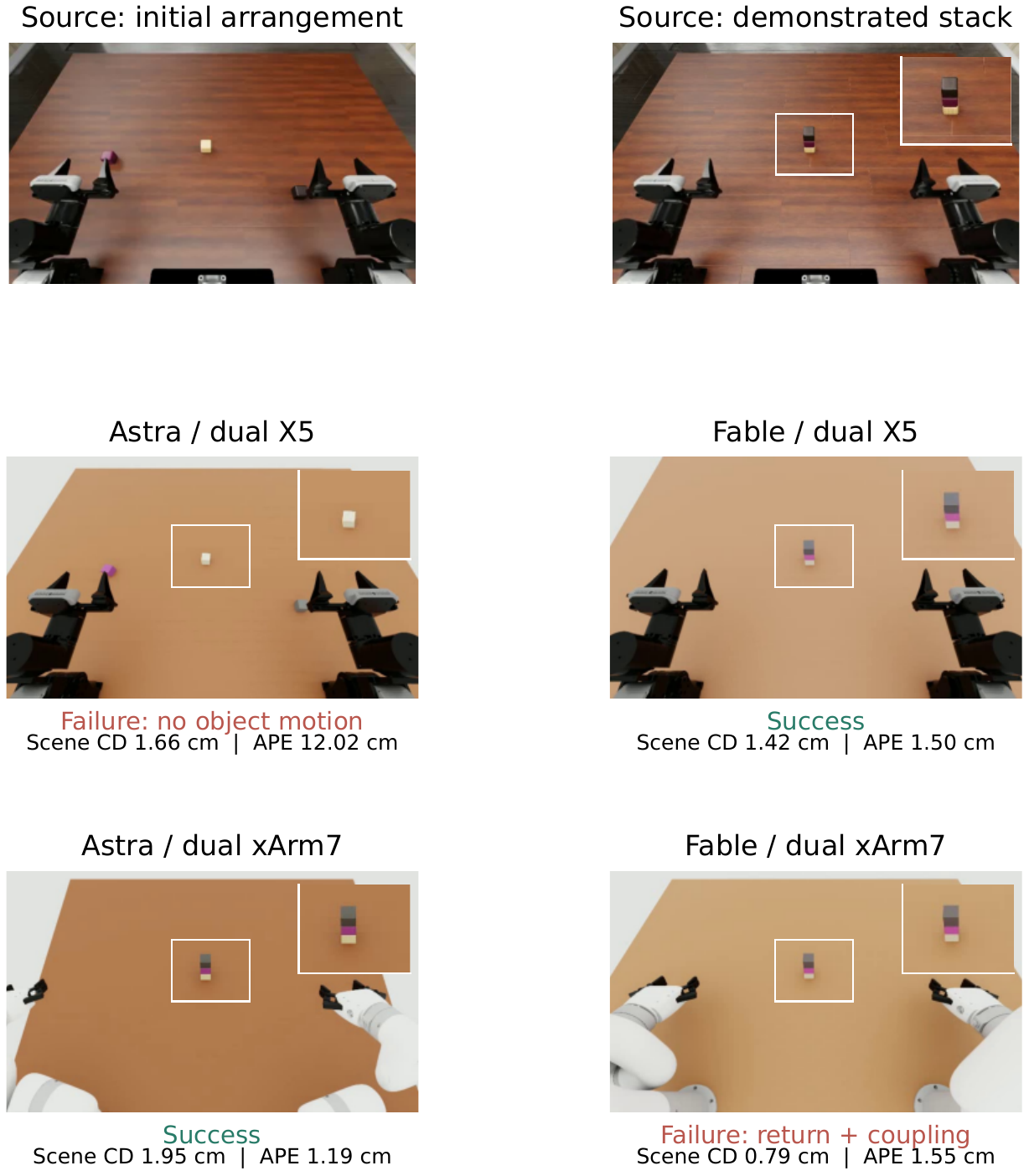}
    \caption{\textbf{One demonstration, two target robots, and distinct execution outcomes.} Top: the RoboDojo source video's initial and terminal arrangements. Middle and bottom: terminal states of Astra and Fable with X5 and xArm7 arms. Each panel is drawn from its own recorded physical execution; insets magnify the outlined receiver region. Astra's X5 objects remain stationary, while its xArm7 submission succeeds. Fable succeeds with X5; its xArm7 stack satisfies the object-stacking subpredicate but fails robot return and gripper-coupling checks. Scene CD and translation APE are shown in centimetres. A visually plausible final stack therefore does not certify the complete interaction.}
    \label{fig:case_twin}
\end{figure}

The final demonstration is a RoboDojo three-block stacking task, assigned separately to dual ARX X5 arms and dual xArm7 arms. Both configurations use the same source video and semantic objective. We examine the formally admitted instances and the submitted executions of Astra and Fable. Figure~\ref{fig:case_twin} shows the demonstrated arrangement and the four recorded terminal states.

\paragraph{An accurate scene without the required object motion.}
Astra's X5 reconstruction has Scene CD of 1.66\,cm but zero Task Progress and zero Task Success. The three blocks remain stationary to numerical precision throughout the recorded execution, and the native stacking predicate is false. Translation APE is 12.02\,cm, averaged over the task objects, while terminal relative-position error is 33.99\,cm. These observations locate the failure in the realized interaction without attributing an unmeasured contact or actuator mechanism to it.

Under the xArm7 embodiment, Astra succeeds on the same demonstration. Scene CD is 1.95\,cm, translation APE is 1.19\,cm, and terminal relative-position error is 0.16\,cm. The different outcomes illustrate why the target robot and its control interface are part of the reconstruction instance: scene correspondence alone does not establish that the submitted behavior will manipulate the objects under that embodiment.

\paragraph{A correct-looking stack with unmet execution requirements.}
Fable succeeds with the X5 arms, with Scene CD of 1.42\,cm, translation APE of 1.50\,cm, and terminal relative-position error of 0.67\,cm. Its xArm7 execution produces an apparently correct stack and has even smaller geometric and endpoint errors: Scene CD is 0.79\,cm and terminal relative-position error is 0.45\,cm, with translation APE of 1.55\,cm. Nevertheless, Task Success and reported progress are zero.

The execution record identifies two deficiencies that a still image of the stack does not establish. The stacking subpredicate passes, but the required robot-return condition fails. In addition, coupled gripper joints violate their prescribed mechanical relation: the largest recorded follower-joint residual is approximately 0.685\,rad, against a tolerance of 0.02\,rad. The evaluator rejects this execution and suppresses its raw partial progress of 0.15. Thus, the favorable object arrangement is insufficient both as evidence of the complete task and as evidence of valid robot mechanics. The native checks certify their recorded constraints; they do not constitute a comprehensive collision certificate.

\section{Limitations and Future Work}

Video2World evaluates autonomous video-to-simulation end to end, while several aspects of interactive world reconstruction remain open. First, functional correctness is measured from the joint execution of the reconstructed environment and the agent-generated robot behavior. Consequently, a failed rollout may originate from errors in the reconstructed world, the behavior policy, or their interaction. Future evaluation could disentangle these factors through standardized probe behaviors, multiple independently generated controllers, or controlled interventions on the reconstructed environment.

Second, the current benchmark primarily evaluates whether a reconstructed world can reproduce the demonstrated interaction. This leaves open a stronger notion of world reconstruction: whether the recovered environment remains physically and functionally valid under actions, configurations, and goals that are not observed in the source video. Evaluating such counterfactual behavior could distinguish task-specific reconstructions from reusable interactive worlds and provide a more direct measure of whether an agent has recovered the underlying structure of the physical environment.

Third, measured performance reflects the complete coding-agent system, including the underlying foundation model, agent scaffold, context management, tool use, and execution strategy. This system-level evaluation matches how coding agents are deployed in practice, but does not isolate which component is responsible for performance differences. Future work could independently vary models, scaffolds, tools, and feedback mechanisms to better characterize the sources of progress in autonomous world construction.

More broadly, Video2World focuses on reconstruction from a single embodied demonstration. Extending this setting to multi-view observations, multiple demonstrations, active information gathering, and progressively accumulated experience could move autonomous video-to-simulation from reconstructing individual interactions toward learning persistent, reusable world models.

\end{document}